\documentclass[pdflatex,sn-mathphys-ay]{sn-jnl}

\usepackage{graphicx}%
\usepackage{multirow}%
\usepackage{amsmath,amssymb,amsfonts}%
\usepackage{amsthm}%
\usepackage{mathrsfs}%
\usepackage[title]{appendix}%
\usepackage{xcolor}%
\usepackage{textcomp}%
\usepackage{manyfoot}%
\usepackage{booktabs}%
\usepackage{algorithm}%
\usepackage{algorithmicx}%
\usepackage{algpseudocode}%
\usepackage{listings}%
\usepackage{subcaption}
\usepackage{natbib}
\usepackage{longtable}
\usepackage{bm}
\usepackage{float}
\usepackage{pdflscape}

\newcommand{\R}{\mathbb{R}}

\newcommand{\lb}{\left (}
\newcommand{\rb}{\right )}

\newcommand\eqn[1]{\begin{align*}#1\end{align*}}

\newcommand{\norm}[1]{\left\lVert#1\right\rVert}

\newcommand{\argmin}{\textrm{argmin}}

\theoremstyle{thmstyleone}%
\theoremstyle{thmstyletwo}%

\theoremstyle{thmstylethree}%

\begin{document}


\title[Intrinsic Dimension Estimators]{A Survey and Comparative Evaluation of Intrinsic Dimension Estimators under the Manifold Hypothesis}


\author*[1]{\fnm{Zelong} \sur{Bi}}\email{zelong.bi@unsw.edu.au}

\author[1]{\fnm{Pierre} \sur{Lafaye De Micheaux}}\email{lafaye@unsw.edu.au}


\affil*[1]{\orgdiv{School of Mathematics \& Statistics}, \orgname{University of New South Wales}, \orgaddress{\street{High Street}, \city{Kensington}, \postcode{2052}, \state{NSW}, \country{Australia}}}



\abstract{The manifold hypothesis suggests that high-dimensional data often lie on or near a low-dimensional manifold. Estimating the dimension of this manifold is essential for leveraging its structure, yet existing work on dimension estimation is fragmented and lacks systematic evaluation. This article provides a comprehensive survey for both researchers and practitioners. We review often-overlooked theoretical foundations and present eight representative estimators. Through controlled experiments, we analyze how individual factors, such as noise, curvature, and sample size, affect performance. We also compare the estimators on diverse synthetic and real-world datasets, introducing a principled approach to dataset-specific hyperparameter tuning. Our results offer practical guidance for estimator selection and yield insights that will inform future estimator design.}

\keywords{manifold hypothesis, intrinsic dimension, nonlinear dimension reduction, manifold learning}

\maketitle

\section{Introduction}

\begin{figure}[H]
    \centering
    \begin{subfigure}[b]{0.35\textwidth}
        \centering
        \includegraphics[width=\textwidth]{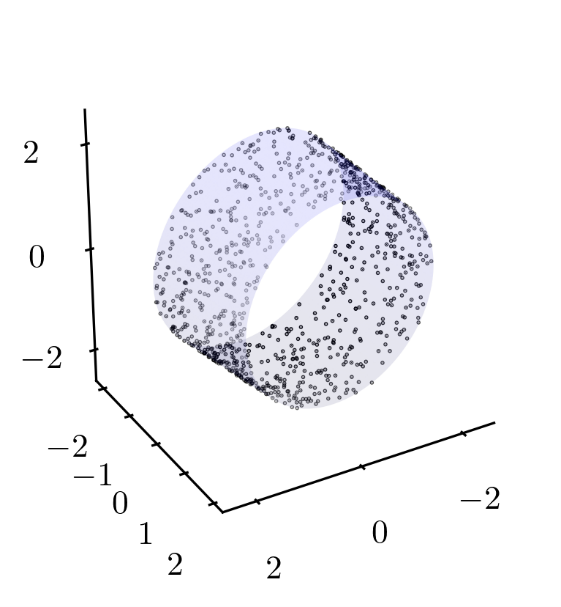}
    \end{subfigure}
    \hfill
    \begin{subfigure}[b]{0.35\textwidth}
        \centering
        \includegraphics[width=\textwidth]{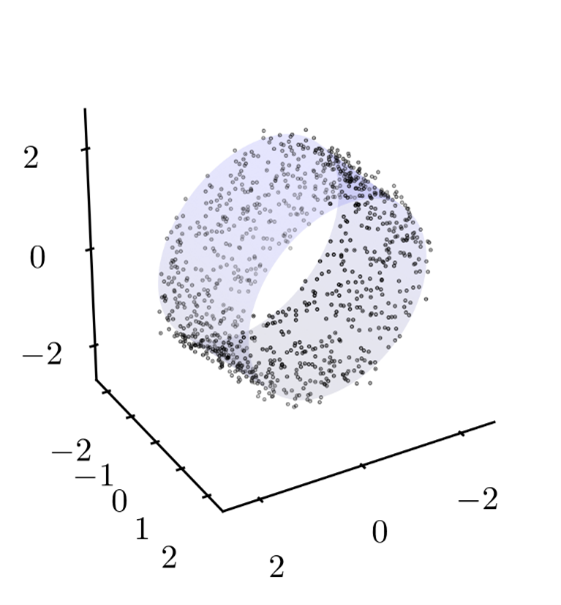}
    \end{subfigure}
    \caption{Data points in $\mathbb{R}^3$. Left: they exactly lie on a 2-dimensional manifold (a cylinder); Right: they lie close to it.}
    \label{fig:mh}
\end{figure}

Scientists and engineers are increasingly training and testing models on high-dimensional datasets \citep{zhou2021machine}. Yet, the success of modern machine learning algorithms appears to defy the curse of dimensionality, which posits that the sample size required to reliably recover underlying structures grows exponentially with the data dimension \citep{keogh2011curse}. One possible explanation is the \emph{manifold hypothesis}, which assumes that despite sitting within a high-dimensional space $\mathbb{R}^p$, the data actually lie on or near a lower $d$-dimensional manifold \citep{aghajanyan2020intrinsic, meilua2024manifold}, as illustrated in Figure~\ref{fig:mh}. Under this assumption, at least locally or approximately, the dataset can be described using only $d<p$ free parameters, thus greatly simplifying the problem. 

A natural task that arises from the manifold hypothesis is the \emph{manifold learning problem}, which aims to recover the underlying manifold from the dataset. This task is also referred to as \emph{nonlinear dimension reduction}; see \citet{balasubramanian2002isomap} and \citet{coifman2006diffusion} for classical methods, and \citet{floryan2022data} and \citet{alberti2024manifold} for more recent, data-driven approaches. The most important output of manifold learning is often a low-dimensional representation of the data. This reduced representation is particularly useful because many classical statistical methods do not scale well in high dimensions and can be more effectively applied to the lower-dimensional coordinates. While the integration of manifold learning with traditional statistical analysis remains relatively underexplored, existing work highlight its potential, such as regression on manifolds \citep{aswani2011regression, cheng2013local} and factor models for time series \citep{chan2017factor}. As high-dimensional data continues to challenge standard statistical techniques, combining manifold learning with classical methods offers a promising yet underutilized alternative.

Among the many properties of the underlying manifold, its dimension is arguably the most important, yet it is often assumed to be known and specified in advance by manifold learning algorithms \citep{meilua2024manifold}. As a result, researchers have long been interested in the subproblem of \emph{manifold dimension estimation}, and numerous methods have been proposed. But existing work is often fragmented, with researchers approaching the problem from domain-specific perspectives with no unified treatment. Moreover, only limited systematic evaluation of existing techniques have been conducted (such as \citet{campadelli2015intrinsic} and \citet{camastra2016intrinsic}, both being now somewhat outdated and limited in scope\footnote{During the final stage of writing this article, we became aware of the arXiv preprint by \citet{binnie2025survey}, which also surveys dimension estimators. However, our work was conducted independently and approaches the problem from different angles, offering unique insights and experimental rigor.}), making it difficult for practitioners to make informed choices. Moreover, much of the existing literature on manifold dimension estimation is not developed from a statistician’s perspective, and the role of statisticians in shaping this area has so far remained limited.

This article presents a comprehensive and theoretically grounded study of the manifold dimension estimation problem, with statistically trained readers in mind. We review often overlooked theoretical aspects of the problem; provide a concise survey of representative manifold dimension estimators organized within a novel design-principle-based framework; and, through a proposed hyperparameter-tuning algorithm, conduct fair and systematic benchmarking on a wide range of synthetic and real-world datasets. These experiments also yield insights into how estimator performance varies with key factors such as neighborhood size, curvature, noise, and sample size, which are analyzed alongside the empirical results. Our goals are to provide practical guidance for estimator selection, to engage statisticians with this problem, and to extract insights from existing methods that may inform future methodological development.

The remainder of the paper is organized as follows. Section~\ref{sec:TheoreticalBackground} introduces the core theoretical foundations, including manifold geometry, sampling procedures, and the statistical difficulty of the problem, providing context that is often omitted in empirical studies. Section~\ref{sec:manifold_dimension_estimators} presents a brief survey of existing manifold dimension estimation methods, focusing on the mathematical construction of eight representative estimators that are subsequently evaluated. Section~\ref{sec-evaluation} provides a systematic empirical evaluation, benchmarking estimator performance under controlled variations in neighborhood size, curvature, noise level, and sample size. These experiments motivate a principled approach to dataset-specific hyperparameter tuning and enable fair comparisons across a diverse set of synthetic and real-world datasets. Section~\ref{sec-conclusion} concludes with practical recommendations and directions for future research. 

Experiments are conducted using both \texttt{Python} and \texttt{R}, with code and datasets publicly available; all numerical results are tabulated in the Appendix.

\section{Theoretical Background}\label{sec:TheoreticalBackground}

In this section, we provide a concise yet comprehensive overview of the theoretical foundations relevant to manifold dimension estimation. We begin with a review of manifolds and their associated structures, then discuss the problem of sampling from manifolds. Finally, we formulate the manifold dimension estimation problem and outline its statistical challenges.

Manifolds, and the mathematical constructions built upon them, can become quite abstract when developed in full generality. Such abstraction, however, is unnecessary for our purposes. Throughout, we adopt the most direct and practical definitions, while avoiding oversimplification. For readers interested in more advanced treatments, we refer to standard references in differential geometry, such as \citet{lee2003smooth} and \citet{lee2018introduction}.

\subsection{Manifold and Dimension}\label{sec:manifold_dimension}

\begin{figure}[H]
    \centering
    \begin{minipage}[t]{0.24\textwidth}
        \centering
        \includegraphics[width=\linewidth]{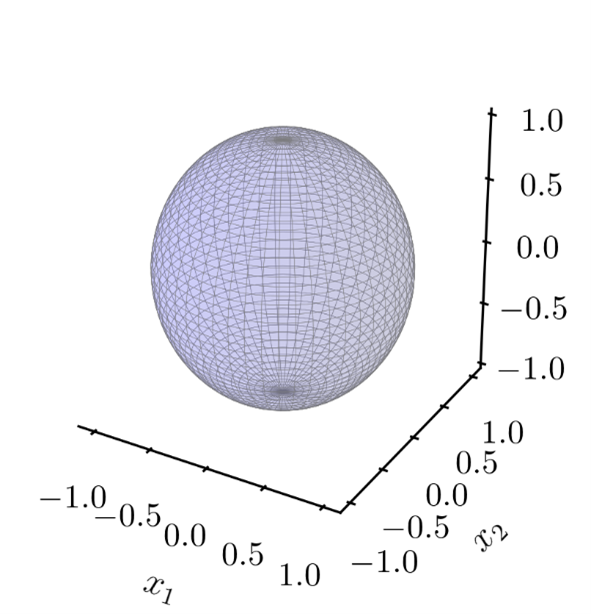}
        \caption*{(a)}
    \end{minipage}
    \begin{minipage}[t]{0.24\textwidth}
        \centering
        \includegraphics[width=\linewidth]{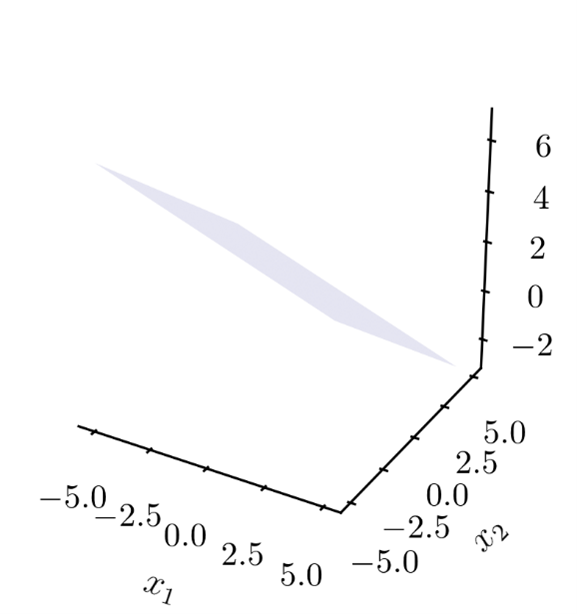}
        \caption*{(b)}
    \end{minipage}
    \begin{minipage}[t]{0.24\textwidth}
        \centering
        \includegraphics[width=\linewidth]{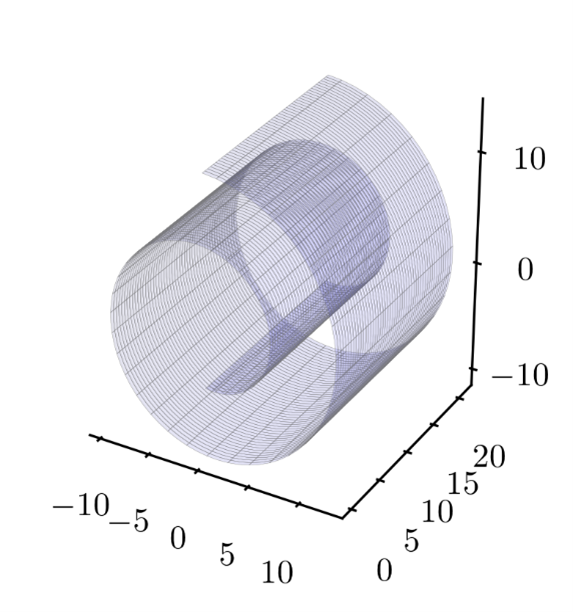}
        \caption*{(c)}
    \end{minipage}
    \begin{minipage}[t]{0.24\textwidth}
        \centering
        \includegraphics[width=\linewidth]{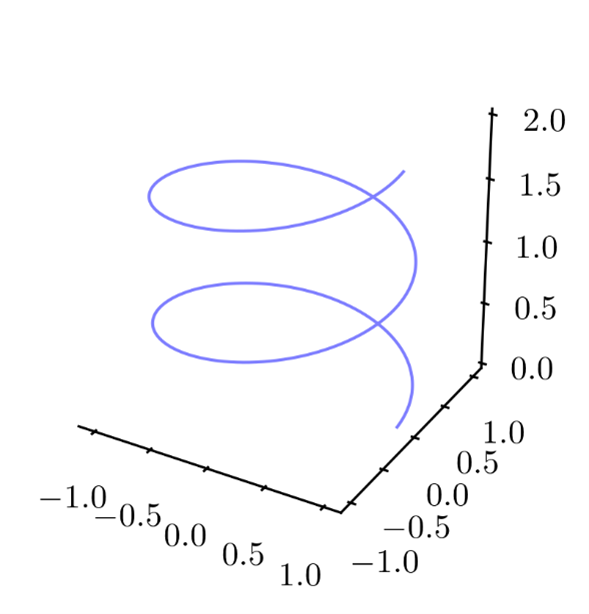}
        \caption*{(d)}
    \end{minipage}

    \vspace{0.5em}

    \begin{minipage}[t]{0.24\textwidth}
        \centering
        \includegraphics[width=\linewidth]{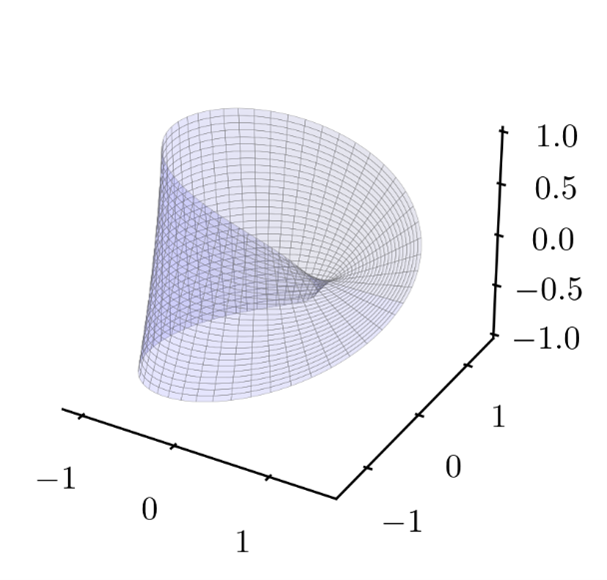}
        \caption*{(e)}
    \end{minipage}
    \begin{minipage}[t]{0.24\textwidth}
        \centering
        \includegraphics[width=\linewidth]{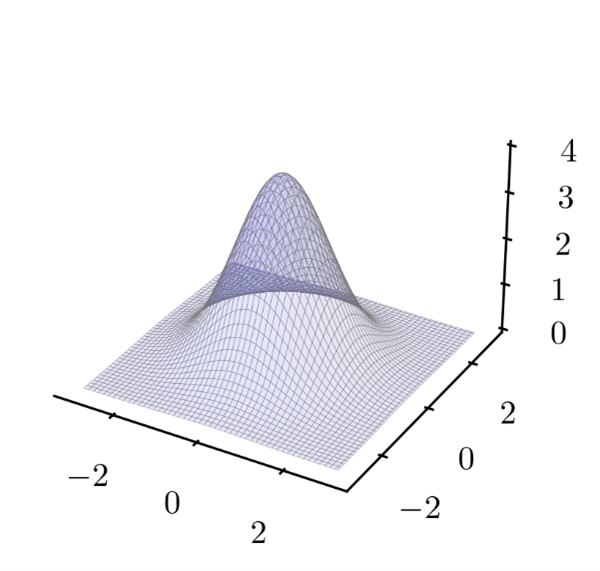}
        \caption*{(f)}
    \end{minipage}
    \begin{minipage}[t]{0.24\textwidth}
        \centering
        \includegraphics[width=\linewidth]{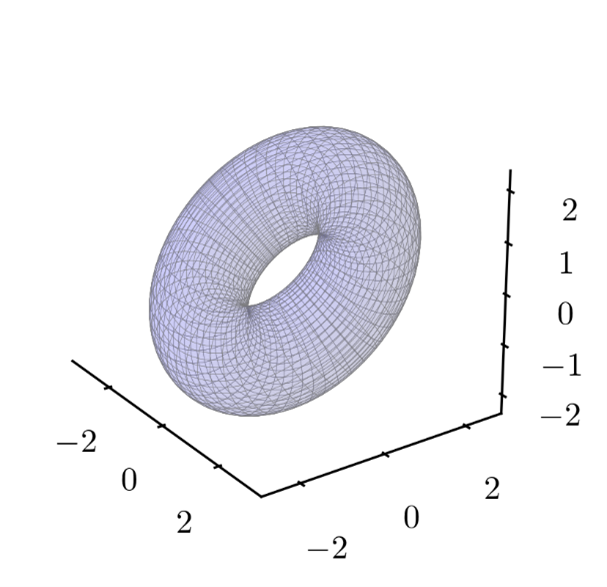}
        \caption*{(g)}
    \end{minipage}
    \begin{minipage}[t]{0.24\textwidth}
        \centering
        \includegraphics[width=\linewidth]{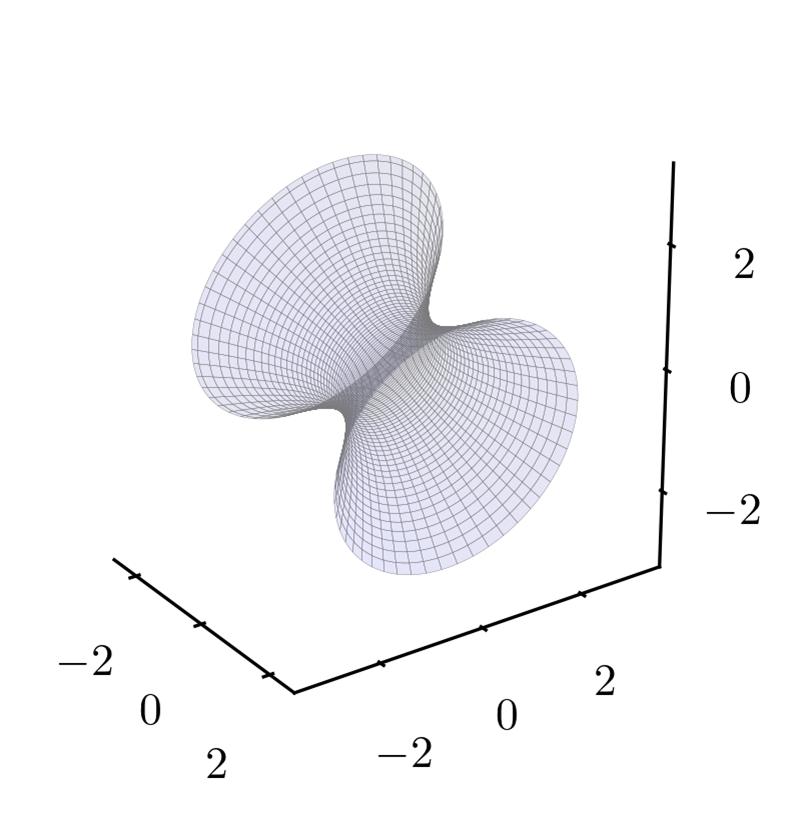}
        \caption*{(h)}
    \end{minipage}

    \caption{Common manifolds embedded in $\R^3$: (a) a $2$-sphere; (b) a hyperplane; (c) a Swiss roll with no boundary; (d) a helix with no endpoints; (e) a Möbius strip with no boundary; (f) a Gaussian density surface; (g) a torus; (h) a hyperboloid of one sheet. }
    \label{fig:manifolds}
\end{figure}

In this article, we take a $d$-dimensional manifold $M$ as a smooth object of dimension $d$, embedded in a higher-dimensional \emph{ambient space} $\mathbb{R}^p$, thereby generalizing the notion of smooth curves and surfaces to arbitrary dimension $d$. Figure~\ref{fig:manifolds} illustrates several classical examples.

\begin{figure}[H]
    \centering
    \begin{subfigure}[b]{0.35\textwidth}
        \centering
        \includegraphics[width=\textwidth]{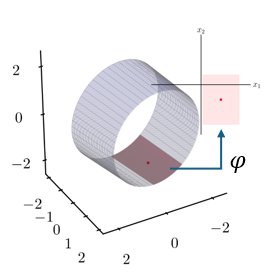}
        \caption{Chart at one point $\bm{x}_0$.}
        \label{fig:chart}
    \end{subfigure}
    \hfill
    \begin{subfigure}[b]{0.35\textwidth}
        \centering
        \includegraphics[width=\textwidth]{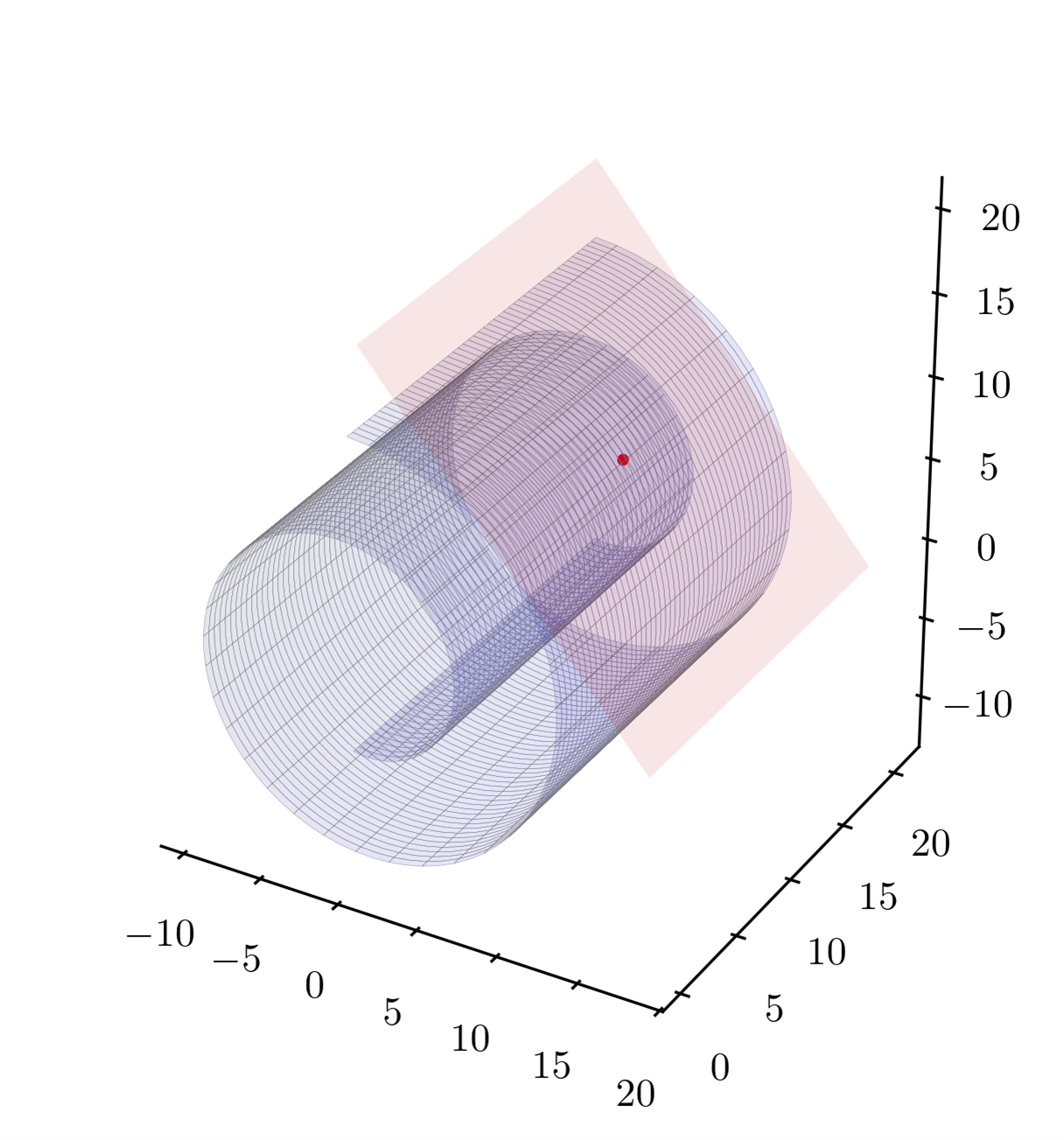}
        \caption{Tangent space at one point $\bm{x}_0$.}
        \label{fig:tangent}
    \end{subfigure}
    \caption{Chart and tangent space.}
    \label{fig:charttangent}
\end{figure}

Real-world datasets do not necessarily lie on any of the classical manifolds illustrated in Figure~\ref{fig:manifolds}. The strength of the manifold model lies in its ability to describe far more general geometric structures beyond these familiar examples. What distinguishes a manifold $M$ from an arbitrary subset of $\mathbb{R}^p$ is that it possesses certain local (and global) properties. The most important of these is that $M$ is \emph{locally Euclidean}: every point $\bm{x}_0 \in M$ has a neighborhood $U \subseteq M$ that can be described using only $d$ coordinates, as illustrated in Figure~\ref{fig:chart}. Consequently, although $M$ is embedded in the higher-dimensional space $\mathbb{R}^p$, it locally resembles $\mathbb{R}^d$ and can be parameterized by only $d$ free variables. We refer to this integer $d$ as the \emph{intrinsic dimension} of $M$, and to $p$ as its \emph{ambient dimension}. In the context of manifold dimension estimation, $p$ is known, while $d$ has to be inferred from the sample.

Beyond the commonly adopted notion of intrinsic dimension $d$ introduced above, several alternative definitions of dimension exist. Some manifold dimension estimators are based on these alternative notions (e.g., \citet{camastra2002estimating}, \citet{bruske2002intrinsic}, \citet{bac2020local}) so we very briefly introduce them below. Notable examples include:

\begin{enumerate}
    \item \emph{Topological dimension} \citep{fedorchuk1990fundamentals}, also known as the \emph{Lebesgue covering dimension}, is defined for general topological spaces and remains invariant under homeomorphisms.
    
    \item \emph{Box-counting dimension} \citep{falconer2013fractal}, also called the \emph{Minkowski dimension}, is obtained by analyzing the asymptotic behavior of the number of boxes (sets with equal-length interval sides) required to cover a set. It is widely used in the study of fractals.
    
    \item \emph{Hausdorff dimension} \citep{falconer2013fractal} is defined in terms of the Hausdorff measure. It applies to metric spaces and plays a central role in fractal geometry as well.
\end{enumerate}

These definitions arise from different contexts and motivations and are not tailored specifically to manifolds; however, for sufficiently regular spaces such as manifolds, the resulting notions of dimension coincide.

\subsection{Tangent Space and Curvature}

As discussed in the previous section, a manifold $M$ is locally Euclidean. Consequently, in a neighborhood of any point $\bm{x}_0 \in M$, it can be well approximated by a flat $d$-dimensional space (e.g., a plane when $d=2$) that ``touches'' the manifold at $\bm{x}_0$. This space is called the \emph{tangent space} at $\bm{x}_0$, denoted by $T_{\bm{x}_0}M$, as illustrated in Figure~\ref{fig:tangent}.

\begin{figure}[H]
    \centering
    \begin{subfigure}[b]{0.40\textwidth}
        \centering
        \includegraphics[width=\textwidth]{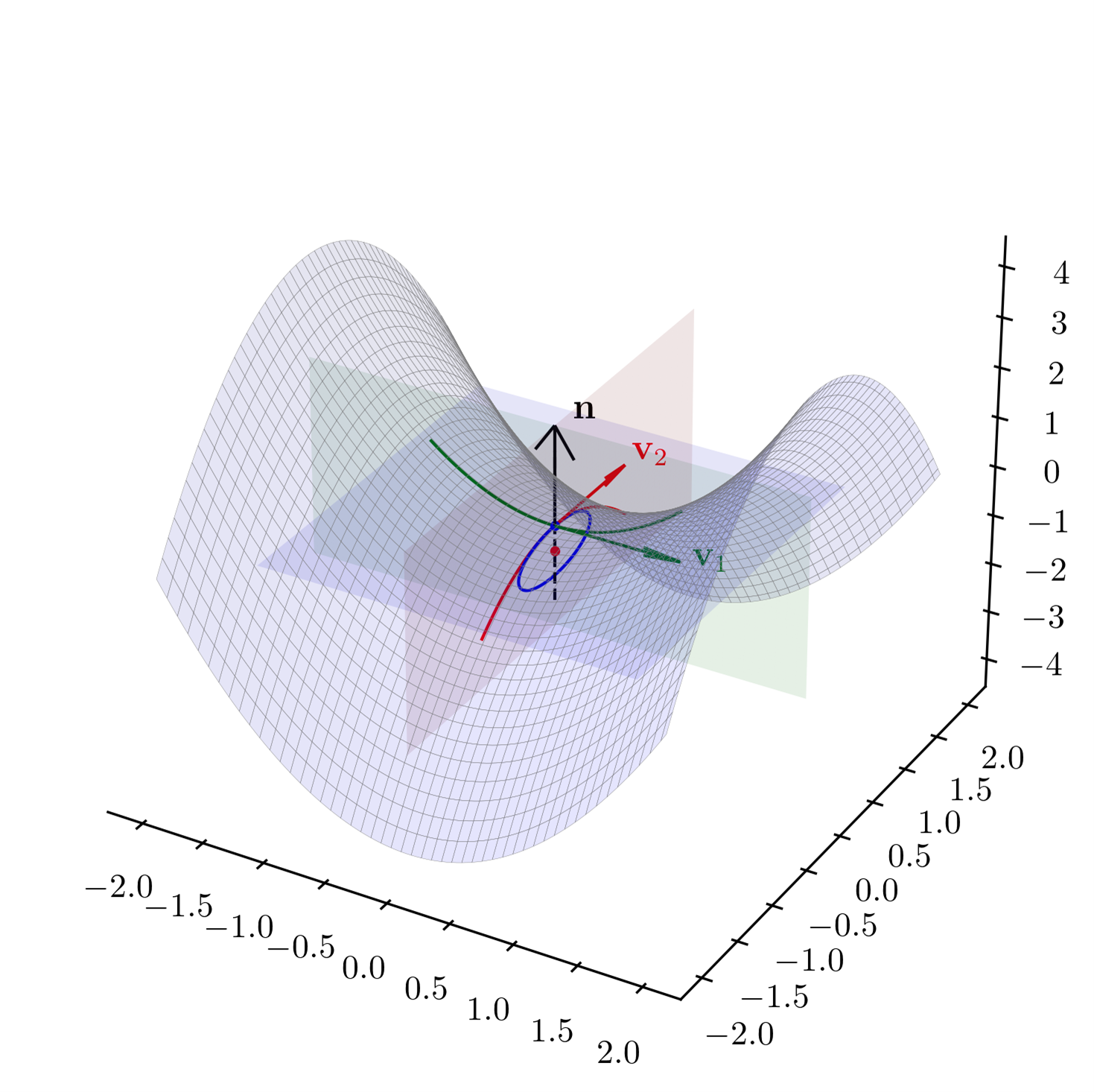}
        \caption{Principal curves (in dark green and red) and osculating circle whose inverse radius gives the  curvature of the red curve at $\bm{x}_0$.}
        \label{fig:principal}
    \end{subfigure}
    \hfill
    \begin{subfigure}[b]{0.40\textwidth}
        \centering
        \includegraphics[width=\textwidth]{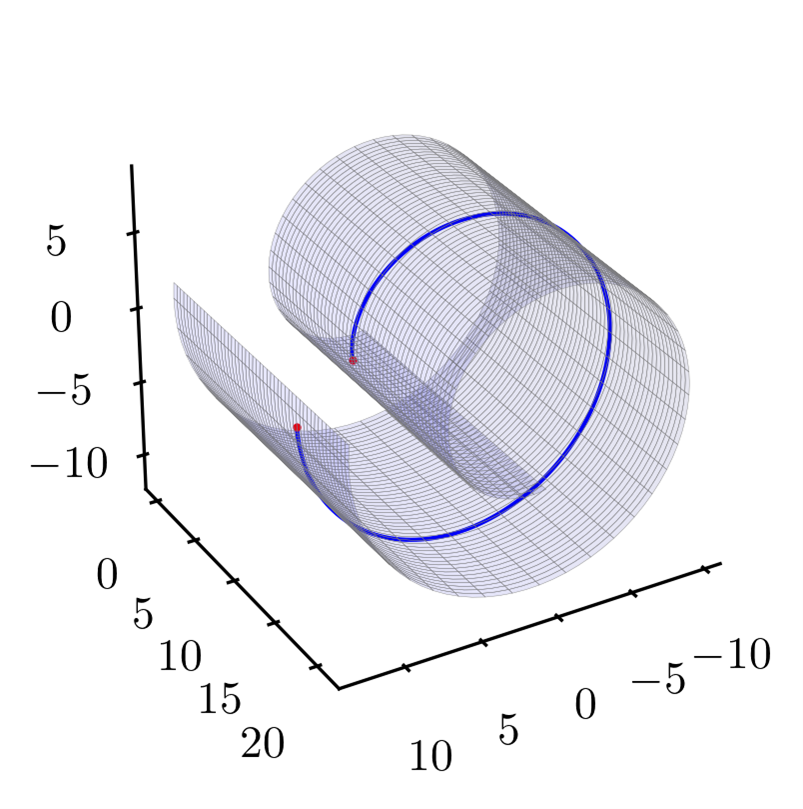}
        \caption{Geodesic (i.e., curve of minimal length) in blue between two points (in red) on a manifold.}
        \label{fig:distance}
    \end{subfigure}
    \caption{Principal curves and their curvature (left) and a geodesic of minimal distance along a manifold (right).}
    \label{fig:principaldistance}
\end{figure}

Formally, let $\varphi : U \to \varphi(U) \subseteq \mathbb{R}^d$ be a coordinate map defined on a neighborhood $U$ of $\bm{x}_0 \in M$. The Jacobian matrix of $\varphi^{-1}$ evaluated at $\varphi(\bm{x}_0) \in \mathbb{R}^d$,
\[
\mathrm{d}\varphi^{-1}\big|_{\varphi(\bm{x}_0)}
\;:=\;
\left[\frac{\partial \varphi^{-1}_i}{\partial u_j}\right]_{\varphi(\bm{x}_0)}
\in \mathbb{R}^{p \times d},
\]
has full rank $d$ and therefore defines an injective linear map from $\mathbb{R}^d$ into $\mathbb{R}^p$. The tangent space to $M$ at $\bm{x}_0$ is then given by
\[
T_{\bm{x}_0}M
\;=\;
\left\{\, \bm{x}_0 + \mathrm{d}\varphi^{-1}\big|_{\varphi(\bm{x}_0)} \bm{u}
: \bm{u} \in \mathbb{R}^d \,\right\}, 
\]
which is a vector space with its operations defined with respect to $\bm{x}_0$. It provides the \emph{best linear approximation} to the manifold in a neighborhood of $\bm{x}_0$ as the definition corresponds to the first-order Taylor expansion of $\varphi^{-1}$ at $\varphi(\bm{x}_0)$. The existence of tangent space is crucial in practice, as many manifold dimension estimators rely on approximating small neighborhoods of $M$ by their tangent spaces, thereby reducing a nonlinear geometric problem to a local linear one.

For trivial manifolds like \(\mathbb{R}^p\) or its subspaces, the tangent space at any point coincides exactly with the manifold itself; i.e., the linear approximation is perfect everywhere and such manifolds are said to be \emph{flat}. More general manifolds are not flat and can curve nontrivially, the farther we move away from $\bm{x}_0$, the less accurate the approximation becomes. For curved manifolds, \emph{principal curvatures} \citep{lee2018introduction} can be used to quantify how a manifold bends at various points. The basic idea is illustrated in Figure~\ref{fig:principal}: each direction (e.g., $\bm{v}_1$ or $\bm{v}_2$) in the tangent space $T_{\bm{x}_0}M$ (in light blue) can be combined with a normal direction (e.g., $\bm{n}$) to form a plane (in light green or light red) whose intersection with the manifold determines a curve (in dark green or dark red). The principal curvature is then defined as the amount by which this curve bends at $\bm{x}_0$.\footnote{The magnitude of the principal curvature is defined as the inverse radius of the osculating circle, which is the unique circle that, when parametrized with unit speed, shares the same position, velocity, and acceleration as the curve at $\bm{x}_0$.} Since there are $d$ linearly independent directions in the tangent space and $p - d$ normal directions, a total of $d(p - d)$ principal curvatures are needed to fully describe the local geometry at $\bm{x}_0$.

\subsection{Geodesics and Distance on a Manifold}

By construction, a manifold is also a metric space\footnote{A smooth manifold becomes a metric space once endowed with a Riemannian metric. In the embedded case, the induced metric.}. Distances \emph{along the manifold} between two points are measured via curves (of minimal length) that sit on the manifold. These curves are  called \emph{geodesics}. They generalize the notion of straight lines in Euclidean spaces. It is important to note that the intrinsic distance between two points on a manifold can differ substantially from their Euclidean distance in the ambient space $\mathbb{R}^p$; points that appear close in $\mathbb{R}^p$ can be far apart along the manifold, as illustrated in Figure~\ref{fig:distance}.

\subsection{Local Graph Representation}

Another way to characterize a manifold $M \subseteq \mathbb{R}^p$ is through \emph{local graph representation}, which expresses $M$ locally as a collection of function graphs. For example, the 2-dimensional unit sphere in $\mathbb{R}^3$ is defined as the set of triples $(x_1,x_2,x_3)$ satisfying $x_1^2 + x_2^2 + x_3^2 = 1$. In a small neighborhood (say, on the upper hemisphere), this relation can then be used to solved for $x_3$, giving
\[
x_3 = g(x_1,x_2) := \sqrt{1 - x_1^2 - x_2^2}.
\]
Thus, in this neighborhood the sphere coincides with the graph of the function $g$.

This idea extends to any $d$-dimensional manifold $M$. By Proposition 5.16 of \citet{lee2003smooth}, together with the inverse function theorem, for any point $\bm{x}_0 \in M$ there exists a neighborhood $U \subseteq M$ in which each $\bm{x} \in U$ can be expressed in the new coordinates satisfying
\[
\bm{x}' = (x'_1, \ldots, x'_d,\ g(x'_1, \ldots, x'_d)),
\]
with $g$ a smooth function satisfying $g(\bm{0}) = \bm{0}$ and $\nabla g(\bm{0}) = \bm{0}$. In other words, $U$ is precisely the graph of $g$. The first $d$ new coordinates $(x_1',\ldots,x_d')$ correspond to directions in the tangent space $T_{\bm{x}_0}M$, while the remaining $p - d$ new coordinates are fully determined by $g$. Setting these last $p - d$ coordinates to zero corresponds to the first-order Taylor approximation to $g$, which is equivalent to the tangent space approximation for $M$ at $\bm{x}_0$. The curvature of the manifold at $\bm{x}_0$ is fully encoded in the higher-order terms of the Taylor expansion of $g$.

\subsection{Embedding}

The major justification for the manifold hypothesis is the celebrated Nash embedding theorem \citep{nash1956imbedding}, which guarantees that any manifold can be embedded into some Euclidean space in a smooth and faithful way, preserving angles and distances. This naturally raises the question of how small the ambient dimension can be while still faithfully representing the manifold.

For instance, all the 2-dimensional manifolds shown in Figure~\ref{fig:manifolds} of Section~\ref{sec:manifold_dimension} are embedded in $\mathbb{R}^3$, where the ambient and intrinsic dimensions are close. We can easily embed them into higher dimensional spaces through linear embeddings with their geometries preserved, but as long as the manifolds remain in some $3$-dimensional subspace, the manifold dimension estimation problem on them should remain relatively easy.

\begin{figure}[H]
   \centering
   \begin{subfigure}[b]{0.30\textwidth}
       \centering
       \includegraphics[width=\textwidth]{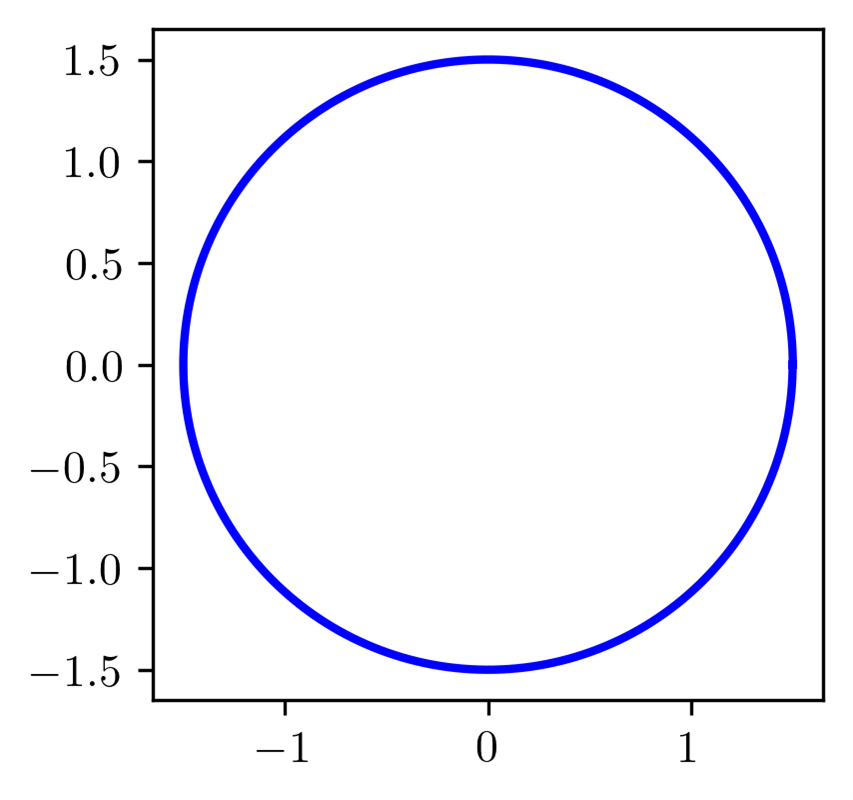}
       \caption{$c = 0.01$}
       \label{fig:c1}
   \end{subfigure}
   \hfill
   \begin{subfigure}[b]{0.30\textwidth}
       \centering
       \includegraphics[width=\textwidth]{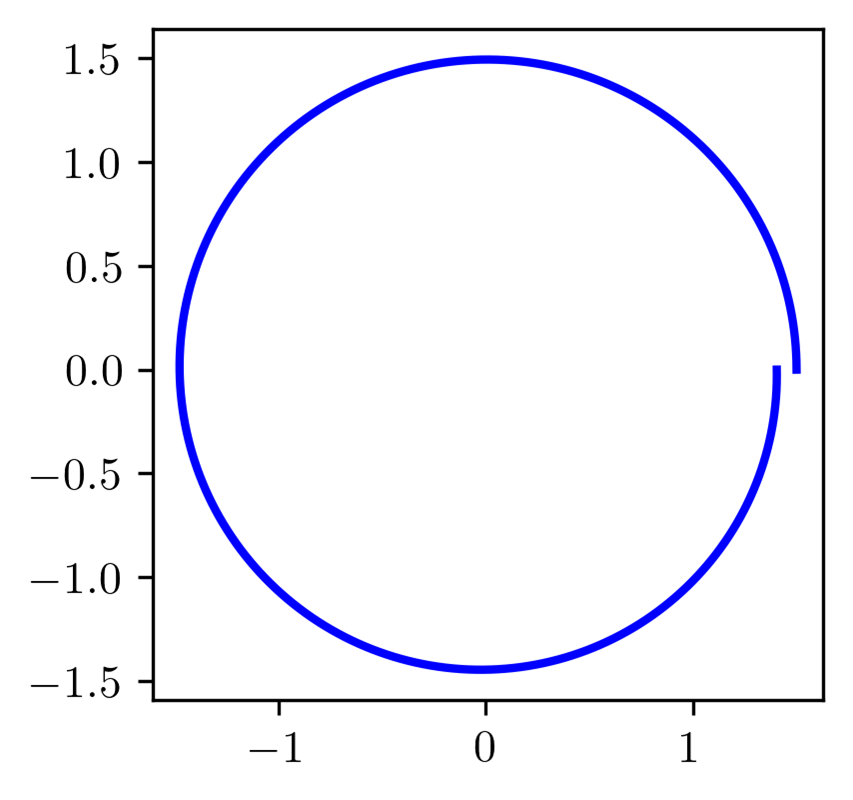}
       \caption{$c = 0.1$}
       \label{fig:c2}
   \end{subfigure}
   \hfill
   \begin{subfigure}[b]{0.30\textwidth}
       \centering
       \includegraphics[width=\textwidth]{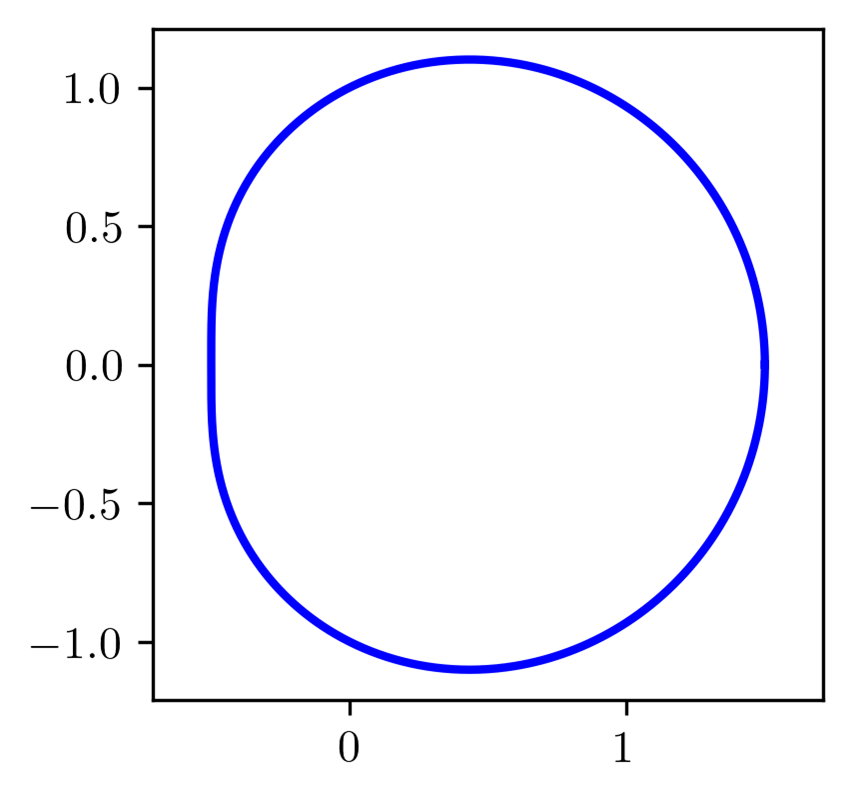}
       \caption{$c=1$}
       \label{fig:c3}
   \end{subfigure}
   \caption{Three deformed spheres ($d=1$).}
   \label{fig:deformed}
\end{figure}

A more challenging scenario is when a $d$-dimensional manifold is nontrivially embedded into~$\R^p$. For example, consider the $d$-dimensional \emph{deformed sphere}, defined through an inverse chart function $\varphi^{-1}: (-\pi,\pi)^d \to \R^p$ with $p = 2d$ as\footnote{Strictly speaking, $\varphi^{-1}$ may fail to be injective for certain values of $c$. In such cases, the manifold should be understood as the image of $\varphi^{-1}$ with the singular points removed.}
\[
\varphi^{-1}(\bm{u}) = (x_1, \ldots, x_{2d}),
\]
with
\[
x_j     = [R + r \cos(2c\pi u_j)] \cos(2\pi u_j) \text{ and }
x_{j+d} = [R + r \cos(2c\pi u_j)] \sin(2\pi u_j)
\]
for $j = 1, \cdots, d$ and $R, r, c > 0$ fixed constants. By definition, a $d$-dimensional deformed sphere resides in $\R^{2d}$ and is not contained in any lower-dimensional subspaces\footnote{This is rather evident from the definition; see Appendix~\ref{sec:mathclaim} for a formal justification.}. Figure~\ref{fig:deformed} shows the deformed sphere when $d = 1$ for  different values of $c$.
The deformed sphere is special in the sense that all its $d^2$ principal curvatures are nontrivial. Manifolds like the deformed sphere are said to be space-filling or nonlinearly embedded in some literature \citep{campadelli2015intrinsic}.


\subsection{Probability Distributions on Manifolds}

Let $(\Omega,\mathcal{F},\mathbb{P})$ be a probability space and let $\bm{X}$ be a random vector defined on it, taking values from $\R^p$. The manifold hypothesis can be restated by saying that the data distribution is supported entirely on a manifold $M$, i.e., $\mathbb{P}(\bm{X} \in M) = 1$. Moreover, it is often convenient to assume that this distribution admits a density function $f_{\bm{X}}$ supported on $M$.

A few technical challenges must be addressed for a rigorous formulation. The most important is that when $d<p$, the standard Lebesgue measure on $\mathbb{R}^p$ is not appropriate for defining the ‘size’ of the manifold, and one must instead use measures adapted to lower-dimensional sets.

It turns out that the appropriate measure here is the $d$-dimensional Hausdorff measure, denoted by $\nu$; see \citet{diaconis2013sampling} for further details. 
Assume that the probability distribution $P_{\bm{X}}$ on the manifold admits a density function $f_{\bm{X}}$ with respect to $\nu$. 
Given a chart function $\varphi$ with domain $U$, one can show that
$$
\int_{\varphi(U)} f_{\bm{X}}(\varphi^{-1}(\bm{u}))\, J_d\varphi^{-1}(\bm{u})\, d\bm{u}
= \int_{U} f_{\bm{X}} \, d\nu,
$$
where
$$
J_d\varphi^{-1}(\bm{u}) \;:=\; \sqrt{\det\!\left(\mathrm{d}\varphi^{-1}(\bm{u})^\top \mathrm{d}\varphi^{-1}(\bm{u})\right)}.
$$
Here $J_d\varphi^{-1}(\bm{u})$ plays the role of the \emph{volume distortion factor}: 
it generalizes the Jacobian determinant from the classical change-of-variables formula, 
adjusting for how the map $\varphi^{-1}$ stretches or compresses $d$-dimensional volumes in the ambient space. 

The above result implies that sampling points from $\varphi(U) \subseteq \mathbb{R}^d$ according to the density 
$ f_{\bm{X}}(\varphi^{-1}(\bm{u}))\, J_d\varphi^{-1}(\bm{u})$, and then mapping them back to $M$ via $\varphi^{-1}$, 
is equivalent in distribution to sampling directly from $M$ with density $f_{\bm{X}}$ relative to $\nu$.

The above result, together with the fact that any compact manifold can be covered by finitely many disjoint neighborhoods (up to a set of probability zero), leads to the rejection sampling algorithm below for generating random samples from a prescribed distribution on $M$. 
It is worth noting, however, that more efficient sampling methods are available for specific cases. 
For example, to sample uniformly from the $d$-dimensional unit sphere in $\mathbb{R}^{d+1}$, one can draw a vector from $N(0,I_{d+1})$ and normalize it \citep{marsaglia1972choosing}.

\begin{algorithm}[H]
\caption{Sampling from a density function $f_{\bm{X}}$ supported on a manifold $M$}
\label{alg:sampling}
\begin{algorithmic}[1]
\State \textbf{Input:} 
\begin{itemize}
  \item $f_{\bm{X}}$: a density function supported on $M$
  \item $\{(U_\alpha, \varphi_\alpha, p_\alpha)\}_{1 \leq \alpha \leq a}$: a partition of $M$ into charts, with chart functions $\varphi_\alpha$, and region probabilities $p_\alpha$, with $\alpha,a\in\mathbb{N}$
  \item $n$: sample size
\end{itemize}
\State \textbf{Output:} a random sample of size $n$ from $f_{\bm{X}}$
\State For each $\alpha$, find a constant $M_\alpha$ such that for all $\bm{u} \in \varphi_\alpha(U_\alpha)$,
\[
f_{\bm{X}}(\varphi_\alpha^{-1}(\bm{u})) \cdot J_d \varphi_\alpha^{-1}(\bm{u}) \leq M_\alpha.
\]
\State Initialize $k = 0$
\While{$k < n$}
    \State Sample $\alpha \in \{1, \ldots, a\}$ according to the distribution $\{p_\alpha\}$
    \Repeat
        \State Sample $\bm{u}$ uniformly from $\varphi_\alpha(U_\alpha)$
        \State Compute acceptance probability:
        \[
        p_{\mathrm{accept}} = \frac{f_{\bm{X}}(\varphi_\alpha^{-1}(\bm{u})) \cdot J_d \varphi_\alpha^{-1}(\bm{u})}{M_\alpha}
        \]
        \State Sample $r \sim \mathrm{Uniform}[0,1]$
    \Until{$r < p_{\mathrm{accept}}$}
    \State Set $\bm{x}_{k+1} = \varphi_\alpha^{-1}(\bm{u})$
    \State $k \gets k + 1$
\EndWhile
\State \Return $\bm{x}_1, \dots, \bm{x}_n$
\end{algorithmic}
\end{algorithm}

\subsection{Manifold Dimension Estimation}

With all the constructions at our disposal, we state the manifold dimension estimation problem as below.

\emph{Let $\bm{x}_{1:n} := \{\bm{x}_k\}_{k=1}^n \subseteq \mathbb{R}^p$ be an observed random sample from a probability distribution with density $f_{\bm{X}}$ supported on a $d$-dimensional manifold $M$. 
The problem of manifold dimension estimation is to estimate the intrinsic dimension $d$ by means of an estimator $\hat{d}(\bm{x}_{1:n})$.}

\citet{kim2019minimax} analyzed the manifold dimension estimation problem by characterizing its fundamental difficulty in terms of minimax risk. 
Specifically, consider a random sample of size $n$ in $\mathbb{R}^p$, drawn from a probability distribution $\mathbb{P} \in \mathcal{P}$, 
where $\mathcal{P}$ is a class of distributions with bounded densities supported on well-behaved manifolds of intrinsic dimension $d \leq p$. 
Let $\hat{d}_n$ be an estimator constructed from the sample, and let $d(\mathbb{P})$ denote the intrinsic dimension of the manifold on which $\mathbb{P}$ is supported. 
The minimax risk is defined as
\[
R_n \;=\; \inf_{\hat{d}_n} \sup_{\mathbb{P} \in \mathcal{P}} \mathbb{E}_\mathbb{P}\!\left[ \mathbf{1}_{\{\hat{d}_n \neq d(\mathbb{P})\}} \right],
\]
which represents the smallest possible worst-case probability of misestimating the intrinsic dimension. 

The authors show that when the manifolds under consideration are compact, complete, and have bounded curvature, the minimax risk satisfies
\[
- c_1 n \;\leq\; \log R_n \;\leq\; - c_2 n,
\]
for some constants $c_1, c_2 > 0$. 
This exponential decay implies that the probability of error decreases at a rate proportional to $e^{-cn}$, 
so the dimension estimation problem becomes exponentially easier as the sample size increases.

\section{Manifold Dimension Estimators}\label{sec:manifold_dimension_estimators}

Over the years, researchers from diverse fields have proposed a wide range of manifold dimension estimators; see, for example, \citet{verveer1995evaluation}, \citet{kegl2002intrinsic}, \citet{yang2007conical}, \citet{carter2009local}, \citet{sricharan2010optimized}, \citet{kalantana2013intrinsic}, \citet{liao2014intrinsic}, \citet{facco2017estimating}, \citet{denti2022generalized} and \citet{lim2024tangent}. Because these estimators are developed from heterogeneous domain-specific perspectives and assumptions, there is no widely accepted unified framework for their classification.

One common approach in existing surveys is to categorize methods as either \textit{local} or \textit{global}, where local methods produce dimension estimates within neighborhoods of the data, while global methods yield a single estimate for the entire dataset. This dichotomy, however, is not entirely satisfactory: local estimates can be readily aggregated to form a global one, and any global method can be applied to subsets of the data, thereby becoming local in a trivial sense.

In this section, which serves as a brief survey, we classify manifold dimension estimators according to their underlying design principles, distinguishing between \textit{flatness-based estimators} and other approaches. Broadly speaking, the majority of existing manifold dimension estimators fall into the class of flatness-based estimators. These methods rely on the assumption that the manifold $M$ exhibits negligible curvature and can therefore be well approximated as locally flat within sufficiently small neighborhoods. As discussed in Section~\ref{sec:TheoreticalBackground}, for any $\bm{x}_k \in M$, the tangent space $T_{\bm{x}_k}M$ provides the best local linear approximation to the manifold, with deviations arising from curvature effects.

Flatness-based estimators typically assume that $\bm{x}_k$ together with its $K$ nearest neighbors are uniformly distributed within a $d$-dimensional ball of radius $R>0$, denoted $B_{\bm{x}_k}(R)$, embedded in the tangent space and centered at $\bm{x}_k$. Under this flatness assumption, statistics computed from $\bm{x}_k$ and its neighbors can be expressed as functions of the intrinsic dimension $d$, thereby motivating a wide range of dimension estimation procedures.

Given the sheer number and diversity of existing manifold dimension estimators, it is neither feasible nor our intention to provide an exhaustive review here. Instead, we focus on eight representative estimators included in our empirical study. These methods span a broad range of design principles, both within and beyond the flatness assumption, ranging from top-performing approaches identified in previous surveys \citep{campadelli2015intrinsic, bac2021scikit} to more recent and promising methods that have not yet been extensively evaluated. 

Several of the selected estimators are classical in nature or give rise to well-established methodological families, and many other estimators can be viewed as variants of these approaches; such variants will be briefly mentioned where appropriate. Along the way, we also discuss key practical considerations, such as computational cost.
 
Throughout this section, let $\{\bm{x}_k\}_{k=1}^{n} \subseteq \mathbb{R}^p$ denote an observed sample drawn from a density function $f:=f_{\bm{X}}$ supported on an underlying $d$-dimensional manifold $M$. 
For each observation~$\bm{x}_k$, we denote $(\bm{x}_k^1, \ldots, \bm{x}_k^K)$ the ordered (from closest to furthest) $K$-tuple of its nearest neighbors in Euclidean distance. We denote $\norm{\bm{x}}$ the Euclidean norm of a vector $\bm{x}\in\mathbb{R}^p$. 

\subsection{Flatness-based Estimators}

\subsubsection{The Local PCA estimator}

\textsc{Local PCA} \citep{fukunaga1971algorithm, fan2010intrinsic} applies principal component analysis (PCA) to each local neighborhood.  
For the neighborhood of $\bm{x}_k$, let $\hat{\lambda}_1 \geq \cdots \geq \hat{\lambda}_p$ denote the eigenvalues of the sample covariance matrix
\[
\hat{\Sigma}_k \;=\; \frac{1}{K}\sum_{\ell=0}^K (\bm{x}_k^\ell - \bar{\bm{x}}_k)(\bm{x}_k^\ell - \bar{\bm{x}}_k)^\top,
\qquad 
\text{where } \bm{x}_k^0 = \bm{x}_k \text{ and } 
\bar{\bm{x}}_k = \frac{1}{K+1}\sum_{\ell=0}^K \bm{x}_k^\ell.
\]
Under the flatness assumption, the smallest $p-d$ eigenvalues of the population covariance matrix $\Sigma_k$ vanish. 
Accordingly, we expect the empirical covariance $\hat{\Sigma}_k$ to approximate this structure, with $\hat{\lambda}_{d+1}, \ldots, \hat{\lambda}_p$ close to zero. 
In practice, each neighborhood yields an estimate $\hat{d}_k$ (expected to be close to $d$), defined as the number of eigenvalues judged to be significantly nonzero, and the corresponding eigenvectors span an ``estimate'' of the tangent space $T_{\bm{x}_k}M$. 
A global estimate is then obtained by averaging over all $N$ neighborhoods:
\[
\hat{d} \;=\; \frac{1}{N}\sum_{k=1}^N \hat{d}_k.
\]

This method is straightforward, computationally fast, and widely used in practice. 
However, many studies on manifold dimension estimators do not compare against it. 
A commonly cited reason is the need to select a threshold for determining which eigenvalues are significant, which introduces some arbitrariness. 
A typical rule, proposed by \citet{fukunaga1971algorithm} is to take the index of the smallest eigenvalue which exceeds $5\%$ of the largest eigenvalue.
Our experiments show that with this simple rule, \textsc{Local PCA} often outperforms most alternative estimators. 

Several methods have been proposed to avoid the subjective threshold choice. 
For example, \citet{bouveyron2011intrinsic} impose an isotropic variance structure within each neighborhood, which yields a consistent maximum likelihood estimator of the local dimension. 
More recently, \citet{lim2024tangent} proposed an alternative criterion:
\[
\hat{d}_k = \underset{1\leq q\leq p}{\argmin} \sum_{j=1}^p \left( \frac{\hat{\lambda}_{j}}{R^2} - \frac{1}{q+2}\,1_{\{j \leq q\}} \right)^2,
\]
motivated by the fact that if $\bm{x}_k$ and its neighbors are uniformly distributed in $B_{\bm{x}_k}(1)$, 
then the population covariance has its first $d$ eigenvalues equal to $(d+2)^{-1}$ and the remaining $p-d$ eigenvalues equal to zero. 
Here the scaling factor $R$ is estimated as the distance between $\bm{x}_k$ and its $K$-th nearest neighbor $\bm{x}_k^K$.


\subsubsection{The MADA estimator}


The manifold-adaptive dimension estimator (\textsc{MADA}) proposed by \citet{farahmand2007manifold} is a refinement of an idea originally introduced by \citet{pettis1979intrinsic}. 
It builds on the fact that the volume of a $d$-dimensional ball grows proportionally to $r^d$, a principle underlying the construction of many other estimators as well; see, e.g., \citet{fan2009intrinsic, kleindessner2015dimensionality}. 

Under the flatness assumption, the probability that a random data point $\bm{X}$ falls within $B_{\bm{x}_k}(R)$ is
\[
\mathbb{P}(\bm{X} \in B_{\bm{x}_k}(R)) = f_{\bm{X}}(\bm{x}_k)\, V_d \, R^d,
\]
where $V_d = \pi^{d/2}/\Gamma(d/2+1)$ is the volume of the unit ball in $\mathbb{R}^d$. 

Given $\bm{x}_k$ and the two neighbors $\bm{x}_k^{\ell_1}$ and $\bm{x}_k^{\ell_2}$, with $\ell_1 = \lceil K/2 \rceil$ and $\ell_2 = K$, let
$r_1$ and $r_2$ be the distances from $\bm{x}_k$ to its $\ell_1$-th and $\ell_2$-th nearest neighbors, respectively. Taking the logarithm of the above expression, we obtain
\[
\log \mathbb{P}(\bm{X} \in B_{\bm{x}_k}(R)) 
= \log \!\big(f_{\bm{X}}(\bm{x}_k)V_d\big) + d \log R,
\]
which holds for both $r_1$ and $r_2$. 
Approximating the probabilities by empirical proportions gives
\[
\log \frac{K}{2n} \;\doteq\; \log \!\big(f_{\bm{X}}(\bm{x}_k)V_d\big) + d \log r_1,
\qquad
\log \frac{K}{n} \;\doteq\; \log \!\big(f_{\bm{X}}(\bm{x}_k)V_d\big) + d \log r_2.
\]
Eliminating the nuisance term $\log \!\big(f_{\bm{X}}(\bm{x}_k)V_d\big)$ yields the local estimator
\[
\hat{d}_k = \frac{\log 2}{\log r_2 - \log r_1}.
\]
A global estimate $\hat{d}$ is then obtained by averaging the $\hat{d}_k$’s across all neighborhoods or, alternatively, by a majority vote.

\subsubsection{The MLE estimator}

The maximum likelihood estimator (\textsc{MLE}) proposed by \citet{levina2004maximum} is one of the most influential methods for intrinsic dimension estimation. 
Although maximum likelihood arguments also appear in the derivation of many other estimators, the approach of \citet{levina2004maximum} is widely recognized under the name \textsc{MLE}. 

Under the flatness assumption, let $N_{\bm{x}_k}(r)$ denote the number of data points falling within the $d$-dimensional ball $B_{\bm{x}_k}(r) \subseteq B_{\bm{x}_k}(R)$ for $0<r\leq R$. 
As $r$ increases from $0$ to $R$, the distribution of $N_{\bm{x}_k}(r)$ can be approximated by an inhomogeneous Poisson process with rate function
\[
\lambda_k(r) = n f(\bm{x}_k) \, d \, r^{d-1} V_d,
\]
where $V_d = \pi^{d/2}/\Gamma(d/2+1)$ is the volume of the $d$-dimensional unit ball. 
This form arises because the expected number of data points in $B_{\bm{x}_k}(r)$ is $n f(\bm{x}_k)$ times the volume of the ball. 

With this approximation, the log-likelihood for the observed neighborhood distances is
\[
\ell(d,\theta) 
= \sum_{\ell=1}^K \log \lambda_k\!\left(\lVert \bm{x}_k^\ell - \bm{x}_k \rVert\right) 
- \int_0^R \lambda_k(t)\,dt,
\]
where $\theta = \log (n f(\bm{x}_k))$ is treated as a nuisance parameter. 
Maximization with respect to $d$ yields the local estimator
\[
\hat{d}_k = \left[ \frac{1}{K} \sum_{\ell=1}^K 
\log \frac{R}{\lVert \bm{x}_k^\ell - \bm{x}_k \rVert} \right]^{-1},
\]
where $R$ is typically taken as the distance from $\bm{x}_k$ to its $K$-th nearest neighbor $\bm{x}_k^K$. 
A global estimate $\hat{d}$ is then obtained by averaging over all local estimates. 

Several refinements of \textsc{MLE} have since been proposed. 
For example, \citet{gupta2012regularized} introduce a regularized likelihood formulation, while \citet{gomtsyan2019geometry} modify the rate function to reduce reliance on the flatness assumption.

\subsubsection{The \textsc{DanCo} estimator}

A major limitation of \textsc{MLE} and many other flatness-based estimators, as will be shown in our experiments, is that their performance deteriorates in the form of underestimation as the intrinsic dimension $d$ increases. The reason is that, as the intrinsic dimension $d$ grows, the local geometry of the manifold becomes increasingly complex. As we discussed earlier, fully characterizing the local structure requires up to $(p - d)d$ principal curvatures. Since $p$ has to grow with $d$ as well, the number of data points needed within a neighborhood for reliable estimation increases rapidly. When the sample size $n$ is fixed, we might not be able to choose a large enough neighborhood size $K$ since a very big neighborhood will jeopardize the flatness assumption.

The dimensionality estimator based on angle and norm concentration, or \textsc{DanCo} \citep{CERUTI20142569}, is primarily designed to improve estimation performance on high-dimensional manifolds from a small sample. It builds upon two of its predecessors \textsc{IDEA} and \textsc{MIND}\textsubscript{KL} \citep{rozza2011idea, rozza2012novel}. \textsc{DanCo} employs two key ideas. First, it leverages both the norm distribution and the angle distribution of data points within a neighborhood. Second, it estimates the Kullback–Leibler (KL) divergence between the estimated likelihood and a simulated likelihood obtained from data under idealized conditions. The later approach mitigates the issue of data sparsity that affects maximum likelihood estimation in high dimensions.

Under the flatness assumption, for a given $\bm{x}_k$ and its neighbors, it can be shown that the ratio of the minimum to maximum distances (norms), defined as
\begin{equation*}
r_k = \frac{\|\bm{x}_k^1 - \bm{x}_k\|}{\|\bm{x}_k^K - \bm{x}_k\|},
\end{equation*}
is distributed according to the density function
\begin{equation*}
f_\text{norm}(r; d) = Kd\,r^{d - 1}(1 - r^{d})^{K - 1}.
\end{equation*}
Hence, given the values $r_1, \ldots, r_n$ computed from the sample, we can obtain the maximum likelihood estimate of $d$, denoted as $\hat{d}_\text{norm}$.

On the other hand, motivated by the observation that random vectors drawn uniformly from the unit sphere in high-dimensional spaces tend to be nearly orthogonal, the authors conjecture that for sufficiently large $d$, the mutual angles among the vectors $\bm{x}_k^\ell - \bm{x}_k$ will concentrate around some theoretical mean $\nu$ (not to be confused with the Hausdorff measure) and can be modeled using the von Mises distribution with density
\begin{equation*}
f_\text{angle}(\theta; \tau, \nu, d) = \frac{e^{\tau\cos(\theta - \nu)}}{2\pi I_{d/2 - 1}(\nu)},
\end{equation*}
where $I_{d/2 - 1}$ is the modified Bessel function of the first kind \citep{mardia2009directional}. Although $d$ appears as a parameter in the above density, obtaining an analytical form for its maximum likelihood estimate is intractable. Instead, \textsc{DanCo} only estimates the parameters $\tau$ and $\nu$ locally using the $K(K - 1)/2$ pairwise angles within each neighborhood, then averages the local estimates to obtain global estimates $\hat{\tau}$ and $\hat{\nu}$.

Next, for $q = 1, \ldots, p$, one simulates a dataset of size $n$ consisting of points uniformly distributed in the $q$-dimensional unit ball, and computes estimates $\hat{d}_q$, $\hat{\tau}_q$, and $\hat{\nu}_q$ using the same procedures for norm and angle distributions. Then, because of the independence between norms and angles, the Kullback–Leibler divergence between the empirical and simulated densities can be estimated as
\begin{equation*}
\text{KL}\big(f_\text{norm}(r;\, \hat{d}_\text{norm}), f_\text{norm}(r;\, \hat{d}_q)\big) + \text{KL}\big(f_\text{angle}(\theta;\, \hat{\tau}, \hat{\nu}, q) , f_\text{angle}(\theta;\, \hat{\tau}_q, \hat{\nu}_q, q)\big),
\end{equation*}
in which $\text{KL}(f_1, f_2) = \int_\R f_1(x)\log[f_1(x)/f_2(x)]dx$. Because of the flatness assumption, we expect the KL divergence to attain its minimum when $q = d$. This value of $q$ is then taken as the \textsc{DanCo} estimate of the intrinsic dimension.

A major drawback of \textsc{DanCo} is that it is arguably the most computationally intensive manifold dimension estimator, especially when the sample size $n$ and the ambient dimension $p$ are large. This is because the method requires simulating data for all dimensions from $1$ up to $p$, and comparing their corresponding estimates with those derived from the observed sample. The problem can become serious if the manifold is nonlinearly embedded into a space with much larger dimensions, as is the case for some real-world datasets (see Section~\ref{sec-realdata}). In this case, besides the computational cost, to obtain an accurate estimate, the method needs to produce a larger KL divergence estimate for every $q \neq d$, which is quite unlikely. 

To mitigate this, one can limit the maximum intrinsic dimension considered or, as in the \texttt{Matlab} implementation provided by  \citet{lombardi2020intrinsic}, use spline interpolation on a coarser set of simulated datasets spanning different sample sizes and intrinsic dimensions (with some fixed neighborhood size). This avoids simulating every dimension by smoothly estimating the KL-divergence curve, stabilizing the estimate by reducing noise. Additionally, if the initial maximum likelihood estimate is very small, the method skips the KL-divergence comparison and relies on the initial estimate to prevent unstable corrections and improve efficiency.

\subsubsection{The TLE estimator}

Like \textsc{DanCo}, the intrinsic dimension estimator with tight localities (\textsc{TLE}) from \citet{amsaleg2019intrinsic} also aims to tackle the problem of insufficient data points when estimating in neighborhoods. It builds on a previously proposed manifold dimension estimator proposed by the same authors \citep{amsaleg2018extreme}.

\textsc{TLE} is constructed on the concept of the so-called local intrinsic dimension (LID), which is a population parameter defined for any given point on the manifold and is determined by the local geometric structure as well as the data distribution. With the help of statistical theory about extreme values, for a given $\bm{x}_k$ and its neighbors, the LID at $\bm{x}_k$ can be estimated as
\eqn{
\hat{d}_k = -\lb\frac{1}{K}\sum_{\ell=1}^K\ln\frac{\norm{\bm{x}_k^\ell - \bm{x}_k}}{\norm{\bm{x}_k^K - \bm{x}_k}}\rb^{-1}.
}

It is easy to show that under the flatness assumption, the LID at $\bm{x}_k$ is the same as the intrinsic dimension $d$, hence, the above expression can be used as a manifold dimension estimator as well, and a global estimate can simply be taken as the average of the $\hat{d}_k$s.

The extreme value theory does not change the problem that $\hat{d}_k$ will suffer if the neighborhood size is too small. To tackle the issue, \textsc{TLE} further makes use of the fact that different points in the neighborhood roughly have the same LID, and improves the estimation by obtaining more estimates within each neighborhood. More specifically, $\hat{d}_k$ is modified as
\[
\hat{d}_k = -\lb\frac{1}{2K(K - 1)}\sum_{\bm{v} \neq \bm{w}}\left[\log\frac{d(\bm{v}, \bm{w})}{R} + \log\frac{d(2\bm{x}_k - \bm{v}, \bm{w})}{R}\right]\rb ^ {-1},
\]
where $\bm{v}, \bm{w}$ are taken from the $K$ nearest neighbors of $\bm{x}_k$, $R$ is estimated as $\norm{\bm{x}_k - \bm{x}_k^K}$, and $d(\bm{v}, \bm{w})$ is defined as\footnote{The formula assumes $\norm{\bm{v} - \bm{x}_k} < R$, when $\norm{\bm{x}_k - \bm{v}} = R$, the formula should be $d(\bm{w}, \bm{v}) = R\norm{\bm{w} - \bm{v}}^2 / [2(\bm{x}_k - \bm{v})\cdot (\bm{w} - \bm{v})]$.}
\eqn{
d(\bm{v}, \bm{w}) = \sqrt{[\bm{u}\cdot (\bm{x}_k - \bm{v})]^2 + R \bm{u}\cdot(\bm{w} - \bm{v})} - \bm{u}\cdot(\bm{x}_k - \bm{v}), \text{ where }\bm{u} = \frac{R(\bm{w} - \bm{v})}{R^2 - \norm{\bm{x}_k - \bm{v}}^2},
}
which is the radius of the circle centred on the line segment between $\bm{x}_k$ and $\bm{v}$ and passing through $\bm{v}$ and $\bm{w}$.

\subsubsection{The \textsc{TwoNN} estimator}

We conclude flatness-based estimators with the two nearest neighbor estimator (\textsc{TwoNN}) proposed by \citet{facco2017estimating}, which pushes the flatness assumption to its extreme by only making use of the two nearest neighbors of every~$\bm{x}_k$, hence avoiding curvatures as much as possible. Let $r_k^1 = \norm{\bm{x}_k - \bm{x}_k^1}$ and $r_k^2 = \norm{\bm{x}_k - \bm{x}_k^2}$ be the distances of $\bm{x}_k$ to its two nearest neighbors. The key insight of \textsc{TwoNN} is that under the flatness assumption, the distribution function of the ratio of the distances, $\mu_k = r_k^2/r_k^1$, is
\eqn{F(\mu) = (1 - \mu^{-d})1_{[0,\infty]}(\mu),}
which not only depends on $d$, but also is independent of the data distribution. As a result, we do not need any manipulation like in \citet{pettis1979intrinsic} and \textsc{MADA}, and can obtain a global estimate of $d$ directly by employing the following relation
\eqn{
-\frac{\log[1 - F(\mu)]}{\log \mu} = d.
}

All ratios $\mu_k$s in the sample are obtained first to construct an empirical distribution function for $F$, and $\hat{d}$ is estimated through the least squares method with $\log[1 - \hat{F}(\mu_k)]$s as responses and $-\log \mu_k$s as predictors. Our experiments show that such a simple and elegant design works very well in many cases, making it a top performer among all flatness-based estimators.

\vspace{1em}

This list of flatness-based estimators is far from exhaustive: several older but well-known estimators also fall into this category, including \textsc{CC} \citep{grassberger1983measuring, kalantana2013intrinsic}, \textsc{kNN} \citep{costa2004geodesic, costa2004learning}, and \textsc{Hein} \citep{hein2005intrinsic}, along with a number of lesser-known approaches \citep{bennett1969intrinsic, trunk1968statistical, johnsson2014low, thordsen2022abid}.  
Many of these have already been reviewed in earlier studies \citep{verveer1995evaluation, campadelli2015intrinsic, camastra2016intrinsic}.  

\subsection{Other Estimators}

Flatness-based estimators behave poorly when the flatness assumption is significantly violated, which typically occurs on manifolds with complex curvature, such as those that are nonlinearly embedded in a Euclidean space.

\subsubsection{The \textsc{CA-PCA} estimator}

Our experimental results indicate that among the flatness-based estimators, \textsc{TwoNN} is the most robust to curvature, likely because it minimizes curvature effects by relying solely on the two nearest neighbors of each data point. At the other end of the design spectrum are methods that explicitly account for curvature. A recent example of this approach is the curvature-adjusted PCA estimator, \textsc{CA-PCA} \citep{gilbert2025pca}, which incorporates curvature information directly into the estimation procedure.

Recall the threshold-free local PCA estimator,
\[
\hat{d}_k = \underset{1\leq q\leq p}{\argmin} \sum_{j = 1}^p \lb \frac{1}{R^2}\hat{\lambda}_{j} - \frac{1}{j + 2}1_{j \leq q} \rb^2,
\]
which is based on the fact that under the flatness assumption (within the unit ball), the true covariance matrix $\Sigma_k$ has its first $d$ eigenvalues equal to $1/(d + 2)$ and the $p - d$ remaining ones equal to $0$ \citep{lim2024tangent}.

If the flatness assumption holds reasonably well, the sample covariance matrix $\hat{\Sigma}_k$ computed from the neighborhood of $\bm{x}_k$ should closely resemble this theoretical structure, which will not be the case if the assumption is seriously violated.

The main idea of \textsc{CA-PCA} is to adjust the eigenvalues to account for curvature. By the local graph representation for manifolds, we know that in a neighborhood of $\bm{x}_k$, points can be expressed in new coordinates as
\[
\big(\bm{x}_{1:d}',\, g(\bm{x}_{1:d}')\big),
\]
where $\bm{x}_{1:d}'$ denotes coordinates with respect to directions within the tangent space $T_{\bm{x}_k}M$, and $g$ is a sufficiently differentiable function. The flatness assumption corresponds to taking $g \doteq 0$, which is equivalent to its first-order Taylor approximation and thus ignores curvature.

To incorporate curvature, we can use a second-order Taylor polynomial to approximate each component of $g$, which leads to
\begin{equation*}
g_{\ell}(\bm{x}_{1:d}') \doteq q_\ell(\bm{x}_{1:d}') = \bm{x}_{1:d}'^\top Q_{\ell} \bm{x}_{1:d}', \ \ell = 1,\cdots, p-d,
\end{equation*}
where $Q_\ell$ is a symmetric $d \times d$ matrix capturing principal curvatures in the $\ell$-th normal direction.

\begin{figure}[H]
    \centering
    \begin{subfigure}[b]{0.35\textwidth}
        \centering
        \includegraphics[width=\textwidth]{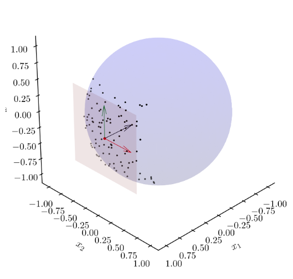}
    \end{subfigure}
    \hfill
    \begin{subfigure}[b]{0.35\textwidth}
        \centering
        \includegraphics[width=\textwidth]{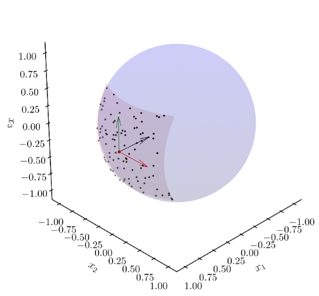}
    \end{subfigure}
    \caption{Approximate a neighborhood with the tangent space (left) and the tangent quadratic space (right).}
    \label{fig:quadraticsurface}
\end{figure}

In other words, instead of assuming the $\bm{x}_k^\ell$s are uniformly distributed in some $d$-dimensional ball within the tangent space, we assume they are uniformly distributed in the \emph{tangent quadratic space} defined by the $q_\ell$s, as illustrated in Figure~\ref{fig:quadraticsurface}, which should be a better approximation. 

\citet{gilbert2025pca} show that, with the quadratic approximation, the theoretical eigenvalues satisfy 
\begin{equation*}
\frac{1}{d + 2} \doteq \lambda_j  + \frac{3d + 4}{d(d + 4)} \sum_{i = d + 1}^p \lambda_i.
\end{equation*}

Using this property, \textsc{CA-PCA} replaces the true eigenvalues with their sample version, and estimates the intrinsic dimension as
\begin{equation*}
\hat{d}_k = \underset{1\leq q\leq p}{\argmin} \left\{\sqrt{\sum_{j=1}^q \left( \frac{1}{q + 2} - \frac{1}{R^2}\lb \hat{\lambda}_{j} + \frac{3q + 4}{q(q + 4)} \sum_{i = q + 1}^p \hat{\lambda}_{ki} \rb \right)^2} + \frac{2}{R^2} \sum_{j = q + 1}^p \hat{\lambda}_{j} \right\},
\end{equation*}
where $R$ can be estimated using $\|\bm{x}_k^K - \bm{x}_k\|$\footnote{The authors of \textsc{CA-PCA} actually suggest using $[\|\bm{x}_k^K - \bm{x}_k\| + \|\bm{x}_k^{K - 1} - \bm{x}_k\|]/2$, which is what we also use in our own  implementation.}, and the last term serves as a penalty for neighborhoods where the quadratic approximation is of poor quality.

A global \textsc{CA-PCA} estimate is then obtained by averaging all the local estimates $\hat{d}_k$ across the dataset.

\subsubsection{The \textsc{Wasserstein} estimator}

All estimators we have introduced so far make explicit use of the underlying manifold's geometry to some extent. In contrast, the manifold dimension estimator based on the Wasserstein distance, \textsc{Wasserstein} \citep{block2022intrinsic}, incorporates the geometry only implicitly in its design.

Given two probability distributions $\mu$ and $\nu$ on $\mathbb{R}^p$, their Wasserstein distance is defined as
\begin{equation*}
W_1(\mu, \nu) = \inf_{(\bm{X}, \bm{Y}) \sim \Gamma(\mu, \nu)} \mathbb{E} \left[ \sum_{j=1}^p |X_j - Y_j| \right],
\end{equation*}
where $\bm{X}$ and $\bm{Y}$ are $p$-dimensional random vectors with marginal distributions $\mu$ and $\nu$, respectively, and $\Gamma(\mu, \nu)$ denotes the set of all joint distributions with these marginals. 

Intuitively, the Wasserstein distance measures the minimum expected transportation cost required to morph one probability distribution into another. When the marginals are discrete, such as empirical distributions constructed from samples, the Wasserstein distance can be computed exactly by solving a discrete optimization problem \citep{villani2021topics}.

\textsc{Wasserstein} is designed based on the theoretical result \citep{dudley1969speed} that the rate at which the difference between a distribution and its empirical estimate (measured in Wasserstein distance) converges as the sample size increases is governed by the intrinsic dimension $d$ of the manifold the distribution is supported on, rather than the ambient dimension $p$. Specifically, let $P$ and $P_n$ be the true and empirical distributions, it holds that
\[
W_1(P_n, P) \propto n^{-\frac{1}{d}},
\]
where the distance used in computing the Wasserstein metric is ideally the geodesic distance along the manifold.

The above result leads to the following expression of $d$ which can be useful for the design of an estimator: given two samples of sizes $m$ and $\alpha m$ ($\alpha>1$ such that $\alpha m\in\mathbb{N}$) distributed according to $P$, we have
\begin{equation*}
d = \frac{\log \alpha}{\log W_1(P_m, P) - \log W_1(P_{\alpha m}, P)}.
\end{equation*}
In practice, $P$ is unknown. To overcome this, the authors propose splitting the full dataset into four parts of sizes $m, m, \alpha m$ and $\alpha m$, which gives $4$ empirical distribution estimates $P_m, P_m'$, $P_{\alpha m}$ and $P_{\alpha m}'$. Using $P_m'$ and $P_{\alpha m}'$ to approximate $P$, the dimension can then be estimated as
\begin{equation*}
\hat{d} = \frac{\log \alpha}{\log W_1(P_m, P_m') - \log W_1(P_{\alpha m}, P_{\alpha m}')}.
\end{equation*}

For simplicity, Euclidean distances can be used as a substitute for geodesic distances when computing Wasserstein distances. Since $\hat{d}$ is a global estimate, we should also generate multiple instances of it through repeated data splits, and to average these estimates in order to reduce uncertainty. For $d \geq 3$, under certain regularity conditions, it can be shown that $\hat{d}$ is close to $d$ with high probability. However, verifying or establishing the conditions is nontrivial. 

\vspace{1em}

Other well-known methods in this category include \textsc{ISOMAP} \citep{balasubramanian2002isomap}, which seeks a low-dimensional embedding into the Euclidean space that preserves geometric properties such as geodesic distances, and \textsc{Diffusion Maps} \citep{coifman2006diffusion}, which constructs a transition matrix that captures the manifold’s intrinsic geometry.  
These two algorithms, however, are primarily designed for the broader task of recovering the full manifold structure rather than solely estimating its dimension, though they can be readily adapted for this purpose.

\section{Empirical Evaluation}\label{sec-evaluation}

In this section, we present a comprehensive set of numerical experiments designed to evaluate the performance of the eight estimators introduced in Section~\ref{sec:manifold_dimension_estimators}. Among them, \textsc{DanCo} was previously identified as the top performer by \citet{campadelli2015intrinsic}, while \textsc{MLE}, \textsc{MADA}, and \textsc{TwoNN} are well-established and widely used. By contrast, \textsc{TLE}, \textsc{Wasserstein}, and \textsc{CA-PCA} are more recent proposals that, to our knowledge, have not yet been systematically compared with earlier methods. We also include \textsc{Local PCA}, despite its frequent omission from benchmarks, using the threshold rule by \citet{fukunaga1971algorithm}.

Although \texttt{R} implementations of manifold dimension estimators are available, they remain relatively underdeveloped compared to their Python counterparts. See \citet{you2022rdimtools} for recent advances in the \texttt{Rdimtools} package \citep{Rdimtools}, which consolidates and extends earlier efforts and will be further introduced in the next section. Moreover, Python implementations are generally faster, which is why we chose to perform all experiments in this section using Python. Specifically, we use the estimator implementations provided by the \texttt{scikit-dimension} (\texttt{skdim}) package \citep{bac2021scikit} for \textsc{Local PCA}, \textsc{MLE}, \textsc{DanCo}, \textsc{MADA}, \textsc{TLE}, and \textsc{TwoNN}, while the \textsc{Wasserstein} and \textsc{CA-PCA} methods are implemented by ourselves. We also implement our own sampling and evaluation procedures.

We set the number of replicates to $100$, meaning that each experiment was repeated 100 times on independently generated random samples under identical conditions. The average and standard deviation of the estimates are reported to evaluate the bias and variance of the estimators.

\subsection{Impact of Individual Factors on Performance}

The performance of a manifold dimension estimator can be affected by numerous factors. Among the most critical ones are the hyperparameters (e.g., the neighborhood ``size'' $K$), the sample size $n$, the ambient dimension $p$, the curvature of the manifold, and the level of noise.

Before undertaking large-scale comparisons across datasets and estimators, it is useful to benchmark each method under controlled variations of these factors. This approach clarifies the strengths, limitations, and sensitivities of the estimators.

In this subsection, we report results from a series of controlled experiments we conducted to examine how the estimators respond to changes in these key factors. Most of our controlled experiments are conducted on a $5$-dimensional sphere, given it is a widely studied manifold that serves as a standard benchmark for dimension estimation. Its geometry is well understood, and its curvature can be easily adjusted by varying the radius. Choosing dimension $d = 5$ offers a suitable level of difficulty for the estimators: challenging enough to be informative, yet not so high as to obscure meaningful differences in performance.

\subsubsection{Effect of hyperparameters}

The choice of hyperparameter values is of crucial importance to the performance of manifold dimension estimators \citep{campadelli2015intrinsic}. Among the eight methods considered, only \textsc{TwoNN} is parameter-free. For \textsc{Local PCA}, \textsc{MLE}, \textsc{DanCo}, \textsc{MADA}, \textsc{TLE} and \textsc{CA-PCA}, the main hyperparameter is the neighborhood size $K$, i.e., a neighborhood consisting of the $K$ nearest neighbors (out of the total $n$ points) of the selected point. By contrast, \textsc{Wasserstein} relies on a different type of hyperparameter: the subsample splitting ratio $\alpha \in (1, \infty)$. This parameter determines how a sample of size $n$ is divided into four parts of sizes \( m \), \( m \), \( \alpha m \), and \( \alpha m \) where $m = n / (2 + 2\alpha)$. For each $\alpha$, the random split is conducted 100 times for a reliable estimation.

When the sample size $n$ is fixed, choosing $K$ involves a trade-off. On the one hand, smaller neighborhoods are desirable because many estimators rely on local approximations of the manifold, which are more accurate when the neighborhood is nearly flat (or quadratic). On the other hand, larger neighborhoods provide more points per estimate, which can improve stability and accuracy, particularly for estimators based on maximum likelihood, whose performance typically benefits from larger sample sizes.

The experiment designed to benchmark the estimators' sensitivity to hyperparameter choices is summarized in Table~\ref{tab:nbh}.

\begin{table}[htbp]
\centering
\caption{Experiment design for evaluating the effect of $K$ or $\alpha$.}
\label{tab:nbh}
\begin{tabular}{@{} l p{0.65\linewidth} @{}}
\toprule
\textbf{Component} & \textbf{Description} \\
\midrule
Manifold $M$       & A $5$-dimensional sphere linearly embedded in $\mathbb{R}^{10}$. \\
Sample             & $n = 1{,}000$ data points uniformly distributed over $M$. \\
Hyperparameter     & Neighborhood size $K$ or regularization parameter $\alpha$. \\
Factor             & • $K \in \{5, 10, 20, 30, 40, 50, 100, 200, 500, 750\}$ (general)\\
                   & • $K \in \{2, 4, 6, \dots, 20\}$ for \textsc{DanCo} \\
                   & • $\alpha \in \{1.01, 1.2, 1.4, 1.6, 1.8, 2, 4, 6, 8, 10\}$ \\
\bottomrule
\end{tabular}
\end{table}

\begin{figure}[htbp]
    \centering
    \begin{subfigure}[b]{0.48\textwidth}
        \centering
        \includegraphics[width=\textwidth]{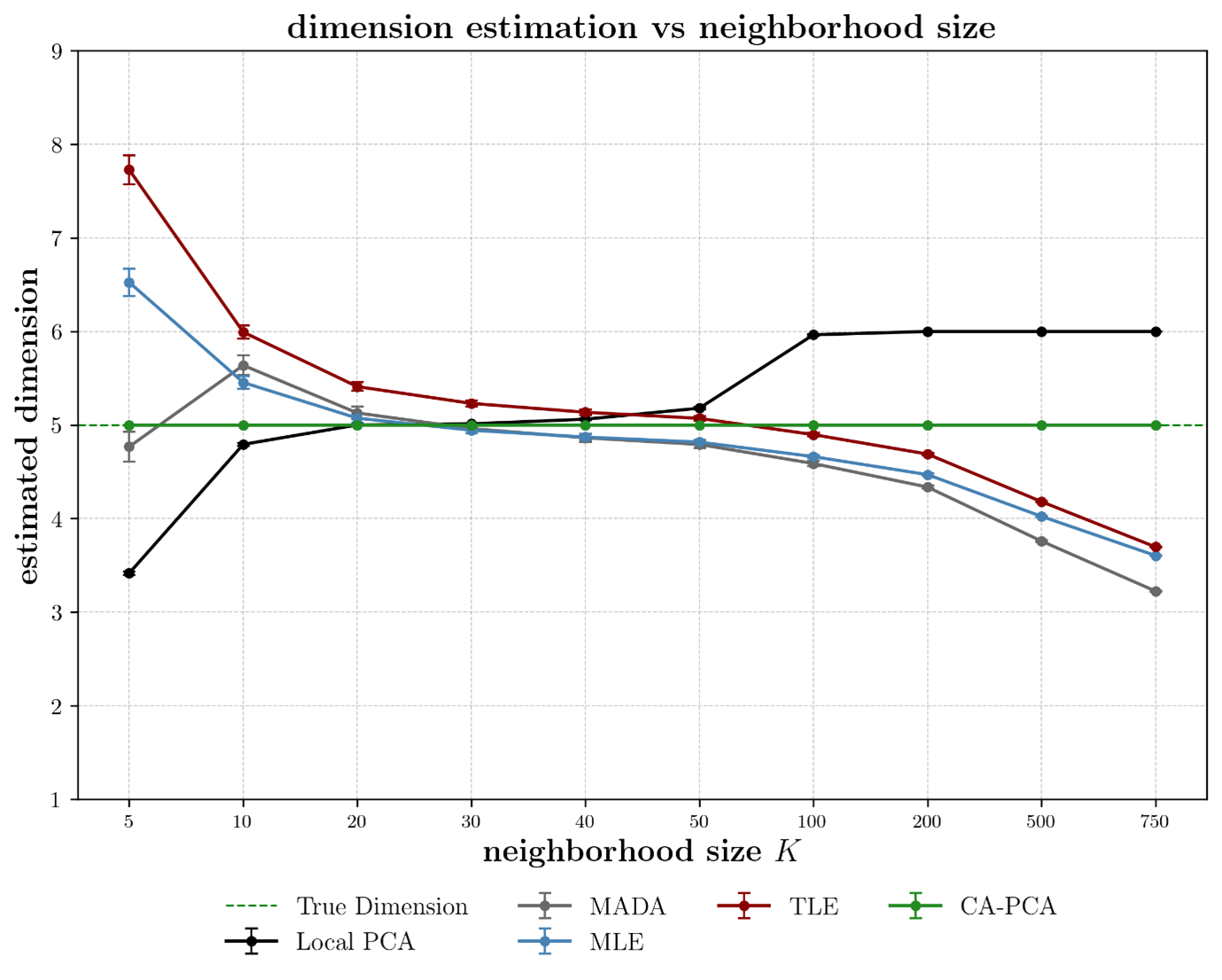}
        \caption{Dimension estimates versus $K$.}
        \label{fig:nbh01}
    \end{subfigure}
    \begin{subfigure}[b]{0.48\textwidth}
        \centering
        \includegraphics[width=\textwidth]{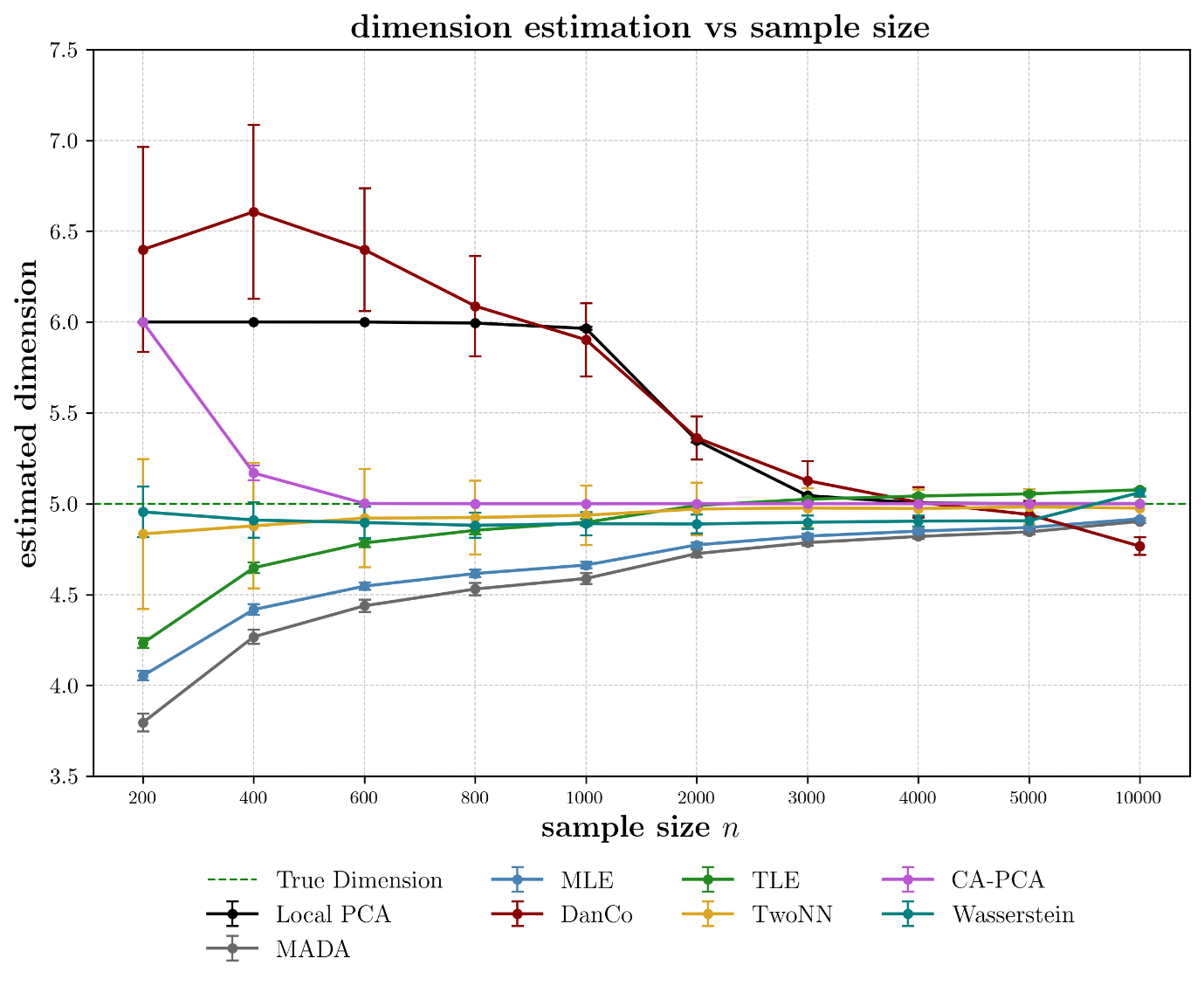}
        \caption{Dimension estimate versus sample size.}
        \label{fig:n}
    \end{subfigure}
    \caption{Dimension estimate versus neighborhood size (left) and sample size (right).}
\end{figure}

In Figure~\ref{fig:nbh01}, we show how the estimates of \textsc{Local PCA}, \textsc{MLE}, \textsc{MADA}, \textsc{TLE} and \textsc{CA-PCA} vary with $K$. We observe that the estimates of \textsc{Local PCA} increase with $K$, while those of \textsc{MLE}, \textsc{MADA}, and \textsc{TLE} decrease. In contrast, \textsc{CA-PCA}'s estimates remain stable across the tested range of $K$, suggesting that accounting for curvature may improve the estimator's robustness to hyperparameter selection. One possible explanation for this phenomenon is that the quadratic space approximation \textsc{CA-PCA} relies on works well for a much larger range of neighborhood sizes. All five estimators tend to stabilize when $K$ is between $20$ and $50$.

\begin{figure}[htbp]
    \centering
    \begin{subfigure}[b]{0.48\textwidth}
        \centering
        \includegraphics[width=\textwidth]{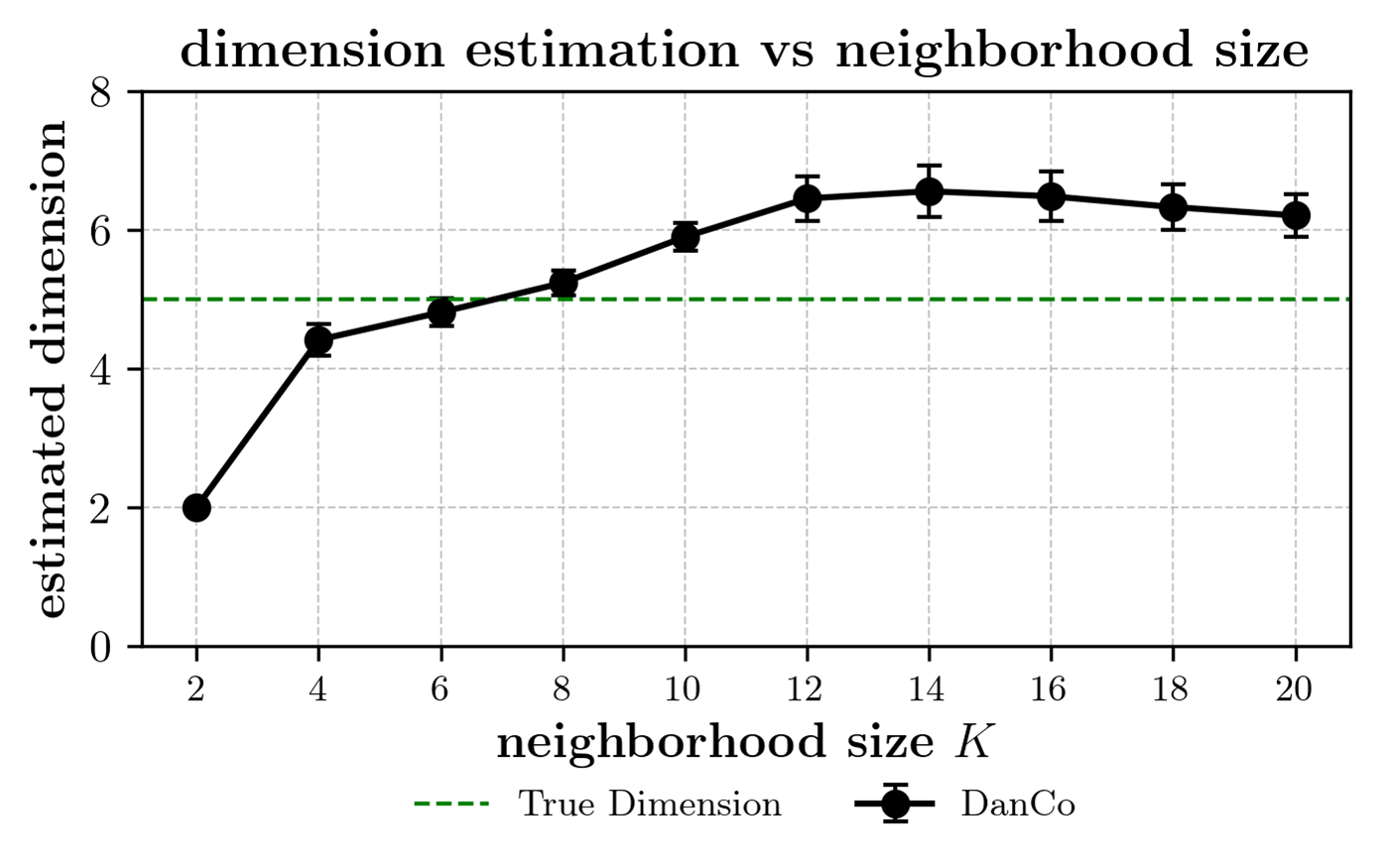}
        \caption{\textsc{DanCo}}
        \label{fig:DanCo}
    \end{subfigure}
    \hfill
    \begin{subfigure}[b]{0.48\textwidth}
        \centering
        \includegraphics[width=\textwidth]{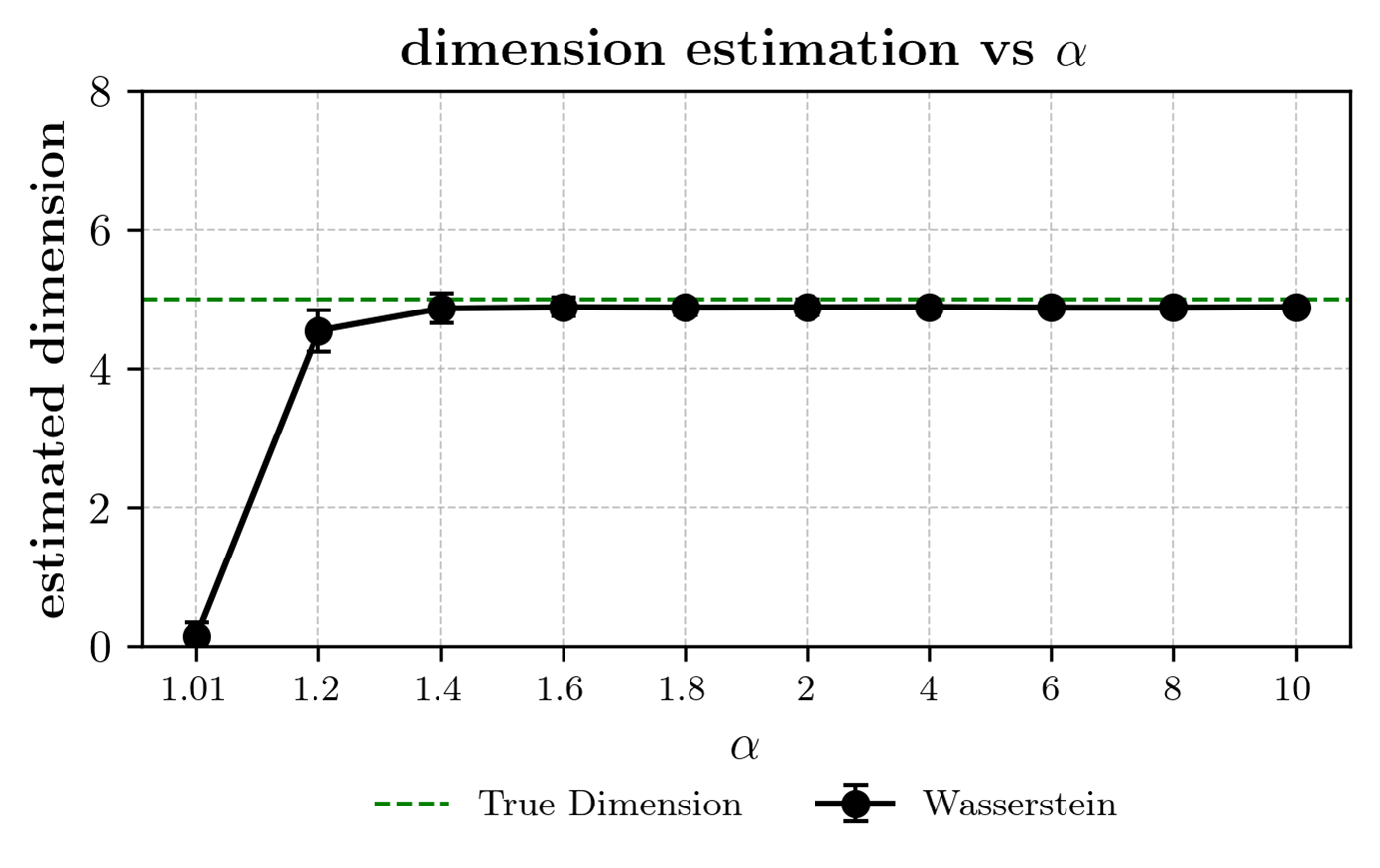}
        \caption{\textsc{Wasserstein}}
        \label{fig:Wasserstein}
    \end{subfigure}
    \caption{Dimension estimates versus $K$ or $\alpha$.}
    \label{fig:nbh02}
\end{figure}

In comparison, \textsc{DanCo} requires a much smaller neighborhood size, and thus its results are plotted separately in Figure~\ref{fig:DanCo}. The estimates of \textsc{Wasserstein} against $\alpha$ are plotted in Figure~\ref{fig:Wasserstein}.

In this particular experiments, \textsc{DanCo} gives the most accurate estimate when $K \doteq 7$. Our other experiments show that in general, \textsc{DanCo} performs the best with neighborhood sizes around $10$, which aligns with the setting used in \citet{CERUTI20142569}, although the authors did not provide an explanation for this choice.

The \textsc{Wasserstein} estimates stabilize for $\alpha > 1.4$.

\subsubsection{Effect of sample size}

\begin{figure}[htbp]
    \centering
    \includegraphics[width=\textwidth]{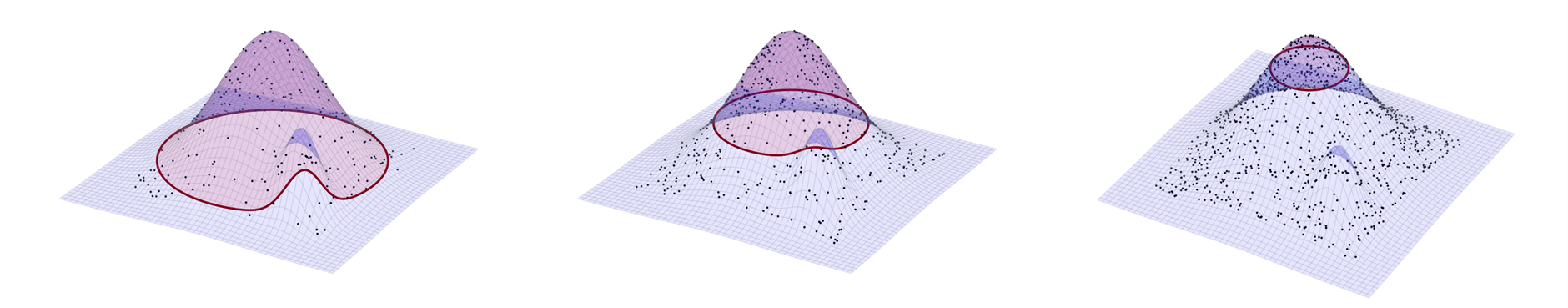}
    \caption{As the sample size $n$ increases, a fixed-size neighborhood $K$ more accurately captures the local structure of the underlying manifold.}
    \label{fig:samplesize}
\end{figure}

For flatness-based estimators, when all other conditions are fixed, the performance of an estimator is expected to improve as the sample size \( n \) increases. The improvement can be understood from two complementary perspectives. First, a fixed neighborhood size \( K \) combined with a larger \( n \) leads to a higher concentration of data points within each neighborhood on the manifold. This local densification improves the quality of linear approximations by the tangent space, and thus enhances estimation accuracy, as illustrated in Figure~\ref{fig:samplesize}. 
At the same time, a larger sample size also allows us to choose larger neighborhood sizes, which can further improve the quality of local estimates by reducing variance without sacrificing locality.

Our experiment design to benchmark the estimators' performance against sample sizes is shown in Table~\ref{tab:n}, which focuses on the asymptotic mode with $K$ fixed and $n$ tending to $\infty$, as it is more relevant for our choice of estimators.  Another asymptotic mode which has been considered in the literature is $K \to \infty, n \to \infty$ with their ratio fixed \citep{costa2004learning}. 

\begin{table}[htbp]
\centering
\caption{Experiment design for evaluating the effect of $n$.}
\label{tab:n}
\begin{tabular}{@{} l p{0.65\linewidth} @{}}
\toprule
\textbf{Component} & \textbf{Description} \\
\midrule
Manifold $M$       & a $5$-dimensional sphere linearly embedded in $\mathbb{R}^{10}$ \\
Sample          & $n = 1,000$ data points uniformly distributed on $M$\\
Hyperparameter  & $K = 100$ ($10$ for \textsc{DanCo}), $\alpha = 5$ \\
Factor          & sample size $n\in\{200,400,\ldots,1000,2000,\ldots,5000,10000\}$ \\
\bottomrule
\end{tabular}
\end{table}

We can see from Figure~\ref{fig:n} that for all estimators, both the bias and the variance of the estimates converge towards~$0$ as the sample size $n$ grows.

In particular, \textsc{CA-PCA} converges much faster than \textsc{Local PCA}, consistent with our previous result that it is more robust with respect to $K$ given fixed $n$. Besides, \textsc{TwoNN} and \textsc{Wasserstein} have very low bias even with small sample sizes, making them strong candidates when we do not have sufficient data. 


\subsubsection{Effect of ambient dimension}

A high ambient dimension $p$ can affect the performance of manifold dimension estimators in two distinct scenarios. As we discussed before, a manifold can be embedded \emph{linearly} into Euclidean spaces with much higher dimensions. In this case, the intrinsic structure of the manifold remains unchanged and the manifold still lies entirely within some lower-dimensional subspace; on the other hand, a nonlinear embedding into a higher-dimensional space results in a different manifold, typically with increased curvature and geometric complexity. In view of the dimension estimators, they will recover the intrinsic dimension of the image manifold, regardless of how it was embedded. Therefore, nonlinear embeddings, which alter the geometry of the manifold, are better studied when we consider curvature as the factor, and we focus on the effect of linear embeddings here.

\begin{table}[htbp]
\centering
\caption{Experiment design for testing the effect of $p$.}
\label{tab:p}
\begin{tabular}{@{} l p{0.65\linewidth} @{}}
\toprule
\textbf{Component} & \textbf{Description} \\
\midrule
Manifold $M$       & a $5$-dimensional sphere linearly embedded in $\mathbb{R}^{p}$ \\
Sample          & $n = 1,000$ data points uniformly distributed on $M$\\
Hyperparameter  & $K = 100$ ($K=10$ for \textsc{DanCo}), $\alpha = 5$ \\
Factor          & ambient dimension $p\in\{6,10,15,\ldots,50\}$ \\
\bottomrule
\end{tabular}
\end{table}

\begin{figure}[htbp]
    \centering
    \begin{subfigure}[b]{0.45\textwidth}
        \centering
        \includegraphics[width=\textwidth]{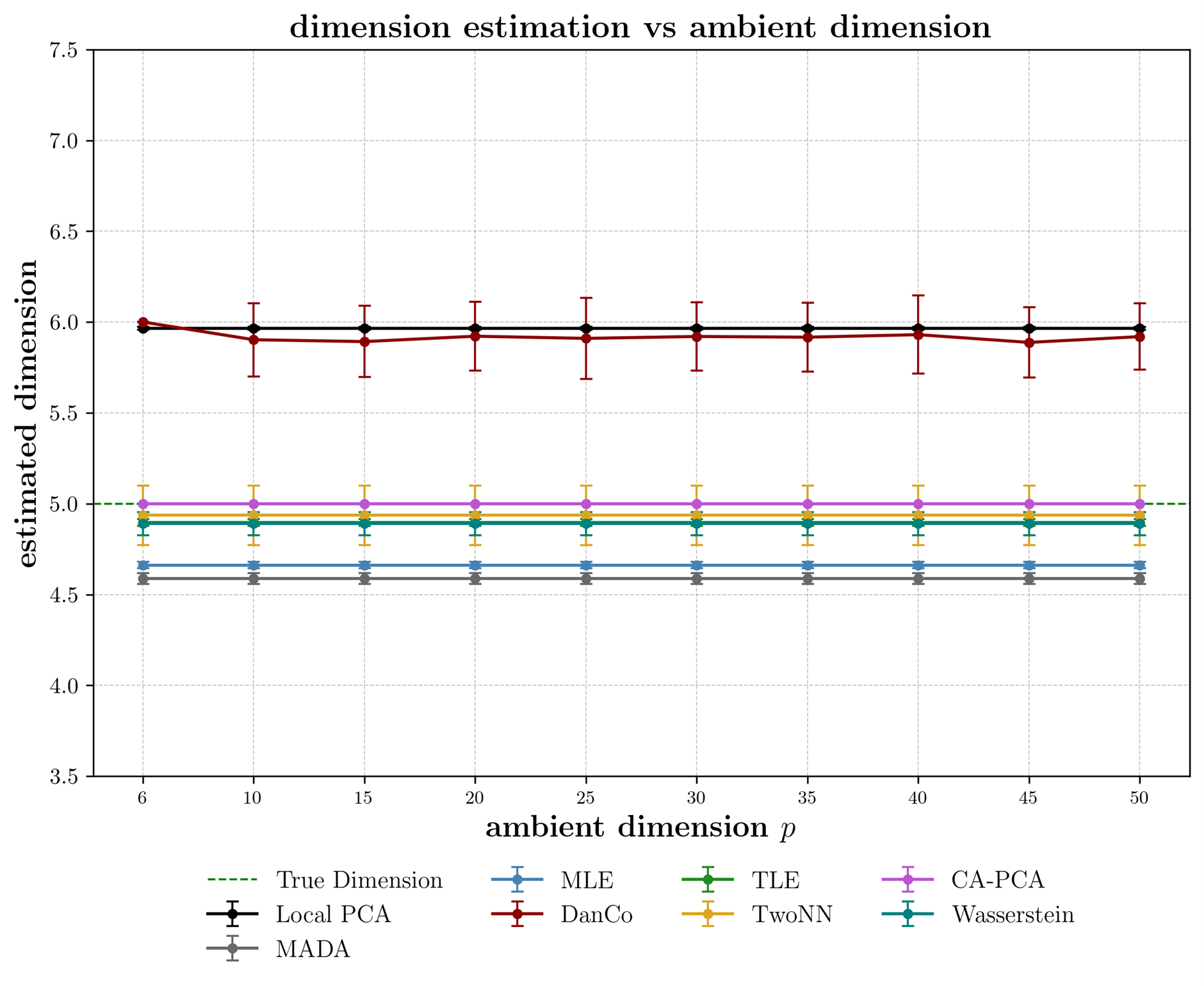}
        \caption{Dimension estimate versus ambient dimension.}
        \label{fig:p}
    \end{subfigure}
    \begin{subfigure}[b]{0.45\textwidth}
        \centering
        \includegraphics[width=\textwidth]{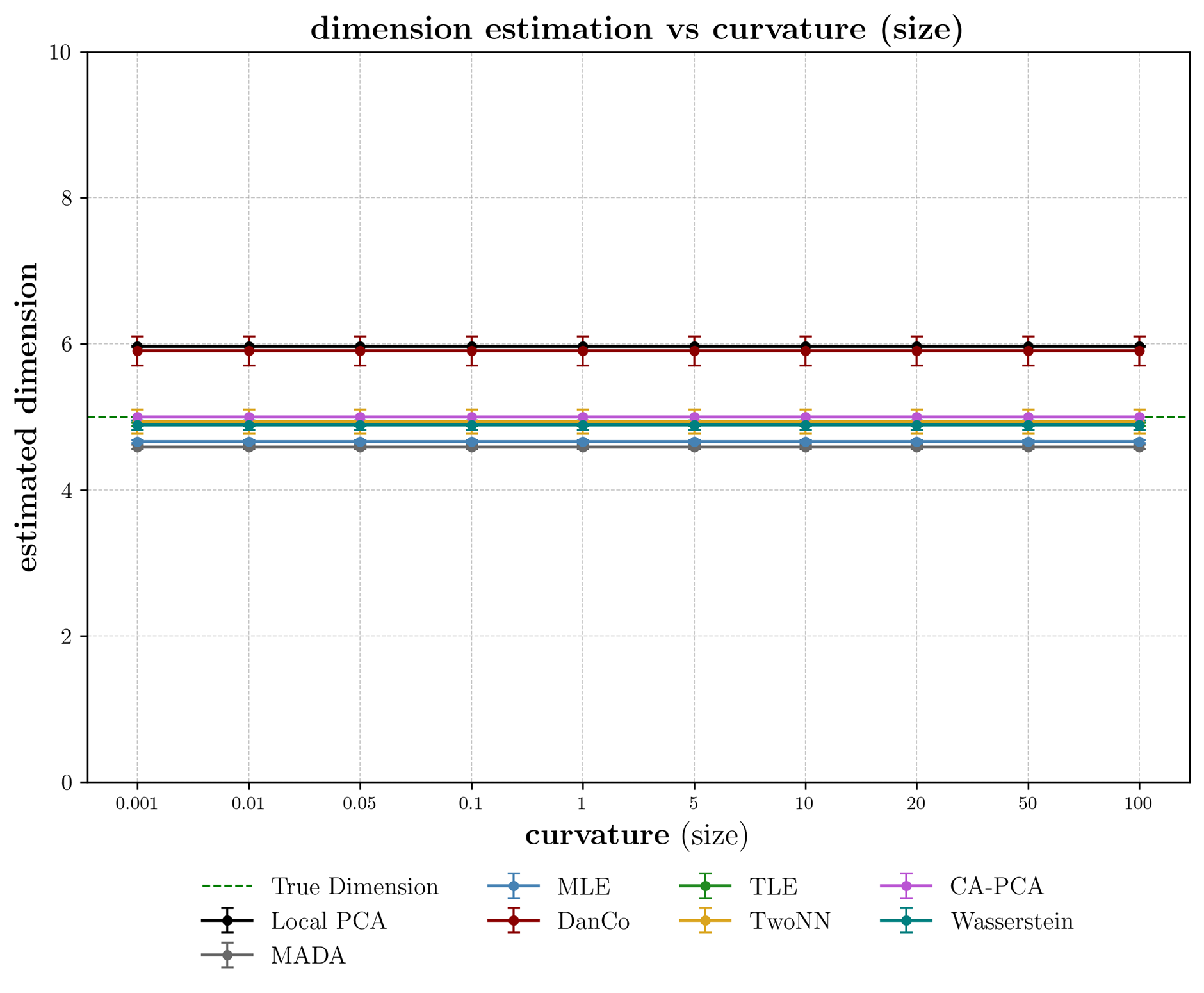}
        \caption{Dimension estimate vs curvature (size)}
        \label{fig:curvature01}
    \end{subfigure}
    \caption{Dimension estimate versus ambient dimension (left) and curvature size (right).}
\end{figure}

Intuitively, a linear embedding into higher ambient dimensions should not significantly impact the performance of dimension estimators. This can be confirmed by their mathematical formulations. For example, the \textsc{MLE} estimate
\eqn{
\hat{d}_k = \left[\frac{1}{K}\sum_{\ell=1}^K \log\frac{R}{\norm{\bm{x}_k^\ell - \bm{x}_k}}\right]^{-1}
}
depends only on norms of pairwise differences between data points in the neighborhood, which are invariant under linear embeddings. Our experiment confirms this analysis: as shown in Figure~\ref{fig:p}, the estimates from all estimators remain constant under linear embeddings into increasingly higher dimensional spaces\footnote{We observe fluctuations in the variance of \textsc{DanCo}, which we attribute to the randomness introduced by sampling from the ideal dataset; see the previous section for further details.}. The experiment design is summarised in Table~\ref{tab:p}. 

\subsubsection{Effect of curvature}

Three aspects of curvature can affect the performance of the estimators: the magnitude of curvature, the number of principal curvatures, and the variation of curvature along the manifold. In this section, we evaluate their effects through three separate experiments.

Intuitively, as suggested in much of the prior literature, it may seem that a larger curvature makes the manifold dimension estimation problem more difficult, since the manifold becomes less flat. However, our experiment shows that this relationship is not so straightforward. The magnitude of the principal curvatures only significantly affects estimator performance when combined with other factors such as noise; by itself, it does not appear to increase the difficulty of the problem.

We start by varying the curvature size through the radius of the sphere, as indicated in Table~\ref{tab:expc01}. 

\begin{table}[htbp]
\centering
\caption{Experiment design for evaluating the effect of curvature size.}
\label{tab:expc01}
\begin{tabular}{@{} l p{0.65\linewidth} @{}}
\toprule
\textbf{Component} & \textbf{Description} \\
\midrule
Manifold $M$        & a $d=5$-dimensional sphere linearly embedded in $\mathbb{R}^{10}$ \\
Sample          & $n = 1,000$ data points uniformly distributed on $M$\\
Hyperparameter  & $K = 100$ ($K=10$ for \textsc{DanCo}), $\alpha = 5$ \\
Factor          & radius $R$ of the sphere, ranging from $0.01$ to $100$ \\
\bottomrule
\end{tabular}
\end{table}

In Figure~\ref{fig:curvature01}, we show the estimates of various estimators on a $5$-dimensional sphere embedded in $\mathbb{R}^{10}$, with varying curvature values. The principal curvature of a sphere is everywhere $1/R$, so a smaller radius corresponds to a higher curvature. As shown in the figure, the performance of all eight estimators remains largely unaffected by changes in the curvature magnitude alone. However, we believe this does not imply a general insensitivity to curvature. 
The likely explanation is the perfect symmetry of the sphere: for a fixed sample size $n$ and neighborhood size $K$, each neighborhood covers the same proportion of the sphere regardless of its radius, producing identical estimates. Moreover, estimators like \textsc{DanCo} and \textsc{TLE} are scale-invariant by construction, so the invariance is expected.

Our next experiment evaluates the effect of the number of principal curvatures. In this setting, we consider spheres of increasing intrinsic dimension; see Table~\ref{tab:expc02} for details. 

\begin{table}[htbp]
\centering
\caption{Experiment design for evaluating the effect of the number of principal curvatures.}
\label{tab:expc02}
\begin{tabular}{@{} l p{0.65\linewidth} @{}}
\toprule
\textbf{Component} & \textbf{Description} \\
\midrule
Manifold $M$       & a $d$-dimensional sphere linearly embedded in $\mathbb{R}^{2d}$ \\
Sample          & $n = 1,000$ data points uniformly distributed on $M$\\
Hyperparameter  & $K = 100$ ($K=10$ for \textsc{DanCo}), $\alpha = 5$ \\
Factor          & $d\in\{2,4,\ldots,20\}$\\
\bottomrule
\end{tabular}
\end{table}

Since a $d$-dimensional sphere resides in a $(d + 1)$-dimensional subspace, the number of principal curvatures required to fully capture the local geometry at any point is $d \times 1 = d$.

As the curvature structure becomes more complex, more data points are generally required to accurately identify the underlying geometry. Therefore, we expect that for a fixed sample size $n$ and neighborhood size $K$, the performance of manifold dimension estimators will deteriorate as $d$ increases.

\begin{figure}[htbp]
    \centering
    \begin{subfigure}[b]{0.48\textwidth}
        \centering
        \includegraphics[width=\textwidth]{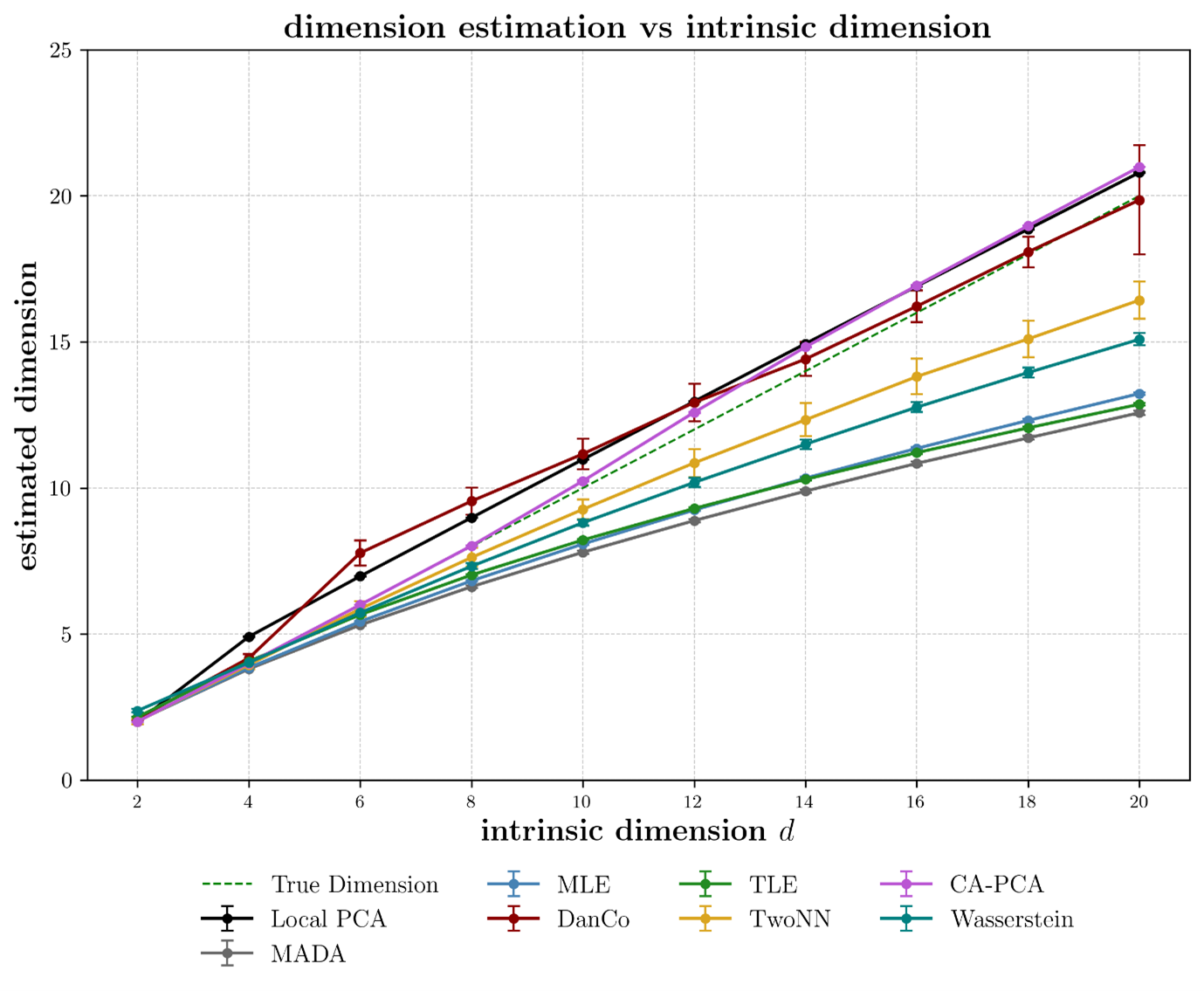}
        \caption{Dimension Estimate vs Curvature ($d$)}
        \label{fig:curvature02}
    \end{subfigure}
    \begin{subfigure}[b]{0.48\textwidth}
        \centering
        \includegraphics[width=\textwidth]{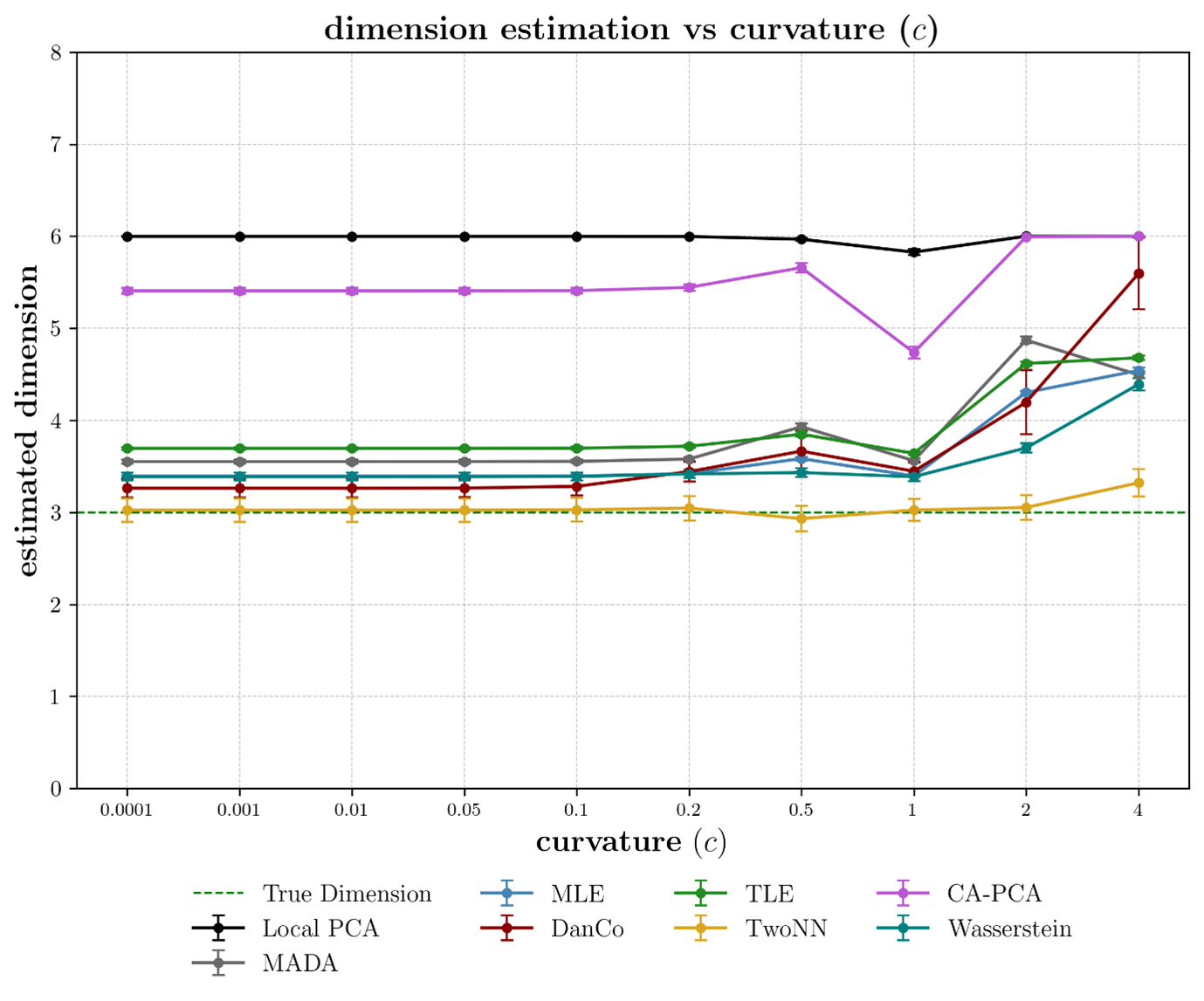}
        \caption{Dimension Estimate vs Curvature ($c$)}
        \label{fig:curvature03}
    \end{subfigure}
    \caption{Dimension estimate versus curvature ($d$, left; $c$, right).}
\end{figure}

The results of this experiment are presented in Figure~\ref{fig:curvature02}.

We observe that three estimators, namely \textsc{Local PCA}, \textsc{CA-PCA}, and \textsc{DanCo}, are quite robust to increasing the intrinsic dimension. In contrast, the remaining five estimators exhibit increasing bias with respect to $d$, consistently underestimating the true dimension.

As previously discussed, \textsc{DanCo} is specifically designed to address the underestimation problem in high-dimensional settings, leveraging an estimated KL divergence. The two PCA-based methods likely benefit from their simplicity and minimal reliance on strong modeling assumptions: since PCA aims to estimate the tangent space directly under the flatness assumption, it can achieve reliable results with as few as $d$ data points.

Our final experiment on curvature, described in Table~\ref{tab:expc03}, examines the effect of curvature variation along the manifold.
\begin{table}[htbp]
\centering
\caption{Experiment design to test the effect of curvature variation.}
\label{tab:expc03}
\begin{tabular}{@{} l p{0.65\linewidth} @{}}
\toprule
\textbf{Component} & \textbf{Description} \\
\midrule
Manifold $M$       & a $3$-dimensional deformed sphere in $\mathbb{R}^{6}$ \\
Sample          & $n=1,000$ data points uniformly distributed on $M$\\
Hyperparameter  & $K = 100$ ($K=10$ for \textsc{DanCo}), $\alpha = 5$ \\
Factor          & $c\in\{0.0001,0.001,0.01,0.05,0.1,0.2,0.5,1,2,4\}$ \\
\bottomrule
\end{tabular}
\end{table}

For this purpose, we use a deformed sphere as the underlying manifold, whose curvature complexity increases globally with the parameter $c$ (see Figure~\ref{fig:deformed} in \ref{sec:TheoreticalBackground} for details). We expect the performance of the estimators to deteriorate as $c$ increases, which is confirmed by the results shown in Figure~\ref{fig:curvature03}.

Notably, the estimates from \textsc{Local PCA} and \textsc{CA-PCA} degrade more significantly than those of other estimators for the given neighborhood size, suggesting a potential weakness when dealing with nonlinear embeddings. However, \textsc{CA-PCA} still outperforms \textsc{Local PCA} in this setting.

We can conjecture that when $c = 1$, the geometry of the resulting manifold happens to align well with the assumptions or mechanisms built into the \textsc{CA-PCA} estimator, as well as the parameter setting, leading to the improved observed performance.

\subsubsection{Effect of noise}

It is quite unlikely that all data points in a sample of data lie exactly on some underlying manifold. A more plausible scenario is that they lie close to the manifold instead. A natural model for this setting assumes that data points originate from the manifold and are subsequently perturbed by some noise, which we assume for simplicity to be $p$-dimensional Gaussian $N_p(\bm{0}, \sigma^2I_p)$ in our experiment.

Almost all state-of-the-art manifold dimension estimators are designed under the assumption of a noiseless sample. Consequently, they are often not well-suited for application to noisy datasets, despite demonstrating strong performance on noise-free examples. The presence of noise significantly complicates the manifold dimension estimation problem, as even small perturbations can easily overwhelm the local structure upon which many estimators rely, and increasing sample size $n$ alone will not necessary be helpful.

The experiment benchmarking the estimators' performance against varying Gaussian noise levels is described in Table~\ref{tab:expnoise}. 
\begin{table}[htbp]
\centering
\caption{Experiment design to test the effect of noise.}
\label{tab:expnoise}
\begin{tabular}{@{} l p{0.65\linewidth} @{}}
\toprule
\textbf{Component} & \textbf{Description} \\
\midrule
Manifold $M$        & a $5$-dimensional unit sphere linearly embedded in $\mathbb{R}^{10}$ \\
Sample          & $n=1,000$ data points uniformly distributed on $M$, with additive Gaussian noise from $N(\bm{0}, \sigma^2I_{10})$ \\
Hyperparameter  & $K = 100$ ($K=10$ for \textsc{DanCo}), $\alpha = 5$ \\
Factor          & $\sigma\in\{0.00,0.01,\ldots,0.09\}$\\
\bottomrule
\end{tabular}
\end{table}

\begin{figure}[htbp]
    \centering
    \begin{subfigure}[b]{0.45\textwidth}
        \centering        \includegraphics[width=\textwidth]{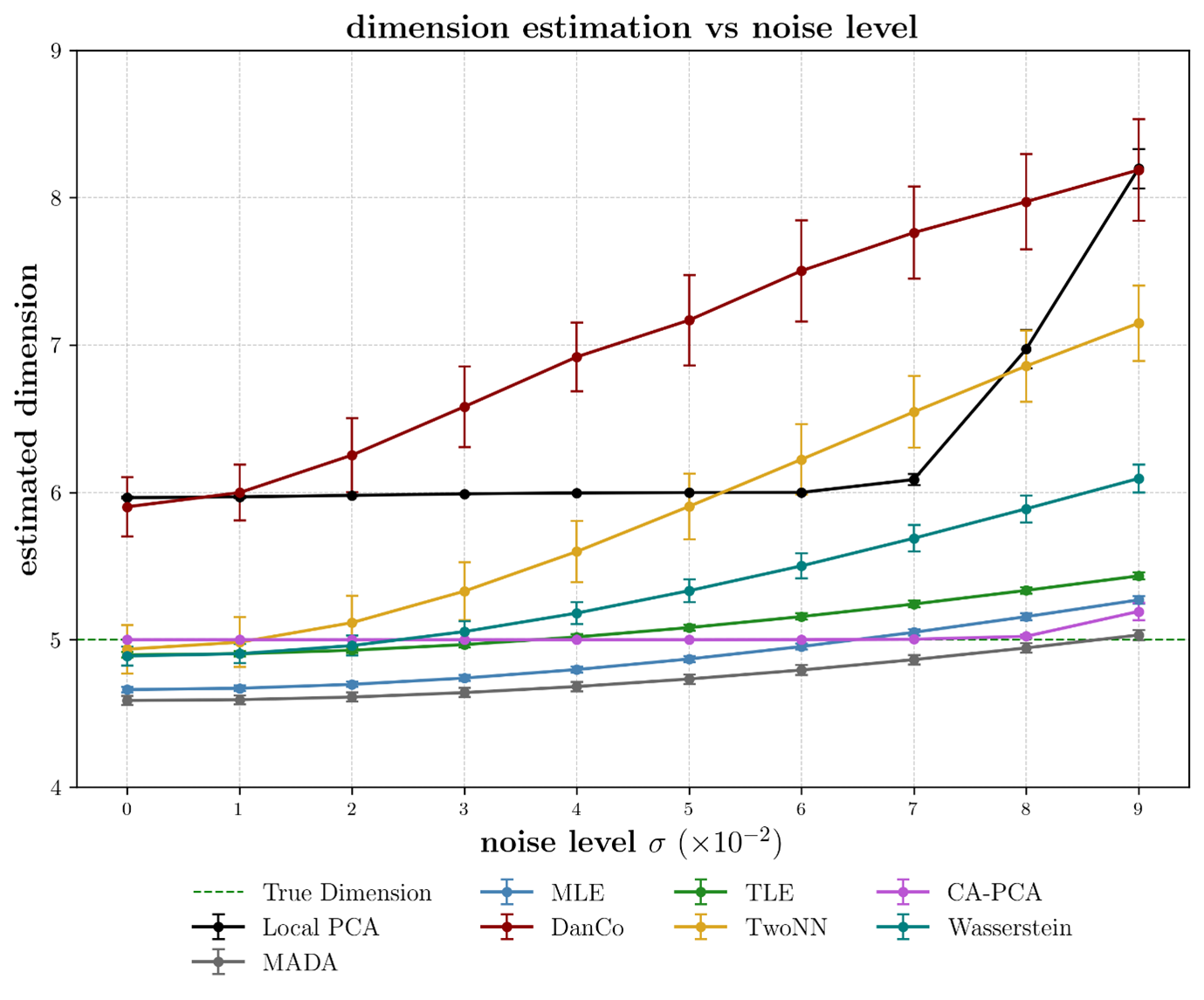}
        \caption{Dimension Estimate vs Noise Level ($\sigma^2$)}
        \label{fig:noise01}
    \end{subfigure}
    \hfill
        \begin{subfigure}[b]{0.45\textwidth}
        \centering
        \includegraphics[width=\textwidth]{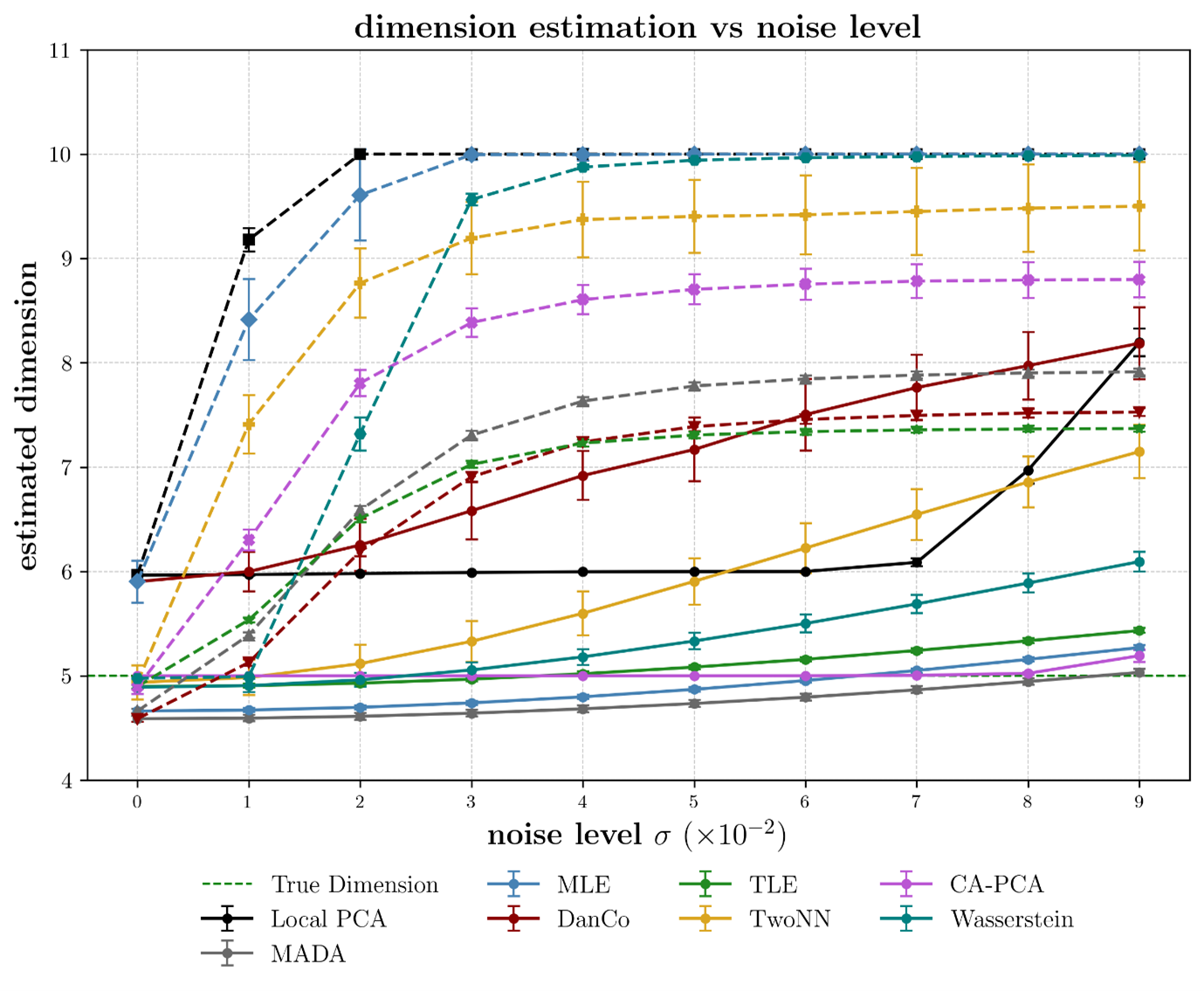}
        \caption{Noise Effect under High and Low Curvature.}
        \label{fig:noise02}
    \end{subfigure}
    \caption{Dimension vs Noise (left) and Noise-Curvature Effect (right).}
\end{figure}

The results are shown in Figure~\ref{fig:noise01}. 

We observe that as the noise level increases, the estimates from all estimators tend to increase, and their performance deteriorates in terms of both bias and variance. This indicates that the estimators become less effective at capturing the true underlying structure under higher noise conditions.

In particular, \textsc{TwoNN} and \textsc{DanCo}, which have been top performers in noiseless scenarios, exhibit greater sensitivity to noise compared to the other methods. This highlights the challenging nature of manifold dimension estimation in noisy environments.

\begin{figure}[htbp]
    \centering
    \begin{subfigure}[b]{0.35\textwidth}
        \centering        \includegraphics[width=\textwidth]{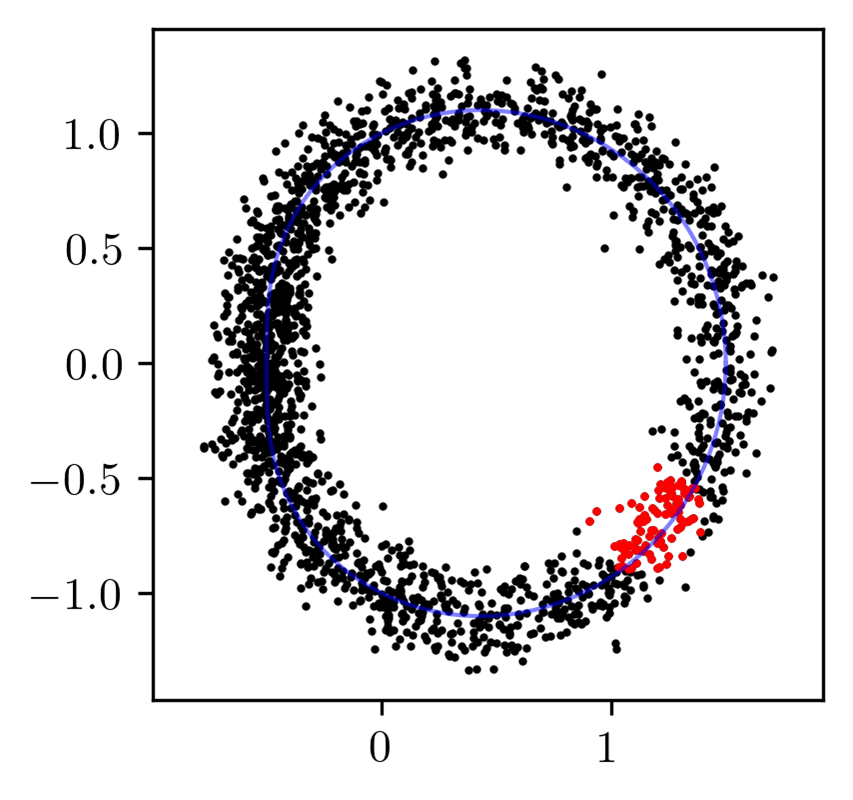}
        \caption{}
        \label{fig:nc01}
    \end{subfigure}
    \hfill
        \begin{subfigure}[b]{0.35\textwidth}
        \centering
        \includegraphics[width=\textwidth]{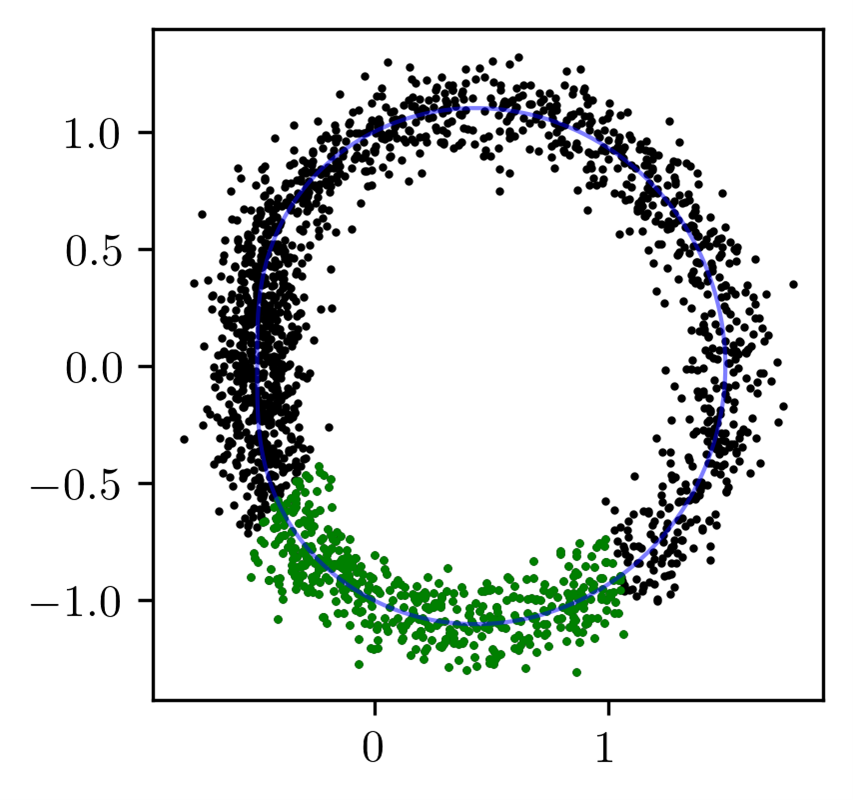}
        \caption{}
        \label{fig:nc02}
    \end{subfigure}
    \caption{Demonstration of Noise-Curvature Dilemma: A small neighborhood can be overwhelmed by noise and fail to capture the true underlying geometry, while a large neighborhood is more robust to noise but will jeopardize the flatness assumption more.}
\end{figure}

We have illustrated in Figure~\ref{fig:curvature01} that the size of curvature alone is not a driver of the performance of manifold dimension estimators. It usually manifests itself when combined with other factors, and noise is one such example. The combination of curvature and noise leads to the so-called \emph{noise curvature dilemma}. We have seen the choice of neighborhood size is important for most flatness-based estimators. For the flatness assumption to hold, the neighborhood size $K$ should be small compared to the sample size $n$, particularly when the manifold exhibits high curvature. However, when Gaussian noise is present, using a very small neighborhood may result in the noise to dominate any meaningful local structure, and we have to look at a large enough neighborhoods instead to not be overwhelmed, hence the dilemma, which is illustrated in Figure~\ref{fig:nc01}--\ref{fig:nc02}.

In Figure~\ref{fig:noise02}, we superimpose on Figure~\ref{fig:noise01} the estimated curves obtained using a sphere with a smaller radius (and thus larger curvature), while keeping all other conditions identical. The new estimates are shown as dashed curves.

We observe that when the data is noiseless, curvature size does not affect the estimates. However, as the noise level increases, estimates on manifolds with higher curvature deteriorate more rapidly in terms of both bias and variance compared to those on manifolds with lower curvature.

To mitigate the noise-curvature dilemma, some researchers have proposed shifting the focus away from recovering the true intrinsic dimension and instead estimating a dimension that is sufficient to describe the data at the chosen neighborhood scale. This notion, often referred to as the \emph{effective dimension}, lacks a standardized definition but has gained attention as a practical alternative in noisy or high-curvature scenarios \citep{wang2008scale, little2009multiscale}.

\begin{figure}[htbp]
    \centering
    \begin{subfigure}[b]{0.35\textwidth}
        \centering
        \includegraphics[width=\textwidth]{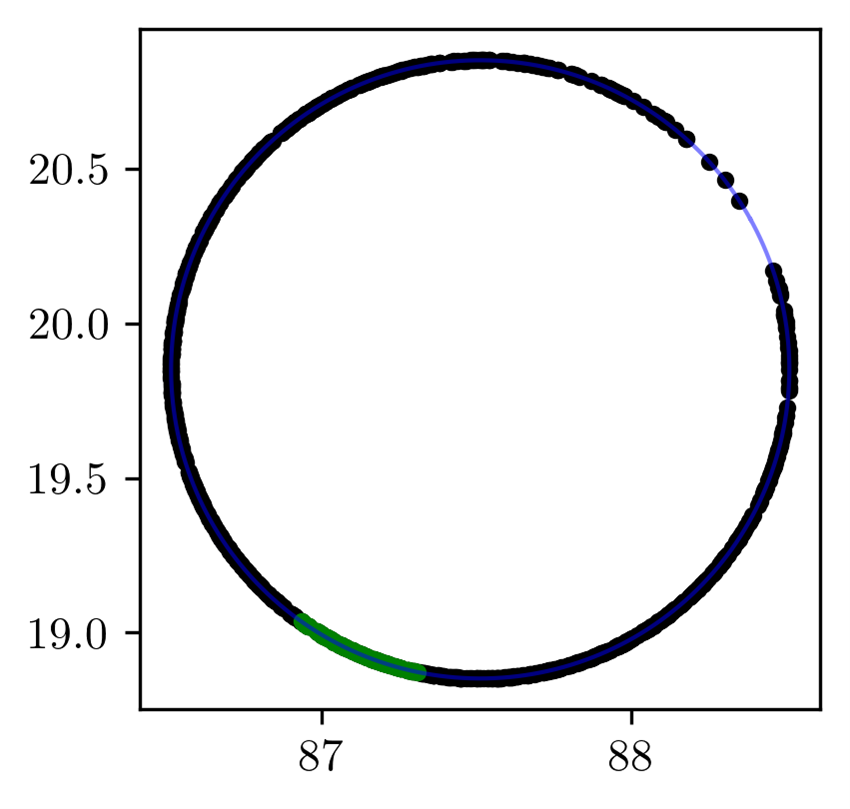}
        \caption{}
        \label{fig:nonuniform01}
    \end{subfigure}
    \hfill
        \begin{subfigure}[b]{0.35\textwidth}
        \centering
        \includegraphics[width=\textwidth]{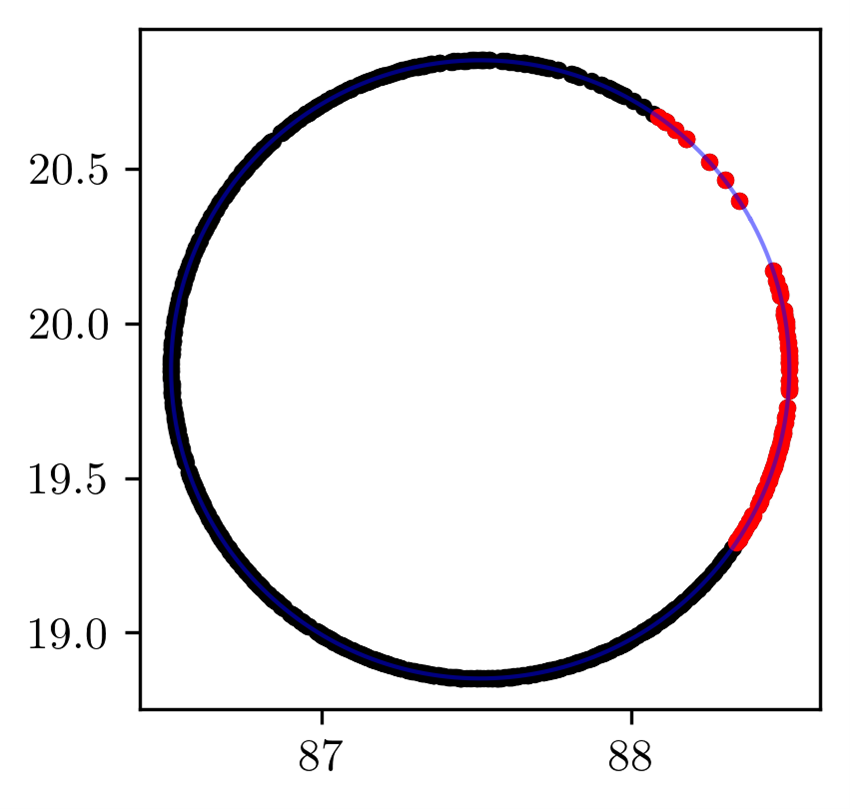}
        \caption{}
        \label{fig:nonuniform02}
    \end{subfigure}
    \caption{Effect of Data Distribution on Effective Neighborhood Sizes: A neighborhood taken from a region with high data density (green) is more local and thus better approximated by a linear space. In contrast, a neighborhood from a low-density region (red) spans a larger area, violating the local flatness assumption to a greater extent.}
\end{figure}

Another factor worth consideration is the data distribution on the manifold. Since most estimators assume at least local uniformity of the data, a highly irregular distribution is expected to also have significant impacts on their performance. For example, as illustrated in Figure~\ref{fig:nonuniform01}--\ref{fig:nonuniform02}, the flatness assumption is clearly violated in the neighborhood where the data density is low. 

To better understand estimator robustness, it is more informative to compare their performance under both uniform and nonuniform distributions across a diverse set of manifolds. We defer this experiment to the next  section, where we conduct a more systematic comparison of manifold dimension estimators across varying data distributions and manifold geometries.

\subsection{Comparative Evaluation of Estimators}

\subsubsection{Choosing hyperparameter values}

As we have shown, the performance of a manifold dimension estimator can vary significantly depending on its hyperparameter values, and the optimal setting may differ across experimental setups. Therefore, when comparing estimators across various manifolds, it is crucial to tune each estimator's hyperparameters appropriately for each sample. However, in practice, the true intrinsic dimension is unknown, necessitating an algorithmic approach for per-sample hyperparameter tuning.

Motivated by the patterns observed in Figure~\ref{fig:nbh01}, we propose an algorithm for selecting hyperparameter values that yield stable estimator performance. The procedure identifies regions of the hyperparameter space where the standard deviation of the estimates is minimized or when the mean estimate remains approximately constant. If no such stable region is detected, the algorithm defaults to reporting the average across all candidate hyperparameter values. The full algorithm is given below\footnote{The window size and standard deviation threshold in Algorithm~\ref{alg:hyperparameter} are chosen heuristically and can be refined in future work.}.

\begin{algorithm}[htbp]
\caption{Choosing a Value of the Hyperparameter $K$ (or $\alpha$)}
\label{alg:hyperparameter}
\begin{algorithmic}[1]
\State \textbf{Input:} Dimension estimates $\hat{d}_{\ell}$ for $\ell = 1, \dots, k_{\max}$, corresponding to hyperparameters $K_{\ell}$
\State \textbf{Output:} Indices $k_1$ and $k_2$, defining the selected stable range for $K$
\State Initialize $k_1 \gets 1$, $k_2 \gets 3$, $k^* \gets 0$
\State Initialize $s_{\min} \gets \infty$, $s_{\max} \gets 0$
\For{$k = 1$ \textbf{to} $k_{\max} - 2$}
    \State $s \gets$ standard deviation of $\hat{d}_k, \hat{d}_{k+1}, \hat{d}_{k+2}$
    \If{$s < s_{\min}$}
        \State $s_{\min} \gets s$
        \State $k^* \gets k$
    \EndIf
    \If{$s > s_{\max}$}
        \State $s_{\max} \gets s$
    \EndIf
\EndFor
\If{$s_{\max} > 1.25 s_{\min}$}
    \State $k_1 \gets k^*$, $k_2 \gets k^* + 2$
\Else
    \State $k_1 \gets 1$, $k_2 \gets k_{\max}$
\EndIf
\State \Return $k_1, k_2$
\end{algorithmic}
\end{algorithm}

\subsubsection{Experimental setup}

We generate 100 random samples of size $n$ (both $n=500$ and $n=2,000$ are considered) on 18 manifolds. Each manifold has a specific intrinsic dimension $d$ (ranging from $\{1,2,3,5,10,20\}$), and a distinct geometry: sphere, ball, Gaussian density surface, deformed sphere, cylinder, helix, Swiss roll, Mobius strip, torus, or hyperbolic surface. The 18 manifolds are detailed in Table~\ref{tab:manifolds}. For each manifold, points are sampled  uniformly and also according to a skewed Beta distribution in the coordinate space, following \citet{campadelli2015intrinsic} as a natural way to model realistic nonuniformity. Both noiseless and noisy settings are considered, the latter obtained by adding the Gaussian noise $N_p(\boldsymbol{0}, 0.01I_p)$. 

These 18 manifolds cover a wide variety of cases, including not only standard toy manifolds commonly found in textbooks ($M_5$ to $M_{10}$; see \citet{lee2003smooth} for example), but also more challenging scenarios such as high-dimensional settings ($M_{11}$ to $M_{33}$) and nonlinear embeddings ($M_{41}$ to $M_{43}$).

Although many of these manifolds can be embedded in $\mathbb{R}^{d+1}$, we further apply a linear embedding into $\mathbb{R}^{2d}$ to increase the potential for estimation error and to evaluate the robustness of the dimension estimators to this factor. 

\begin{table}[htbp]
\centering
\caption{The 18 manifolds considered.}
\label{tab:manifolds}
\begin{tabular}{@{} l p{11cm} @{}}
\toprule
\textbf{Manifold} & \textbf{Description} \\
\midrule
$M_{11}$         & A $5$-dimensional sphere embedded in $\R^{10}$ with $R = 1$. \\
$M_{12}$     & A $10$-dimensional sphere embedded in $\R^{20}$ with $R = 1$. \\
$M_{13}$          & A $20$-dimensional sphere embedded in $\R^{40}$ with $R = 1$. \\
$M_{21}$          & A $5$-dimensional ball embedded in $\R^{10}$ with $R = 1$. \\
$M_{22}$        & A $10$-dimensional ball embedded in $\R^{20}$ with $R = 1$. \\
$M_{23}$          & A $20$-dimensional ball embedded in $\R^{40}$ with $R = 1$. \\
$M_{31}$           & A $5$-dimensional Gaussian density surface embedded in $\R^{10}$ corresponding to $N(0, 0.01I_5)$. \\
$M_{32}$       & A $10$-dimensional Gaussian density surface embedded in $\R^{20}$ corresponding to $N(0, 0.01I_{10})$. \\
$M_{33}$   & A $20$-dimensional Gaussian density surface embedded in $\R^{40}$ corresponding to $N(0, 0.01I_{20})$. \\
$M_{41}$   & A $3$-dimensional deformed sphere with $c = 0.01$. \\
$M_{42}$         & A $3$-dimensional deformed sphere with $c = 0.1$. \\
$M_{43}$   & A $3$-dimensional deformed sphere with $c = 1$. \\
$M_{5}$     & A $2$-dimensional cylinder embedded in $\R^4$. \\
$M_{6}$   & A $1$-dimensional helix embedded in $\R^3$. \\
$M_{7}$ & A $2$-dimensional Swiss roll embedded in $\R^4$. \\
$M_{8}$      & A $2$-dimensional Mobius strip embedded in $\R^4$. \\
$M_{9}$     & A $2$-dimensional torus embedded in $\R^4$. \\
$M_{10}$         & A $2$-dimensional hyperbolic surface embedded in $\R^4$. \\
\bottomrule
\end{tabular}
\end{table}

\subsubsection{Analysis of the results}

The results are presented in six numerical tables, each comprising 18 rows (corresponding to the manifolds under study) and 8 columns (corresponding to the dimension estimators). Each cell reports the mean and standard deviation computed from 100 replications. For ease of presentation, the complete set of tables is relegated to Appendix~B. In what follows, we restrict attention to the principal findings, which are examined through a detailed analysis of these tables.

Tables~\ref{tab:02} and~\ref{tab:03} present results on uniformly distributed, noiseless samples with respective sample sizes \( n = 500 \) and \( n = 2,000 \). Beyond what is already known from individual factor analyses, we uncover the following insights.

Despite being excluded from many prior comparative studies, \textsc{Local PCA} consistently delivers strong performance when $n=2,000$, achieving the smallest bias and variance across almost all manifolds. However, it struggles when $n=500$, particularly on manifolds exhibiting complex curvature such as \( M_{41} \) to \( M_{43} \).
    
\textsc{CA-PCA} further enhances this performance, not only when $n=500$ but also on nonlinearly embedded manifolds like \( M_{41} \) to \( M_{43} \). When $n=2,000$, it achieves near-unbiased estimates with zero variance, highlighting the strength of the PCA procedure.

Manifolds characterized by high curvature or complex geometry, such as \( M_{41} \) to \( M_{43} \), present challenges for nearly all estimators when $n=500$. While estimation improves when $n=2,000$, substantial bias remains for most methods.

High-dimensional manifolds tend to have their dimension underestimated. This trend, already observed with spheres previously, is further confirmed by results on balls and Gaussian density surfaces. Notable exceptions include \textsc{Local PCA}, \textsc{CA-PCA}, and \textsc{DanCo}, which handle high-dimensional cases better.

\textsc{DanCo} and \textsc{TwoNN} are also competitive. When $n=500$, \textsc{TwoNN} achieves the best overall performance on low-dimensional manifolds but tends to underestimate dimensions in high-dimensional settings. \textsc{DanCo} performs slightly worse than \textsc{TwoNN} in low dimensions, in particular the deformed spheres, but gets better with high-dimensional manifolds. As the sample size increases from $n=500$ to $n=2,000$, \textsc{TwoNN} shows greater improvement, becoming nearly unbiased on the manifolds where it performs well. However, it continues to face challenges with high-dimensional cases.

The remaining estimators, \textsc{MLE}, \textsc{MADA}, \textsc{TLE}, and \textsc{Wasserstein}, can be classified as bottom-tier. They all tend to underestimate the dimension of high-dimensional manifolds and struggle with nonlinearly embedded manifolds. Even for simple manifolds with $n=2,000$, a relatively large bias still persists.

Tables~\ref{tab:04} and~\ref{tab:05} present the experimental results under nonuniform data distributions, where nonuniformity is introduced by sampling from skewed Beta distributions in the parameter space. Our findings are summarized below.

The bias of all estimators increases under nonuniform sampling, especially for high-dimensional manifolds. However, an interesting exception occurs with $M_{41}$ to $M_{43}$, where reduced bias is observed. We hypothesize that this is due to the complex curvature of these manifolds, that is, curvature varies significantly across different regions. Under nonuniform sampling, data tends to concentrate in one or a few high-density regions, potentially skipping areas with high curvature. This effectively simplifies the estimation task. Since we do not fully control the nonuniformity, such outcomes can occur.

With nonuniform data distributions, the underestimation problem for high-dimensional spheres and balls becomes significantly more severe. This time, even \textsc{Local PCA}, \textsc{DanCo}, and \textsc{CA-PCA} exhibit underestimation, although \textsc{CA-PCA} underestimates significantly less than \textsc{Local PCA}. For the PCA-based methods, this underestimation is intuitive: with nonuniform sampling, a neighborhood may include data points that are far from the center; see Figure~\ref{fig:nonuniform02} for example. The principal components corresponding to these distant points can dominate the local structure and pull down the estimated dimension.

Despite the increased difficulty introduced by nonuniformity, the best-performing estimators remain \textsc{Local PCA}, \textsc{CA-PCA}, \textsc{DanCo}, and \textsc{TwoNN}. However, in this setting, they perform well primarily on low-dimensional manifolds and also fail noticeably on high-dimensional ones.

Lastly, when additive Gaussian noise is present, as shown in Tables~\ref{tab:06} and~\ref{tab:07}, the performance of all manifold dimension estimators deteriorates, primarily in the form of increased bias. In particular, our previous top performers (\textsc{Local PCA}, \textsc{DanCo}, \textsc{TwoNN} and \textsc{CA-PCA}) exhibit the most significant degradation. Conversely, the previous underdogs (\textsc{MLE}, \textsc{MADA}, \textsc{TLE} and \textsc{Wasserstein}) do not degrade as severely. A possible explanation is that these estimators do not rely on geometric information as delicate as the others and therefore are less overwhelmed by noise. 

Notably, increasing the sample size does not lead to performance improvement in the presence of noise. We also observe an apparent decrease in bias for high-dimensional manifolds compared to the noiseless case for most estimators; however, this reduction appears to be an artifact of the noise rather than a genuine improvement. This is evidenced by \textsc{TwoNN}, whose bias shifts from negative to positive as the sample size increases. For the same reason, although \textsc{MADA} appears to perform best on most manifolds with low to moderate dimensions, we believe this may not accurately reflect its true effectiveness.

\subsection{Real-World Data Examples}\label{sec-realdata}

In this section, we further assess the performance of the eight estimators on three real-world datasets which are commonly used to benchmark manifold dimension estimators, all of which are widely believed to satisfy the manifold hypothesis. Our experiments were conducted in a similar manner as for the synthetic datasets, hyperparameters are selected according to Algorithm~\ref{alg:hyperparameter}. 

Since all three of our real-world datasets have ambient dimension much larger than the intrinsic dimension, as discussed in Section~\ref{sec:manifold_dimension_estimators}, \textsc{DanCo} is expected to struggle. Indeed, our experiments with \textsc{DanCo} produced dimensionality estimates significantly higher than expected (often close to the ambient dimension). These discrepancies led to results that differ substantially from those reported in prior studies such as \citet{CERUTI20142569} and \citet{campadelli2015intrinsic}, which in fact used the mechanisms we introduced before, and thus no longer reflect pure \textsc{DanCo} outputs. Consequently, we cite their reported estimates (in brackets) for comparative purposes. All real-world datasets are fed for estimation directly without centering or scaling.

\subsubsection{The \texttt{ISOMAP} Dataset}

\begin{figure}[htbp]
    \centering

    \begin{subfigure}[t]{0.14\textwidth}
        \includegraphics[width=\linewidth]{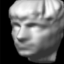}
    \end{subfigure}
    \begin{subfigure}[t]{0.14\textwidth}
        \includegraphics[width=\linewidth]{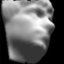}
    \end{subfigure}
    \begin{subfigure}[t]{0.14\textwidth}
        \includegraphics[width=\linewidth]{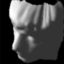}
    \end{subfigure}
    \begin{subfigure}[t]{0.14\textwidth}
        \includegraphics[width=\linewidth]{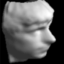}
    \end{subfigure}
    \begin{subfigure}[t]{0.14\textwidth}
        \includegraphics[width=\linewidth]{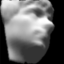}
    \end{subfigure}
    \begin{subfigure}[t]{0.14\textwidth}
        \includegraphics[width=\linewidth]{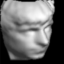}
    \end{subfigure}

    \vspace{2mm} 

    \begin{subfigure}[t]{0.14\textwidth}
        \includegraphics[width=\linewidth]{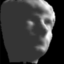}
    \end{subfigure}
    \begin{subfigure}[t]{0.14\textwidth}
        \includegraphics[width=\linewidth]{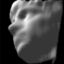}
    \end{subfigure}
    \begin{subfigure}[t]{0.14\textwidth}
        \includegraphics[width=\linewidth]{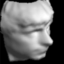}
    \end{subfigure}
    \begin{subfigure}[t]{0.14\textwidth}
        \includegraphics[width=\linewidth]{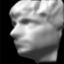}
    \end{subfigure}
    \begin{subfigure}[t]{0.14\textwidth}
        \includegraphics[width=\linewidth]{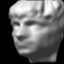}
    \end{subfigure}
    \begin{subfigure}[t]{0.14\textwidth}
        \includegraphics[width=\linewidth]{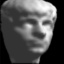}
    \end{subfigure}

    \caption{Sample images from \texttt{ISOMAP} dataset.}
    \label{fig:faces}
\end{figure}

The \texttt{ISOMAP} dataset \citep{balasubramanian2002isomap} consists of 698 grayscale images of a face sculpture, each of size $64 \times 64$, as shown in Figure~\ref{fig:faces}. 

It is widely believed that the vectorized dataset has an intrinsic dimension of $d = 3$, corresponding to the two degrees of freedom for camera position and one for lighting condition.

Table~\ref{tab:realdata_ISOMAP} reports the estimation results.

\renewcommand{\arraystretch}{0.8}
\begin{table}[htbp]
\centering
\caption{Dimension Estimate on \texttt{ISOMAP}}
\label{tab:realdata_ISOMAP}
\begin{tabular}{ *{9}{c} @{}}
\toprule
 \ & \textsc{Local PCA} & \textsc{MADA} & \textsc{MLE} & \textsc{DanCo}  & \textsc{TLE} & \textsc{TwoNN} & \textsc{CA-PCA} & \textsc{Wass.}  \\
\midrule
\texttt{Python} & $10.05$ & $5.86$ & $4.34$ & ($4.00$)  & $4.70$ & $3.49$ & $25.71$ & $4.13$  \\
\bottomrule
\end{tabular}
\end{table}

\subsubsection{The \texttt{MNIST} Dataset}

\begin{figure}[htbp]
    \centering
    \includegraphics[width=0.90\textwidth]{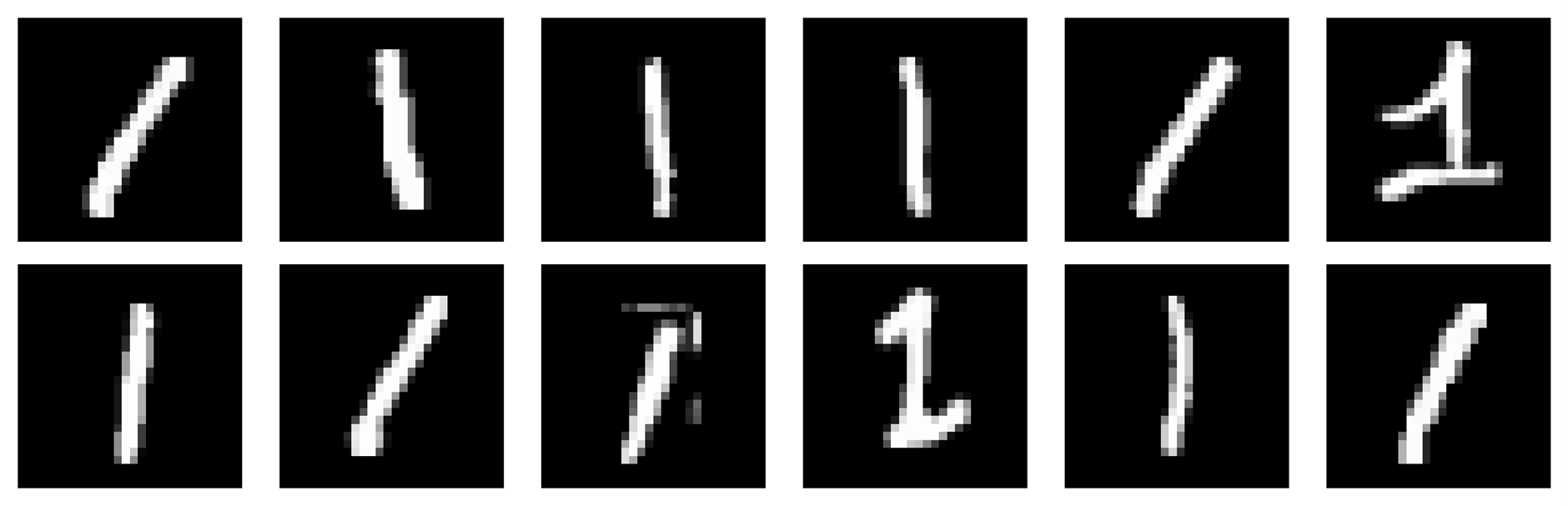}
    \caption{Sample images from \texttt{MNIST} dataset.}
    \label{fig:digit}
\end{figure}

The now well known \texttt{MNIST} dataset \citep{lecun2002gradient} contains images of handwritten digits from $0$ to $9$. From this, we extract 6,742 grayscale images of digit $1$, each of size $28 \times 28$, as shown in Figure~\ref{fig:digit}. 

The manifold hypothesis is believed to hold for this subset, with an intrinsic dimension estimated to lie between 8 and 11.

Table~\ref{tab:realdata_MNIST} reports the estimation results.

\renewcommand{\arraystretch}{0.8}
\begin{table}[htbp]
\centering
\caption{Dimension Estimate on \texttt{MNIST}}
\label{tab:realdata_MNIST}
\begin{tabular}{ *{9}{c} @{}}
\toprule
 \ & \textsc{Local PCA} & \textsc{MADA} & \textsc{MLE} & \textsc{DanCo}  & \textsc{TLE} & \textsc{TwoNN} & \textsc{CA-PCA} & \textsc{Wass.}  \\
\midrule
 \texttt{Python} & $15.41$ & $8.67$ & $8.97$ & ($9.98$)  & $7.92$ & $12.98$ & $84.66$ & $9.49$  \\
\bottomrule
\end{tabular}
\end{table}

\subsubsection{The \texttt{ISOLET} Dataset}

\begin{figure}[H]
    \centering
    \includegraphics[width=0.95\textwidth]{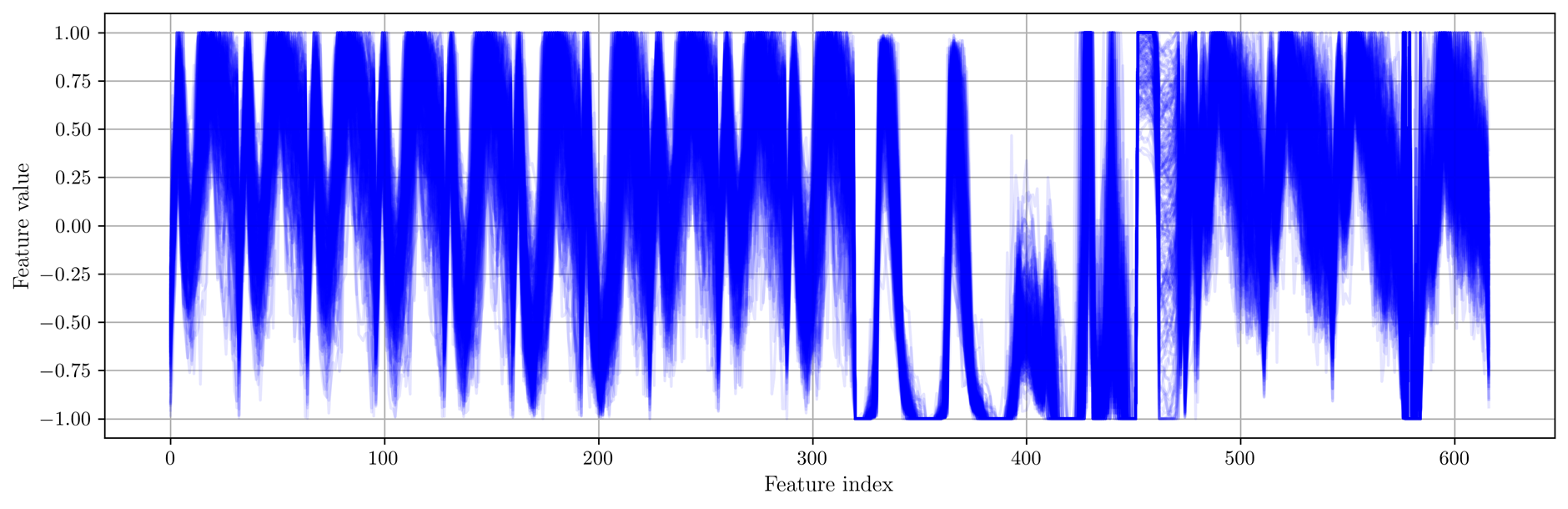}
    \caption{Frequency characteristics for letter ``A" from \texttt{ISOLET} dataset.}
    \label{fig:isolet}
\end{figure}

The \texttt{ISOLET} dataset \citep{asuncion2007uci} includes 6,240 vectors of length 616. Each vector represents frequency characteristics extracted from audio recordings of individuals pronouncing letters from the English alphabet, as illustrated in Figure~\ref{fig:isolet}. 

The manifold hypothesis is again believed to apply, with an estimated intrinsic dimension between 16 and 22.

Table~\ref{tab:realdata_ISOLET} reports the estimation results.

\renewcommand{\arraystretch}{0.8}
\begin{table}[htbp]
\centering
\caption{Dimension Estimate on \texttt{ISOLET}}
\label{tab:realdata_ISOLET}
\begin{tabular}{ *{9}{c} @{}}
\toprule
 \ & \textsc{Local PCA} & \textsc{MADA} & \textsc{MLE} & \textsc{DanCo}  & \textsc{TLE} & \textsc{TwoNN} & \textsc{CA-PCA} & \textsc{Wass.}  \\
\midrule
 \texttt{Python} & $34.44$ & $13.63$ & $14.29$ & ($19.00$)  & $12.58$ & $9.11$ & $54.22$ & $13.38$  \\
\bottomrule
\end{tabular}
\end{table}

\subsubsection{Analysis of the results}

Among all estimators, (the adapted) \textsc{DanCo} produces estimates within the expected ranges for all three datasets. While \textsc{MLE}, \textsc{MADA}, \textsc{TLE}, \textsc{TwoNN}, and \textsc{Wasserstein} perform well on the \texttt{ISOMAP} and \texttt{MNIST} datasets, they underestimate the dimension of \texttt{ISOLET}, which is known to have higher intrinsic dimensionality. This behavior is consistent with our findings on the synthetic datasets.

In contrast, \textsc{Local PCA} and \textsc{CA-PCA}, despite being top performers on simulated data, perform considerably worse on these real-world datasets. This is likely because the underlying manifolds are embedded in spaces with much higher ambient dimensionality. Since both methods rely, at least partly, on identifying the number of significant eigenvalues, the accumulation of many small (but nonzero) eigenvalues can mislead the estimators and degrade their performance. This highlights the potential limitations of applying the two methods in practice.

\section{Conclusion}\label{sec-conclusion}

In this article, we consolidate the core theoretical foundations underlying manifold dimension estimation, including manifold geometry and sampling mechanisms, and present a principled survey of representative estimators. Building on this framework, we conduct the most up-to-date and comprehensive empirical evaluation to date, featuring extensive controlled experiments that systematically isolate the effects of key factors and enable fair, meaningful comparisons across methods, thereby yielding clear insights into estimator behavior and performance trade-offs.

Perhaps the most valuable takeaway is for practitioners. Although an estimator's performace can depend on  various factors such as hyperparameter, noise level, curvature complexity, and intrinsic dimension range, our experiments identify \textsc{CA-PCA} and \textsc{TwoNN} as the overall top performers, making them strong default choices. However, caution is warranted: \textsc{TwoNN} tends to underestimate the dimension when the true intrinsic dimensionality is high, while \textsc{CA-PCA} may overestimate when the manifold is nonlinearly embedded in a high-dimensional ambient space. When these two estimators produce substantially different values, and computational cost is not a major constraint, we recommend considering \textsc{DanCo} for a third opinion. We also suggest exploring a wide range of neighborhood sizes for the estimators to help identify stable estimates.

Our experimental results lead to a perhaps unsurprising conclusion: for a problem of this complexity, simpler methods often perform better. Given the flexibility of the manifold model, any strong assumption made by an estimator about the underlying structure is more likely to be violated than satisfied. For researchers developing new manifold dimension estimators, we recommend moving beyond the extensively exploited flatness assumption. Instead, focusing on methods that explicitly incorporate curvature and are robust to practical challenges such as noise and non-uniform sampling may yield more broadly applicable and reliable results.

\section*{Acknowledgements}

This research includes computations using the computational cluster Katana supported by Research Technology Services at UNSW Sydney.

This paper is part of Z.~Bi's Ph.D.\ thesis.

\section*{Disclosure statement}\label{disclosure-statement}

The authors do not have any conflicts of interest to declare.

\section*{Data Availability Statement}\label{data-availability-statement}

All code and data used in this research is available through our  \href{https://github.com/loong-bi/manifold-dimension-estimation/tree/main}{Github page}.

\appendix
\section{Mathematical claim}\label{sec:mathclaim}


\noindent\emph{Claim.} A $d$-dimensional deformed sphere $M$ fills the entire $\mathbb{R}^{2d}$; in other words, it is not contained in any proper linear subspace.

\noindent\emph{Proof.}
It suffices to show the collection of tangent vectors of $M$ spans $\R^{2d}$. From the definition of $M$, it has tangent vectors in the form of 
\eqn{
\bm{v}_j(\bm{u}) = \textrm{d}\varphi^{-1}|_{\bm{u}}(\bm{e}_j) \in \R^{2d} \text{ with }\bm{u \in }(-\pi, \pi)^d,
}
where $\{\bm{e}_j\}_{j=1}^{d}$ are standard basis of $\mathbb{R}^{d}$. The components of $\bm{v}_j(\bm{u})$ satisfy
\[
\left\{
\begin{aligned}
v_{jj}(\bm{u}) &=  - 2\pi c r\sin(2c\pi u_j)\cos(2\pi u_j) - 2\pi[R + r\cos(2c\pi u_j)]\sin(2\pi u_j), \\
v_{j,j + d}(\bm{u}) &=  - 2\pi c r\sin(2c\pi u_j)\sin(2\pi u_j) + 2\pi[R + r\cos(2c\pi u_j)]\cos(2\pi u_j)
\end{aligned}
\right.
\]
for $j = 1, \cdots, d$ and $v_{ji}(\bm{u}) = v_{j,i + d}(\bm{u}) = 0$ for $i \neq j$. When $2c$ is an integer, we take $2d$ tangent vectors to be $\bm{v}_j(\pm\bm{e}_j)$ for $j = 1, \cdots, d$. Otherwise, we set $\bm{v}_j(\pm\bm{e}_j / 2)$. In both cases, one can verify that the resulting $2d$ tangent vectors are linearly independent. This establishes that the tangent vectors span $\mathbb{R}^{2d}$, and the proof is complete.
\qed

\section{Tables of results}\label{sec:TablesResults}


\begin{landscape}
{\small
\setlength{\tabcolsep}{2pt}
\begin{longtable}{@{} l| *{8}{c} @{}}
\caption{Mean dimension estimates ($\pm$ standard deviation) for 18 manifolds (Table~\ref{tab:manifolds}) with true dimension $d$, based on 100 replicates of $n=500$ uniform samples. Bold indicates the estimate with minimal MSE.} \label{tab:02} \\
\toprule
\textbf{Manifold} ($d$) & \textsc{Local PCA} & \textsc{MADA} & \textsc{MLE} & \textsc{DanCo}  & \textsc{TLE} & \textsc{TwoNN}  & \textsc{CA-PCA} & \textsc{Wasserstein} \\
\midrule
\endfirsthead
\multicolumn{9}{c}%
{\tablename\ \thetable\ -- \textit{continued from previous page}} \\
\toprule
\textbf{Manifold} ($d$) & \textsc{Local PCA} & \textsc{MADA} & \textsc{MLE} & \textsc{DanCo}  & \textsc{TLE} & \textsc{TwoNN}  & \textsc{CA-PCA} & \textsc{Wasserstein} \\
\midrule
\endhead
\midrule \multicolumn{9}{r}{\textit{Continued on next page}} \\
\endfoot
\bottomrule
\endlastfoot
$M_{11} (5)$ & $\phantom{1}5.03  \pm  0.01$ & $\phantom{1}4.65  \pm  0.05$ & $\phantom{1}4.78  \pm  0.04$ & $\phantom{1}6.18  \pm  0.26$ & $\phantom{1}5.06  \pm  0.04$ & $\phantom{1}4.90  \pm  0.28$ & $\phantom{1}\bm{5.00  \pm  0.00}$ & $\phantom{1}4.88  \pm  0.08$ \\
$M_{12} (10)$ & \bm{$10.10 \pm 0.01$} & $\phantom{1}7.93  \pm  0.10$ & $\phantom{1}8.25  \pm  0.07$ & $11.16 \pm 0.37$ & $\phantom{1}8.48  \pm  0.07$ & $\phantom{1}9.17  \pm  0.53$ & $10.21 \pm 0.02$ & $\phantom{1}8.67  \pm  0.16$ \\
$M_{13} (20)$ & $19.76 \pm 0.02$ & $12.89 \pm 0.15$ & $13.54 \pm 0.10$ & \bm{$20.13 \pm 0.37$} & $13.33 \pm 0.10$ & $15.90 \pm 0.90$ & $20.68 \pm 0.02$ & $14.73 \pm 0.35$ \\
$M_{21} (5)$ & $\phantom{1}\bm{5.00  \pm  0.00}$ & $\phantom{1}4.24  \pm  0.05$ & $\phantom{1}4.41  \pm  0.04$ & $\phantom{1}4.97  \pm  0.13$ & $\phantom{1}4.58  \pm  0.04$ & $\phantom{1}4.64  \pm  0.30$ & $\phantom{1}4.91  \pm  0.03$ & $\phantom{1}4.52  \pm  0.07$ \\
$M_{22} (10)$ & $\phantom{1}\bm{9.99  \pm  0.01}$ & $\phantom{1}7.45  \pm  0.08$ & $\phantom{1}7.75  \pm  0.06$ & $10.03 \pm 0.17$ & $\phantom{1}7.88  \pm  0.05$ & $\phantom{1}8.79  \pm  0.51$ & $\phantom{1}9.93  \pm  0.03$ & $\phantom{1}8.28  \pm  0.26$ \\
$M_{23} (20)$ & $19.10 \pm 0.04$ & $12.42 \pm 0.15$ & $13.05 \pm 0.12$ & \bm{$20.01 \pm 0.28$} & $12.79 \pm 0.11$ & $15.53 \pm 0.92$ & $19.93 \pm 0.02$ & $14.44 \pm 0.50$ \\
$M_{31} (5)$ & $\phantom{1}\bm{5.00  \pm  0.00}$ & $\phantom{1}4.21  \pm  0.06$ & $\phantom{1}4.38  \pm  0.06$ & $\phantom{1}4.97  \pm  0.13$ & $\phantom{1}4.54  \pm  0.05$ & $\phantom{1}4.55  \pm  0.27$ & $\phantom{1}5.00  \pm  0.01$ & $\phantom{1}4.40  \pm  0.07$ \\
$M_{32} (10)$ & \bm{$10.00 \pm 0.00$} & $\phantom{1}7.29  \pm  0.08$ & $\phantom{1}7.55  \pm  0.07$ & $\phantom{1}9.55  \pm  0.39$ & $\phantom{1}7.56  \pm  0.06$ & $\phantom{1}8.28  \pm  0.51$ & $10.01 \pm 0.02$ & $\phantom{1}7.82  \pm  0.15$ \\
$M_{33} (20)$ & $19.43 \pm 0.04$ & $12.04 \pm 0.13$ & $12.69 \pm 0.10$ & $19.59 \pm 0.54$ & $12.14 \pm 0.09$ & $14.75 \pm 0.83$ & \bm{$20.07 \pm 0.05$} & $13.66 \pm 0.28$ \\
$M_{41} (3)$ & $\phantom{1}5.83  \pm  0.02$ & $\phantom{1}3.53  \pm  0.04$ & $\phantom{1}3.36  \pm  0.03$ & $\phantom{1}3.42  \pm  0.12$ & $\phantom{1}3.67  \pm  0.03$ & $\phantom{1}\bm{3.05  \pm  0.17}$ & $\phantom{1}3.24  \pm  0.02$ & $\phantom{1}3.54  \pm  0.05$ \\
$M_{42} (3)$ & $\phantom{1}5.83  \pm  0.02$ & $\phantom{1}3.53  \pm  0.04$ & $\phantom{1}3.36  \pm  0.03$ & $\phantom{1}3.44  \pm  0.12$ & $\phantom{1}3.67  \pm  0.03$ & $\phantom{1}\bm{3.05  \pm  0.17}$ & $\phantom{1}3.25  \pm  0.02$ & $\phantom{1}3.55  \pm  0.05$ \\
$M_{43} (3)$ & $\phantom{1}5.52  \pm  0.05$ & $\phantom{1}3.45  \pm  0.04$ & $\phantom{1}3.36  \pm  0.03$ & $\phantom{1}3.58  \pm  0.14$ & $\phantom{1}3.63  \pm  0.03$ & $\phantom{1}\bm{3.06  \pm  0.16}$ & $\phantom{1}3.44  \pm  0.03$ & $\phantom{1}3.58  \pm  0.04$ \\
$M_{5} (2)$ & $\phantom{1}\bm{2.00  \pm  0.00}$ & $\phantom{1}1.86  \pm  0.03$ & $\phantom{1}1.91  \pm  0.02$ & $\phantom{1}2.16  \pm  0.03$ & $\phantom{1}2.01  \pm  0.02$ & $\phantom{1}1.98  \pm  0.12$ & $\phantom{1}1.95  \pm  0.01$ & $\phantom{1}2.35  \pm  0.06$ \\
$M_{6} (1)$ & $\phantom{1}2.75  \pm  0.04$ & $\phantom{1}1.11  \pm  0.02$ & $\phantom{1}1.09  \pm  0.01$ & $\phantom{1}1.01  \pm  0.02$ & $\phantom{1}1.21  \pm  0.01$ & $\phantom{1}1.00  \pm  0.07$ & $\phantom{1}\bm{1.00  \pm  0.00}$ & $\phantom{1}2.21  \pm  0.05$ \\
$M_{7} (2)$ & $\phantom{1}2.99  \pm  0.01$ & $\phantom{1}2.96  \pm  0.07$ & $\phantom{1}2.14  \pm  0.03$ & $\phantom{1}2.18  \pm  0.03$ & $\phantom{1}2.30  \pm  0.02$ & $\phantom{1}\bm{1.97  \pm  0.12}$ & $\phantom{1}2.78  \pm  0.04$ & $\phantom{1}2.52  \pm  0.14$ \\
$M_{8} (2)$ & $\phantom{1}\bm{2.00  \pm  0.00}$ & $\phantom{1}1.83  \pm  0.02$ & $\phantom{1}1.88  \pm  0.01$ & $\phantom{1}2.18  \pm  0.03$ & $\phantom{1}1.97  \pm  0.01$ & $\phantom{1}1.96  \pm  0.12$ & $\phantom{1}1.93  \pm  0.01$ & $\phantom{1}2.29  \pm  0.07$ \\
$M_{9} (2)$ & $\phantom{1}3.00  \pm  0.00$ & $\phantom{1}2.34  \pm  0.03$ & $\phantom{1}2.19  \pm  0.02$ & $\phantom{1}2.16  \pm  0.03$ & $\phantom{1}2.35  \pm  0.02$ & $\phantom{1}2.00  \pm  0.13$ & $\phantom{1}\bm{2.00  \pm  0.01}$ & $\phantom{1}2.15  \pm  0.11$ \\
$M_{10} (2)$ & $\phantom{1}2.07  \pm  0.01$ & $\phantom{1}2.11  \pm  0.03$ & $\phantom{1}2.11  \pm  0.02$ & $\phantom{1}2.18  \pm  0.03$ & $\phantom{1}2.14  \pm  0.01$ & $\phantom{1}\bm{1.99  \pm  0.12}$ & $\phantom{1}1.95  \pm  0.01$ & $\phantom{1}2.44  \pm  0.07$ \\
\end{longtable}}
\end{landscape}

\begin{landscape}
{\small
\setlength{\tabcolsep}{2pt}
\begin{longtable}{@{} l| *{8}{c} @{}}
\caption{Mean dimension estimates ($\pm$ standard deviation) for 18 manifolds (Table~\ref{tab:manifolds}) with true dimension $d$, based on 100 replicates of $n=2{,}000$ uniform samples. Bold indicates the estimate with minimal MSE.} \label{tab:03} \\
\toprule
\textbf{Manifold} ($d$) & \textsc{Local PCA} & \textsc{MADA} & \textsc{MLE} & \textsc{DanCo}  & \textsc{TLE} & \textsc{TwoNN}  & \textsc{CA-PCA} & \textsc{Wasserstein} \\
\midrule
\endfirsthead
\multicolumn{9}{c}%
{\tablename\ \thetable\ -- \textit{continued from previous page}} \\
\toprule
\textbf{Manifold} ($d$) & \textsc{Local PCA} & \textsc{MADA} & \textsc{MLE} & \textsc{DanCo}  & \textsc{TLE} & \textsc{TwoNN}  & \textsc{CA-PCA} & \textsc{Wasserstein} \\
\midrule
\endhead
\midrule \multicolumn{9}{r}{\textit{Continued on next page}} \\
\endfoot
\bottomrule
\endlastfoot
$M_{11}(5)$ & $\phantom{1}5.00  \pm  0.00$ & $\phantom{1}4.86  \pm  0.03$ & $\phantom{1}4.92  \pm  0.02$ & $\phantom{1}6.54  \pm  0.17$ & $\phantom{1}5.15  \pm  0.02$ & $\phantom{1}4.97  \pm  0.14$ & $\phantom{1}\bm{5.00  \pm  0.00}$ & $\phantom{1}4.89  \pm  0.04$ \\
$M_{12}(10)$ & $10.01 \pm 0.00$ & $\phantom{1}8.68  \pm  0.05$ & $\phantom{1}8.89  \pm  0.04$ & $10.92 \pm 0.14$ & $\phantom{1}9.20  \pm  0.04$ & $\phantom{1}9.40  \pm  0.24$ & \bm{$10.00 \pm 0.00$} & $\phantom{1}9.06  \pm  0.15$ \\
$M_{13}(20)$ & $19.70 \pm 0.01$ & $14.61 \pm 0.08$ & $15.12 \pm 0.06$ & $19.97 \pm 0.24$ & $15.13 \pm 0.06$ & $16.74 \pm 0.52$ & \bm{$20.17 \pm 0.01$} & $15.52 \pm 0.14$ \\
$M_{21}(5)$ & $\phantom{1}\bm{5.00  \pm  0.00}$ & $\phantom{1}4.54  \pm  0.03$ & $\phantom{1}4.62  \pm  0.02$ & $\phantom{1}5.00  \pm  0.06$ & $\phantom{1}4.76  \pm  0.02$ & $\phantom{1}4.76  \pm  0.14$ & $\phantom{1}4.93  \pm  0.01$ & $\phantom{1}4.56  \pm  0.03$ \\
$M_{22}(10)$ & $\phantom{1}9.99  \pm  0.00$ & $\phantom{1}8.20  \pm  0.05$ & $\phantom{1}8.41  \pm  0.04$ & \bm{$10.00 \pm 0.11$} & $\phantom{1}8.61  \pm  0.04$ & $\phantom{1}8.96  \pm  0.22$ & $\phantom{1}9.91  \pm  0.02$ & $\phantom{1}8.45  \pm  0.08$ \\
$M_{23}(20)$ & $19.00 \pm 0.03$ & $14.09 \pm 0.08$ & $14.59 \pm 0.06$ & \bm{$19.98 \pm 0.15$} & $14.51 \pm 0.05$ & $16.17 \pm 0.43$ & $19.79 \pm 0.03$ & $14.94 \pm 0.13$ \\
$M_{31}(5)$ & $\phantom{1}\bm{5.00  \pm  0.00}$ & $\phantom{1}4.41  \pm  0.02$ & $\phantom{1}4.50  \pm  0.02$ & $\phantom{1}5.11  \pm  0.08$ & $\phantom{1}4.62  \pm  0.02$ & $\phantom{1}4.68  \pm  0.13$ & $\phantom{1}\bm{5.00  \pm  0.00}$ & $\phantom{1}4.42  \pm  0.03$ \\
$M_{32}(10)$ & \bm{$10.00 \pm 0.00$} & $\phantom{1}7.86  \pm  0.05$ & $\phantom{1}8.09  \pm  0.04$ & $\phantom{1}9.71  \pm  0.31$ & $\phantom{1}8.17  \pm  0.03$ & $\phantom{1}8.61  \pm  0.25$ & \bm{$10.00 \pm 0.00$} & $\phantom{1}7.99  \pm  0.07$ \\
$M_{33}(20)$ & $19.38 \pm 0.02$ & $13.65 \pm 0.07$ & $14.09 \pm 0.05$ & $20.87 \pm 0.37$ & $13.71 \pm 0.05$ & $15.41 \pm 0.38$ & \bm{$19.99 \pm 0.01$} & $14.37 \pm 0.14$ \\
$M_{41}(3)$ & $\phantom{1}3.11  \pm  0.01$ & $\phantom{1}3.22  \pm  0.02$ & $\phantom{1}3.19  \pm  0.01$ & $\phantom{1}2.99  \pm  0.05$ & $\phantom{1}3.40  \pm  0.01$ & $\phantom{1}3.03  \pm  0.09$ & $\phantom{1}\bm{3.00  \pm  0.00}$ & $\phantom{1}3.30  \pm  0.02$ \\
$M_{42}(3)$ & $\phantom{1}3.12  \pm  0.01$ & $\phantom{1}3.22  \pm  0.02$ & $\phantom{1}3.19  \pm  0.01$ & $\phantom{1}3.01  \pm  0.05$ & $\phantom{1}3.41  \pm  0.01$ & $\phantom{1}3.03  \pm  0.09$ & $\phantom{1}\bm{3.00  \pm  0.00}$ & $\phantom{1}3.30  \pm  0.02$ \\
$M_{43}(3)$ & $\phantom{1}3.17  \pm  0.01$ & $\phantom{1}3.23  \pm  0.02$ & $\phantom{1}3.20  \pm  0.01$ & $\phantom{1}3.76  \pm  0.09$ & $\phantom{1}3.39  \pm  0.01$ & $\phantom{1}3.03  \pm  0.09$ & $\phantom{1}\bm{3.01  \pm  0.00}$ & $\phantom{1}3.28  \pm  0.02$ \\
$M_{5}(2)$ & $\phantom{1}\bm{2.00  \pm  0.00}$ & $\phantom{1}1.95  \pm  0.01$ & $\phantom{1}1.97  \pm  0.01$ & $\phantom{1}2.14  \pm  0.02$ & $\phantom{1}2.03  \pm  0.01$ & $\phantom{1}1.97  \pm  0.05$ & $\phantom{1}1.97  \pm  0.00$ & $\phantom{1}2.26  \pm  0.06$ \\
$M_{6}(1)$ & $\phantom{1}\bm{1.00  \pm  0.00}$ & $\phantom{1}1.01  \pm  0.01$ & $\phantom{1}1.02  \pm  0.00$ & $\phantom{1}1.00  \pm  0.00$ & $\phantom{1}1.10  \pm  0.00$ & $\phantom{1}1.00  \pm  0.03$ & $\phantom{1}\bm{1.00  \pm  0.00}$ & $\phantom{1}2.09  \pm  0.04$ \\
$M_{7}(2)$ & $\phantom{1}2.03  \pm  0.00$ & $\phantom{1}\bm{1.99  \pm  0.01}$ & $\phantom{1}\bm{1.99  \pm  0.01}$ & $\phantom{1}2.16  \pm  0.01$ & $\phantom{1}2.09  \pm  0.01$ & $\phantom{1}1.97  \pm  0.06$ & $\phantom{1}1.98  \pm  0.01$ & $\phantom{1}2.44  \pm  0.03$ \\
$M_{8}(2)$ & $\phantom{1}\bm{2.00  \pm  0.00}$ & $\phantom{1}1.94  \pm  0.01$ & $\phantom{1}1.97  \pm  0.01$ & $\phantom{1}2.15  \pm  0.01$ & $\phantom{1}2.02  \pm  0.01$ & $\phantom{1}1.97  \pm  0.05$ & $\phantom{1}1.96  \pm  0.00$ & $\phantom{1}1.70  \pm  0.10$ \\
$M_{9}(2)$ & $\phantom{1}\bm{2.00  \pm  0.00}$ & $\phantom{1}2.09  \pm  0.01$ & $\phantom{1}2.08  \pm  0.01$ & $\phantom{1}2.15  \pm  0.02$ & $\phantom{1}2.19  \pm  0.01$ & $\phantom{1}2.00  \pm  0.06$ & $\phantom{1}\bm{2.00  \pm  0.00}$ & $\phantom{1}2.29  \pm  0.10$ \\
$M_{10}(2)$ & $\phantom{1}\bm{2.00  \pm  0.00}$ & $\phantom{1}2.06  \pm  0.01$ & $\phantom{1}2.05  \pm  0.01$ & $\phantom{1}2.16  \pm  0.01$ & $\phantom{1}2.10  \pm  0.01$ & $\phantom{1}1.98  \pm  0.06$ & $\phantom{1}1.98  \pm  0.00$ & $\phantom{1}2.31  \pm  0.05$ \\
\end{longtable}}
\end{landscape}

\begin{landscape}
{\small
\setlength{\tabcolsep}{2pt}
\begin{longtable}{@{} l| *{8}{c} @{}}
\caption{Mean dimension estimates ($\pm$ standard deviation) for 18 manifolds (Table~\ref{tab:manifolds}) with true dimension $d$, based on 100 replicates of $n=500$ nonuniform samples. Bold indicates the estimate with minimal MSE.} \label{tab:04} \\
\toprule
\textbf{Manifold} ($d$) & \textsc{Local PCA} & \textsc{MADA} & \textsc{MLE} & \textsc{DanCo}  & \textsc{TLE} & \textsc{TwoNN}  & \textsc{CA-PCA} & \textsc{Wasserstein} \\
\midrule
\endfirsthead
\multicolumn{9}{c}%
{\tablename\ \thetable\ -- \textit{continued from previous page}} \\
\toprule
\textbf{Manifold} ($d$) & \textsc{Local PCA} & \textsc{MADA} & \textsc{MLE} & \textsc{DanCo}  & \textsc{TLE} & \textsc{TwoNN}  & \textsc{CA-PCA} & \textsc{Wasserstein} \\
\midrule
\endhead
\midrule \multicolumn{9}{r}{\textit{Continued on next page}} \\
\endfoot
\bottomrule
\endlastfoot
$M_{11}(5)$ & $\phantom{1}\bm{4.94  \pm  0.02}$ & $\phantom{1}4.52  \pm  0.07$ & $\phantom{1}4.71  \pm  0.06$ & $\phantom{1}5.69  \pm  0.15$ & $\phantom{1}4.21  \pm  0.06$ & $\phantom{1}4.91  \pm  0.25$ & $\phantom{1}5.23  \pm  0.04$ & $\phantom{1}4.59  \pm  0.14$ \\
$M_{12}(10)$ & $\phantom{1}5.79  \pm  0.11$ & $\phantom{1}4.60  \pm  0.09$ & $\phantom{1}4.85  \pm  0.08$ & $\phantom{1}6.60  \pm  0.19$ & $\phantom{1}4.23  \pm  0.06$ & $\phantom{1}5.87  \pm  0.31$ & $\phantom{1}\bm{9.10  \pm  0.32}$ & $\phantom{1}4.94  \pm  0.17$ \\
$M_{13}(20)$ & $\phantom{1}5.89  \pm  0.12$ & $\phantom{1}4.59  \pm  0.10$ & $\phantom{1}4.83  \pm  0.10$ & $\phantom{1}6.58  \pm  0.19$ & $\phantom{1}4.22  \pm  0.07$ & $\phantom{1}5.73  \pm  0.35$ & \bm{$15.75 \pm 0.76$} & $\phantom{1}4.84  \pm  0.21$ \\
$M_{21}(5)$ & $\phantom{1}\bm{4.99  \pm  0.01}$ & $\phantom{1}4.81  \pm  0.08$ & $\phantom{1}4.95  \pm  0.06$ & $\phantom{1}6.04  \pm  0.14$ & $\phantom{1}4.17  \pm  0.05$ & $\phantom{1}5.01  \pm  0.24$ & $\phantom{1}5.48  \pm  0.06$ & $\phantom{1}4.77  \pm  0.14$ \\
$M_{22}(10)$ & $\phantom{1}5.98  \pm  0.10$ & $\phantom{1}4.96  \pm  0.12$ & $\phantom{1}5.20  \pm  0.11$ & $\phantom{1}6.46  \pm  0.16$ & $\phantom{1}4.23  \pm  0.06$ & $\phantom{1}6.00  \pm  0.31$ & \bm{$10.34 \pm 0.34$} & $\phantom{1}4.79  \pm  0.33$ \\
$M_{23}(20)$ & $\phantom{1}5.97  \pm  0.12$ & $\phantom{1}4.94  \pm  0.11$ & $\phantom{1}5.17  \pm  0.09$ & $\phantom{1}6.64  \pm  0.18$ & $\phantom{1}4.23  \pm  0.06$ & $\phantom{1}5.99  \pm  0.33$ & \bm{$18.32 \pm 0.80$} & $\phantom{1}4.75  \pm  0.34$ \\
$M_{31}(5)$ & $\phantom{1}\bm{5.00  \pm  0.00}$ & $\phantom{1}4.57  \pm  0.06$ & $\phantom{1}4.76  \pm  0.05$ & $\phantom{1}5.50  \pm  0.17$ & $\phantom{1}4.75  \pm  0.04$ & $\phantom{1}4.90  \pm  0.30$ & $\phantom{1}5.05  \pm  0.02$ & $\phantom{1}4.72  \pm  0.10$ \\
$M_{32}(10)$ & \bm{$10.00 \pm 0.00$} & $\phantom{1}7.44  \pm  0.07$ & $\phantom{1}7.90  \pm  0.07$ & $10.04 \pm 0.27$ & $\phantom{1}7.60  \pm  0.05$ & $\phantom{1}8.91  \pm  0.42$ & $10.38 \pm 0.07$ & $\phantom{1}8.16  \pm  0.19$ \\
$M_{33}(20)$ & \bm{$18.81 \pm 0.06$} & $11.84 \pm 0.13$ & $12.72 \pm 0.11$ & $21.74 \pm 1.29$ & $11.76 \pm 0.12$ & $15.31 \pm 0.89$ & $21.74 \pm 0.30$ & $13.77 \pm 0.38$ \\
$M_{41}(3)$ & $\phantom{1}4.03  \pm  0.07$ & $\phantom{1}3.17  \pm  0.04$ & $\phantom{1}3.20  \pm  0.03$ & $\phantom{1}3.40  \pm  0.12$ & $\phantom{1}3.31  \pm  0.03$ & $\phantom{1}\bm{3.01  \pm  0.19}$ & $\phantom{1}2.95  \pm  0.01$ & $\phantom{1}3.33  \pm  0.08$ \\
$M_{42}(3)$ & $\phantom{1}4.05  \pm  0.07$ & $\phantom{1}3.17  \pm  0.04$ & $\phantom{1}3.20  \pm  0.03$ & $\phantom{1}3.40  \pm  0.12$ & $\phantom{1}3.31  \pm  0.03$ & $\phantom{1}\bm{3.01  \pm  0.19}$ & $\phantom{1}2.95  \pm  0.01$ & $\phantom{1}3.33  \pm  0.08$ \\
$M_{43}(3)$ & $\phantom{1}3.88  \pm  0.06$ & $\phantom{1}3.04  \pm  0.04$ & $\phantom{1}3.15  \pm  0.03$ & $\phantom{1}3.38  \pm  0.12$ & $\phantom{1}3.22  \pm  0.03$ & $\phantom{1}3.02  \pm  0.20$ & $\phantom{1}\bm{3.00  \pm  0.01}$ & $\phantom{1}3.34  \pm  0.07$ \\
$M_{5}(2)$ & $\phantom{1}\bm{2.00  \pm  0.00}$ & $\phantom{1}1.83  \pm  0.02$ & $\phantom{1}1.89  \pm  0.01$ & $\phantom{1}2.23  \pm  0.05$ & $\phantom{1}1.93  \pm  0.01$ & $\phantom{1}2.00  \pm  0.13$ & $\phantom{1}1.88  \pm  0.01$ & $\phantom{1}2.33  \pm  0.08$ \\
$M_{6}(1)$ & $\phantom{1}1.61  \pm  0.08$ & $\phantom{1}1.43  \pm  0.15$ & $\phantom{1}1.15  \pm  0.02$ & $\phantom{1}\bm{1.00  \pm  0.00}$ & $\phantom{1}1.22  \pm  0.01$ & $\phantom{1}0.99  \pm  0.06$ & $\phantom{1}1.03  \pm  0.01$ & $\phantom{1}2.23  \pm  0.05$ \\
$M_{7}(2)$ & $\phantom{1}2.59  \pm  0.04$ & $\phantom{1}2.67  \pm  0.11$ & $\phantom{1}2.29  \pm  0.05$ & $\phantom{1}2.19  \pm  0.04$ & $\phantom{1}2.28  \pm  0.02$ & $\phantom{1}\bm{1.99  \pm  0.11}$ & $\phantom{1}2.13  \pm  0.02$ & $\phantom{1}2.40  \pm  0.15$ \\
$M_{8}(2)$ & $\phantom{1}\bm{2.00  \pm  0.00}$ & $\phantom{1}1.70  \pm  0.03$ & $\phantom{1}1.78  \pm  0.02$ & $\phantom{1}2.26  \pm  0.05$ & $\phantom{1}1.83  \pm  0.02$ & $\phantom{1}2.00  \pm  0.13$ & $\phantom{1}1.86  \pm  0.02$ & $\phantom{1}2.23  \pm  0.08$ \\
$M_{9}(2)$ & $\phantom{1}2.69  \pm  0.05$ & $\phantom{1}2.20  \pm  0.04$ & $\phantom{1}2.17  \pm  0.02$ & $\phantom{1}2.20  \pm  0.03$ & $\phantom{1}2.22  \pm  0.02$ & $\phantom{1}\bm{1.99  \pm  0.12}$ & $\phantom{1}1.98  \pm  0.01$ & $\phantom{1}2.07  \pm  0.15$ \\
$M_{10}(2)$ & $\phantom{1}2.02  \pm  0.01$ & $\phantom{1}1.94  \pm  0.03$ & $\phantom{1}2.02  \pm  0.02$ & $\phantom{1}2.22  \pm  0.04$ & $\phantom{1}\bm{2.00  \pm  0.01}$ & $\phantom{1}2.01  \pm  0.12$ & $\phantom{1}1.93  \pm  0.01$ & $\phantom{1}2.38  \pm  0.08$ \\
\end{longtable}
}
\end{landscape}

\begin{landscape}
{\small
\setlength{\tabcolsep}{2pt}
\begin{longtable}{@{} l| *{8}{c} @{}}
\caption{Mean dimension estimates ($\pm$ standard deviation) for 18 manifolds (Table~\ref{tab:manifolds}) with true dimension $d$, based on 100 replicates of $n=2{,}000$ nonuniform samples. Bold indicates the estimate with minimal MSE.} \label{tab:05} \\
\toprule
\textbf{Manifold} ($d$) & \textsc{Local PCA} & \textsc{MADA} & \textsc{MLE} & \textsc{DanCo}  & \textsc{TLE} & \textsc{TwoNN}  & \textsc{CA-PCA} & \textsc{Wasserstein} \\
\midrule
\endfirsthead
\multicolumn{9}{c}%
{\tablename\ \thetable\ -- \textit{continued from previous page}} \\
\toprule
\textbf{Manifold} ($d$) & \textsc{Local PCA} & \textsc{MADA} & \textsc{MLE} & \textsc{DanCo}  & \textsc{TLE} & \textsc{TwoNN}  & \textsc{CA-PCA} & \textsc{Wasserstein} \\
\midrule
\endhead
\midrule \multicolumn{9}{r}{\textit{Continued on next page}} \\
\endfoot
\bottomrule
\endlastfoot
$M_{11}(5)$ & $\phantom{1}\bm{4.99  \pm  0.00}$ & $\phantom{1}4.98  \pm  0.03$ & $\phantom{1}5.03  \pm  0.03$ & $\phantom{1}6.63  \pm  0.14$ & $\phantom{1}4.68  \pm  0.02$ & $\phantom{1}5.01  \pm  0.14$ & $\phantom{1}5.06  \pm  0.01$ & $\phantom{1}4.71  \pm  0.06$ \\
$M_{12}(10)$ & $\phantom{1}6.42  \pm  0.06$ & $\phantom{1}5.46  \pm  0.06$ & $\phantom{1}5.71  \pm  0.05$ & $\phantom{1}6.97  \pm  0.06$ & $\phantom{1}5.05  \pm  0.03$ & $\phantom{1}6.42  \pm  0.19$ & $\phantom{1}\bm{8.29  \pm  0.16}$ & $\phantom{1}5.54  \pm  0.09$ \\
$M_{13}(20)$ & $\phantom{1}6.41  \pm  0.06$ & $\phantom{1}5.42  \pm  0.05$ & $\phantom{1}5.66  \pm  0.04$ & $\phantom{1}6.94  \pm  0.06$ & $\phantom{1}5.03  \pm  0.03$ & $\phantom{1}6.36  \pm  0.19$ & \bm{$11.44 \pm 0.26$} & $\phantom{1}5.47  \pm  0.10$ \\
$M_{21}(5)$ & $\phantom{1}\bm{5.00  \pm  0.00}$ & $\phantom{1}5.17  \pm  0.04$ & $\phantom{1}5.18  \pm  0.02$ & $\phantom{1}6.38  \pm  0.13$ & $\phantom{1}4.67  \pm  0.02$ & $\phantom{1}5.05  \pm  0.14$ & $\phantom{1}5.14  \pm  0.02$ & $\phantom{1}4.80  \pm  0.07$ \\
$M_{22}(10)$ & $\phantom{1}6.60  \pm  0.07$ & $\phantom{1}5.80  \pm  0.07$ & $\phantom{1}6.03  \pm  0.06$ & $\phantom{1}7.01  \pm  0.04$ & $\phantom{1}5.10  \pm  0.04$ & $\phantom{1}6.58  \pm  0.17$ & $\phantom{1}\bm{9.19  \pm  0.14}$ & $\phantom{1}5.90  \pm  0.20$ \\
$M_{23}(20)$ & $\phantom{1}6.51  \pm  0.06$ & $\phantom{1}5.74  \pm  0.06$ & $\phantom{1}5.97  \pm  0.05$ & $\phantom{1}6.99  \pm  0.04$ & $\phantom{1}5.08  \pm  0.03$ & $\phantom{1}6.48  \pm  0.18$ & \bm{$13.84 \pm 0.31$} & $\phantom{1}5.71  \pm  0.08$ \\
$M_{31}(5)$ & $\phantom{1}\bm{5.00  \pm  0.00}$ & $\phantom{1}4.75  \pm  0.03$ & $\phantom{1}4.89  \pm  0.02$ & $\phantom{1}5.59  \pm  0.09$ & $\phantom{1}4.89  \pm  0.02$ & $\phantom{1}4.94  \pm  0.15$ & $\phantom{1}5.01  \pm  0.00$ & $\phantom{1}4.73  \pm  0.04$ \\
$M_{32}(10)$ & \bm{$10.00 \pm 0.00$} & $\phantom{1}8.32  \pm  0.05$ & $\phantom{1}8.66  \pm  0.04$ & $10.02 \pm 0.06$ & $\phantom{1}8.47  \pm  0.03$ & $\phantom{1}9.20  \pm  0.27$ & $10.09 \pm 0.02$ & $\phantom{1}8.49  \pm  0.09$ \\
$M_{33}(20)$ & $18.79 \pm 0.02$ & $13.70 \pm 0.07$ & $14.43 \pm 0.06$ & $21.77 \pm 0.86$ & $13.55 \pm 0.06$ & $16.21 \pm 0.42$ & \bm{$20.47 \pm 0.07$} & $14.71 \pm 0.18$ \\
$M_{41}(3)$ & $\phantom{1}3.02  \pm  0.00$ & $\phantom{1}3.13  \pm  0.02$ & $\phantom{1}3.12  \pm  0.01$ & $\phantom{1}3.18  \pm  0.06$ & $\phantom{1}3.21  \pm  0.01$ & $\phantom{1}3.01  \pm  0.08$ & $\phantom{1}\bm{3.00  \pm  0.00}$ & $\phantom{1}3.18  \pm  0.04$ \\
$M_{42}(3)$ & $\phantom{1}3.02  \pm  0.00$ & $\phantom{1}3.13  \pm  0.02$ & $\phantom{1}3.12  \pm  0.01$ & $\phantom{1}3.18  \pm  0.06$ & $\phantom{1}3.21  \pm  0.01$ & $\phantom{1}3.01  \pm  0.08$ & $\phantom{1}\bm{3.00  \pm  0.00}$ & $\phantom{1}3.18  \pm  0.04$ \\
$M_{43}(3)$ & $\phantom{1}3.01  \pm  0.00$ & $\phantom{1}3.13  \pm  0.02$ & $\phantom{1}3.13  \pm  0.01$ & $\phantom{1}3.41  \pm  0.07$ & $\phantom{1}3.17  \pm  0.01$ & $\phantom{1}3.02  \pm  0.08$ & $\phantom{1}\bm{3.00  \pm  0.00}$ & $\phantom{1}3.17  \pm  0.04$ \\
$M_{5}(2)$ & $\phantom{1}\bm{2.00  \pm  0.00}$ & $\phantom{1}2.02  \pm  0.01$ & $\phantom{1}2.03  \pm  0.01$ & $\phantom{1}2.18  \pm  0.01$ & $\phantom{1}2.05  \pm  0.01$ & $\phantom{1}2.01  \pm  0.06$ & $\phantom{1}1.97  \pm  0.00$ & $\phantom{1}2.24  \pm  0.06$ \\
$M_{6}(1)$ & $\phantom{1}1.01  \pm  0.00$ & $\phantom{1}1.19  \pm  0.07$ & $\phantom{1}1.06  \pm  0.00$ & $\phantom{1}\bm{1.00  \pm  0.00}$ & $\phantom{1}1.11  \pm  0.00$ & $\phantom{1}1.00  \pm  0.03$ & $\phantom{1}\bm{1.00  \pm  0.00}$ & $\phantom{1}2.09  \pm  0.04$ \\
$M_{7}(2)$ & $\phantom{1}2.02  \pm  0.00$ & $\phantom{1}2.42  \pm  0.06$ & $\phantom{1}2.14  \pm  0.02$ & $\phantom{1}2.18  \pm  0.02$ & $\phantom{1}2.14  \pm  0.01$ & $\phantom{1}2.01  \pm  0.06$ & $\phantom{1}\bm{2.00  \pm  0.00}$ & $\phantom{1}2.46  \pm  0.04$ \\
$M_{8}(2)$ & $\phantom{1}\bm{2.00  \pm  0.00}$ & $\phantom{1}1.99  \pm  0.01$ & $\phantom{1}2.02  \pm  0.01$ & $\phantom{1}2.18  \pm  0.03$ & $\phantom{1}2.04  \pm  0.01$ & $\phantom{1}2.00  \pm  0.06$ & $\phantom{1}1.95  \pm  0.01$ & $\phantom{1}2.20  \pm  0.06$ \\
$M_{9}(2)$ & $\phantom{1}2.01  \pm  0.00$ & $\phantom{1}2.11  \pm  0.01$ & $\phantom{1}2.10  \pm  0.01$ & $\phantom{1}2.18  \pm  0.02$ & $\phantom{1}2.13  \pm  0.01$ & $\phantom{1}2.01  \pm  0.06$ & $\phantom{1}\bm{2.00  \pm  0.00}$ & $\phantom{1}2.35  \pm  0.07$ \\
$M_{10}(2)$ & $\phantom{1}\bm{2.00  \pm  0.00}$ & $\phantom{1}2.10  \pm  0.01$ & $\phantom{1}2.11  \pm  0.01$ & $\phantom{1}2.21  \pm  0.03$ & $\phantom{1}2.10  \pm  0.01$ & $\phantom{1}2.01  \pm  0.06$ & $\phantom{1}1.98  \pm  0.00$ & $\phantom{1}2.29  \pm  0.05$ \\
\end{longtable}
}
\end{landscape}

\begin{landscape}
{\small
\setlength{\tabcolsep}{2pt}
\begin{longtable}{@{} l| *{8}{c} @{}}
\caption{Mean dimension estimates ($\pm$ standard deviation) for 18 manifolds (Table~\ref{tab:manifolds}) with true dimension $d$, based on 100 replicates of $n=500$ uniform samples perturbed by Gaussian noise. Bold indicates the estimate with minimal MSE.} \label{tab:06} \\
\toprule
\textbf{Manifold} ($d$) & \textsc{Local PCA} & \textsc{MADA} & \textsc{MLE} & \textsc{DanCo}  & \textsc{TLE} & \textsc{TwoNN}  & \textsc{CA-PCA} & \textsc{Wasserstein} \\
\midrule
\endfirsthead
\multicolumn{9}{c}%
{\tablename\ \thetable\ -- \textit{continued from previous page}} \\
\toprule
\textbf{Manifold} ($d$) & \textsc{Local PCA} & \textsc{MADA} & \textsc{MLE} & \textsc{DanCo}  & \textsc{TLE} & \textsc{TwoNN}  & \textsc{CA-PCA} & \textsc{Wasserstein} \\
\midrule
\endhead
\midrule \multicolumn{9}{r}{\textit{Continued on next page}} \\
\endfoot
\bottomrule
\endlastfoot
$M_{11}(5)$ & $\phantom{1}7.19  \pm  0.07$ & $\phantom{1}\bm{5.22  \pm  0.07}$ & $\phantom{1}5.59  \pm  0.05$ & $\phantom{1}7.49  \pm  0.24$ & $\phantom{1}5.80  \pm  0.05$ & $\phantom{1}6.99  \pm  0.35$ & $\phantom{1}7.35  \pm  0.32$ & $\phantom{1}6.06  \pm  0.11$ \\
$M_{12}(10)$ & $13.27 \pm 0.08$ & $\phantom{1}9.17  \pm  0.10$ & $\phantom{1}\bm{9.67  \pm  0.08}$ & $14.69 \pm 0.41$ & $\phantom{1}9.61  \pm  0.07$ & $11.98 \pm 0.61$ & $19.06 \pm 0.06$ & $10.70 \pm 0.23$ \\
$M_{13}(20)$ & $25.73 \pm 0.11$ & $15.41 \pm 0.15$ & $16.37 \pm 0.12$ & $34.36 \pm 1.01$ & $15.32 \pm 0.11$ & \bm{$20.84 \pm 1.13$} & $35.93 \pm 0.07$ & $18.90 \pm 0.56$ \\
$M_{21}(5)$ & $\phantom{1}8.77  \pm  0.08$ & $\phantom{1}\bm{5.19  \pm  0.06}$ & $\phantom{1}5.65  \pm  0.05$ & $\phantom{1}8.22  \pm  0.22$ & $\phantom{1}5.74  \pm  0.05$ & $\phantom{1}7.52  \pm  0.43$ & $\phantom{1}9.89  \pm  0.03$ & $\phantom{1}6.07  \pm  0.09$ \\
$M_{22}(10)$ & $15.20 \pm 0.12$ & $\phantom{1}8.99  \pm  0.10$ & $\phantom{1}\bm{9.53  \pm  0.08}$ & $14.65 \pm 0.57$ & $\phantom{1}9.35  \pm  0.07$ & $12.25 \pm 0.63$ & $19.50 \pm 0.04$ & $10.88 \pm 0.39$ \\
$M_{23}(20)$ & $25.89 \pm 0.12$ & $15.25 \pm 0.16$ & $16.24 \pm 0.13$ & $35.01 \pm 1.25$ & $15.11 \pm 0.12$ & $21.03 \pm 0.97$ & $36.23 \pm 0.08$ & \bm{$18.97 \pm 0.76$} \\
$M_{31}(5)$ & $\phantom{1}\bm{5.04  \pm  0.02}$ & $\phantom{1}4.45  \pm  0.06$ & $\phantom{1}4.74  \pm  0.06$ & $\phantom{1}6.02  \pm  0.15$ & $\phantom{1}4.91  \pm  0.06$ & $\phantom{1}5.66  \pm  0.33$ & $\phantom{1}5.06  \pm  0.03$ & $\phantom{1}4.80  \pm  0.08$ \\
$M_{32}(10)$ & \bm{$10.00 \pm 0.00$} & $\phantom{1}7.51  \pm  0.09$ & $\phantom{1}7.82  \pm  0.07$ & $10.16 \pm 0.28$ & $\phantom{1}7.79  \pm  0.06$ & $\phantom{1}8.93  \pm  0.50$ & $10.22 \pm 0.08$ & $\phantom{1}8.18  \pm  0.15$ \\
$M_{33}(20)$ & $19.55 \pm 0.03$ & $12.30 \pm 0.14$ & $12.99 \pm 0.10$ & \bm{$20.42 \pm 0.57$} & $12.39 \pm 0.09$ & $15.28 \pm 0.76$ & $21.15 \pm 0.24$ & $14.08 \pm 0.28$ \\
$M_{41}(3)$ & $\phantom{1}5.90  \pm  0.01$ & $\phantom{1}3.59  \pm  0.04$ & $\phantom{1}3.47  \pm  0.03$ & $\phantom{1}4.04  \pm  0.14$ & $\phantom{1}3.78  \pm  0.03$ & $\phantom{1}3.47  \pm  0.18$ & $\phantom{1}\bm{3.44  \pm  0.03}$ & $\phantom{1}3.68  \pm  0.06$ \\
$M_{42}(3)$ & $\phantom{1}5.90  \pm  0.01$ & $\phantom{1}3.59  \pm  0.04$ & $\phantom{1}3.48  \pm  0.03$ & $\phantom{1}4.07  \pm  0.14$ & $\phantom{1}3.78  \pm  0.03$ & $\phantom{1}3.48  \pm  0.18$ & $\phantom{1}\bm{3.46  \pm  0.03}$ & $\phantom{1}3.69  \pm  0.06$ \\
$M_{43}(3)$ & $\phantom{1}5.79  \pm  0.03$ & $\phantom{1}\bm{3.57  \pm  0.05}$ & $\phantom{1}3.58  \pm  0.03$ & $\phantom{1}4.73  \pm  0.19$ & $\phantom{1}3.83  \pm  0.03$ & $\phantom{1}3.88  \pm  0.19$ & $\phantom{1}3.89  \pm  0.05$ & $\phantom{1}3.85  \pm  0.05$ \\
$M_{5}(2)$ & $\phantom{1}3.93  \pm  0.01$ & $\phantom{1}\bm{2.44  \pm  0.05}$ & $\phantom{1}2.68  \pm  0.04$ & $\phantom{1}4.00  \pm  0.00$ & $\phantom{1}2.77  \pm  0.03$ & $\phantom{1}3.85  \pm  0.21$ & $\phantom{1}3.52  \pm  0.10$ & $\phantom{1}3.02  \pm  0.07$ \\
$M_{6}(1)$ & $\phantom{1}2.97  \pm  0.01$ & $\phantom{1}\bm{1.58  \pm  0.04}$ & $\phantom{1}1.67  \pm  0.03$ & $\phantom{1}2.94  \pm  0.09$ & $\phantom{1}2.01  \pm  0.03$ & $\phantom{1}2.77  \pm  0.15$ & $\phantom{1}1.73  \pm  0.04$ & $\phantom{1}2.51  \pm  0.06$ \\
$M_{7}(2)$ & $\phantom{1}2.99  \pm  0.01$ & $\phantom{1}2.96  \pm  0.07$ & $\phantom{1}\bm{2.17  \pm  0.03}$ & $\phantom{1}2.27  \pm  0.04$ & $\phantom{1}2.33  \pm  0.03$ & $\phantom{1}2.18  \pm  0.11$ & $\phantom{1}2.78  \pm  0.04$ & $\phantom{1}2.55  \pm  0.14$ \\
$M_{8}(2)$ & $\phantom{1}3.94  \pm  0.02$ & $\phantom{1}\bm{2.22  \pm  0.03}$ & $\phantom{1}2.45  \pm  0.03$ & $\phantom{1}4.00  \pm  0.02$ & $\phantom{1}2.55  \pm  0.02$ & $\phantom{1}3.75  \pm  0.20$ & $\phantom{1}3.20  \pm  0.10$ & $\phantom{1}2.78  \pm  0.09$ \\
$M_{9}(2)$ & $\phantom{1}3.00  \pm  0.00$ & $\phantom{1}2.40  \pm  0.03$ & $\phantom{1}2.35  \pm  0.02$ & $\phantom{1}2.89  \pm  0.13$ & $\phantom{1}2.55  \pm  0.02$ & $\phantom{1}2.98  \pm  0.15$ & $\phantom{1}\bm{2.10  \pm  0.01}$ & $\phantom{1}2.31  \pm  0.12$ \\
$M_{10}(2)$ & $\phantom{1}3.19  \pm  0.06$ & $\phantom{1}2.23  \pm  0.04$ & $\phantom{1}2.35  \pm  0.03$ & $\phantom{1}3.86  \pm  0.16$ & $\phantom{1}2.37  \pm  0.02$ & $\phantom{1}3.29  \pm  0.18$ & $\phantom{1}\bm{2.15  \pm  0.04}$ & $\phantom{1}2.67  \pm  0.07$ \\
\end{longtable}
}
\end{landscape}

\begin{landscape}
{\small
\setlength{\tabcolsep}{2pt}
\begin{longtable}{@{} l| *{8}{c} @{}}
\caption{Mean dimension estimates ($\pm$ standard deviation) for 18 manifolds (Table~\ref{tab:manifolds}) with true dimension $d$, based on 100 replicates of $n=2{,}000$ uniform samples perturbed by Gaussian noise. Bold indicates the estimate with minimal MSE.} \label{tab:07} \\
\toprule
\textbf{Manifold} ($d$) & \textsc{Local PCA} & \textsc{MADA} & \textsc{MLE} & \textsc{DanCo}  & \textsc{TLE} & \textsc{TwoNN}  & \textsc{CA-PCA} & \textsc{Wasserstein} \\
\midrule
\endfirsthead
\multicolumn{9}{c}%
{\tablename\ \thetable\ -- \textit{continued from previous page}} \\
\toprule
\textbf{Manifold} ($d$) & \textsc{Local PCA} & \textsc{MADA} & \textsc{MLE} & \textsc{DanCo}  & \textsc{TLE} & \textsc{TwoNN}  & \textsc{CA-PCA} & \textsc{Wasserstein} \\
\midrule
\endhead
\midrule \multicolumn{9}{r}{\textit{Continued on next page}} \\
\endfoot
\bottomrule
\endlastfoot
$M_{11}(5)$ & $\phantom{1}8.40  \pm  0.03$ & $\phantom{1}\bm{5.90  \pm  0.04}$ & $\phantom{1}6.22  \pm  0.03$ & $10.00 \pm 0.02$ & $\phantom{1}6.38  \pm  0.03$ & $\phantom{1}7.77  \pm  0.20$ & $\phantom{1}9.78  \pm  0.04$ & $\phantom{1}6.53  \pm  0.05$ \\
$M_{12}(10)$ & $14.56 \pm 0.05$ & \bm{$10.50 \pm 0.05$} & $10.98 \pm 0.05$ & $14.89 \pm 0.21$ & $10.96 \pm 0.04$ & $13.03 \pm 0.32$ & $19.60 \pm 0.02$ & $11.83 \pm 0.20$ \\
$M_{13}(20)$ & $26.17 \pm 0.05$ & $18.08 \pm 0.09$ & $19.00 \pm 0.08$ & $36.49 \pm 0.90$ & $17.99 \pm 0.07$ & $22.68 \pm 0.64$ & $35.38 \pm 0.04$ & \bm{$20.44 \pm 0.21$} \\
$M_{21}(5)$ & $\phantom{1}9.61  \pm  0.02$ & $\phantom{1}\bm{6.14  \pm  0.04}$ & $\phantom{1}6.53  \pm  0.03$ & $10.00 \pm 0.00$ & $\phantom{1}6.58  \pm  0.03$ & $\phantom{1}8.25  \pm  0.24$ & $\phantom{1}9.99  \pm  0.00$ & $\phantom{1}7.06  \pm  0.11$ \\
$M_{22}(10)$ & $16.21 \pm 0.06$ & \bm{$10.46 \pm 0.06$} & $11.01 \pm 0.05$ & $15.02 \pm 0.13$ & $10.84 \pm 0.04$ & $13.33 \pm 0.30$ & $19.83 \pm 0.01$ & $11.77 \pm 0.13$ \\
$M_{23}(20)$ & $26.38 \pm 0.06$ & $18.00 \pm 0.10$ & $18.94 \pm 0.08$ & $37.51 \pm 0.96$ & $17.81 \pm 0.08$ & $22.81 \pm 0.58$ & $35.68 \pm 0.04$ & \bm{$20.46 \pm 0.20$} \\
$M_{31}(5)$ & $\phantom{1}5.87  \pm  0.05$ & $\phantom{1}4.81  \pm  0.03$ & $\phantom{1}\bm{5.05  \pm  0.02}$ & $\phantom{1}6.97  \pm  0.09$ & $\phantom{1}5.18  \pm  0.02$ & $\phantom{1}6.34  \pm  0.17$ & $\phantom{1}5.08  \pm  0.02$ & $\phantom{1}5.27  \pm  0.11$ \\
$M_{32}(10)$ & \bm{$10.00 \pm 0.00$} & $\phantom{1}8.22  \pm  0.05$ & $\phantom{1}8.50  \pm  0.04$ & $10.49 \pm 0.33$ & $\phantom{1}8.55  \pm  0.03$ & $\phantom{1}9.47  \pm  0.27$ & $10.09 \pm 0.02$ & $\phantom{1}8.48  \pm  0.07$ \\
$M_{33}(20)$ & $19.52 \pm 0.02$ & $14.03 \pm 0.06$ & $14.52 \pm 0.05$ & $22.82 \pm 0.62$ & $14.07 \pm 0.05$ & $16.13 \pm 0.37$ & \bm{$20.27 \pm 0.04$} & $14.90 \pm 0.14$ \\
$M_{41}(3)$ & $\phantom{1}3.85  \pm  0.02$ & $\phantom{1}3.32  \pm  0.02$ & $\phantom{1}3.36  \pm  0.01$ & $\phantom{1}4.46  \pm  0.12$ & $\phantom{1}3.58  \pm  0.01$ & $\phantom{1}3.93  \pm  0.10$ & $\phantom{1}\bm{3.03  \pm  0.00}$ & $\phantom{1}3.55  \pm  0.02$ \\
$M_{42}(3)$ & $\phantom{1}3.89  \pm  0.02$ & $\phantom{1}3.33  \pm  0.02$ & $\phantom{1}3.37  \pm  0.01$ & $\phantom{1}4.52  \pm  0.13$ & $\phantom{1}3.59  \pm  0.01$ & $\phantom{1}3.96  \pm  0.10$ & $\phantom{1}\bm{3.04  \pm  0.01}$ & $\phantom{1}3.56  \pm  0.02$ \\
$M_{43}(3)$ & $\phantom{1}4.55  \pm  0.03$ & $\phantom{1}\bm{3.45  \pm  0.02}$ & $\phantom{1}3.55  \pm  0.01$ & $\phantom{1}5.04  \pm  0.19$ & $\phantom{1}3.73  \pm  0.01$ & $\phantom{1}4.47  \pm  0.14$ & $\phantom{1}3.53  \pm  0.04$ & $\phantom{1}3.78  \pm  0.03$ \\
$M_{5}(2)$ & $\phantom{1}4.00  \pm  0.00$ & $\phantom{1}3.26  \pm  0.02$ & $\phantom{1}3.46  \pm  0.02$ & $\phantom{1}4.00  \pm  0.00$ & $\phantom{1}3.49  \pm  0.02$ & $\phantom{1}4.00  \pm  0.12$ & $\phantom{1}3.98  \pm  0.00$ & $\phantom{1}\bm{3.10  \pm  0.16}$ \\
$M_{6}(1)$ & $\phantom{1}2.85  \pm  0.01$ & $\phantom{1}\bm{1.66  \pm  0.01}$ & $\phantom{1}1.88  \pm  0.01$ & $\phantom{1}3.00  \pm  0.00$ & $\phantom{1}2.18  \pm  0.01$ & $\phantom{1}3.01  \pm  0.08$ & $\phantom{1}2.54  \pm  0.01$ & $\phantom{1}2.47  \pm  0.06$ \\
$M_{7}(2)$ & $\phantom{1}2.13  \pm  0.01$ & $\phantom{1}\bm{2.01  \pm  0.01}$ & $\phantom{1}2.04  \pm  0.01$ & $\phantom{1}2.39  \pm  0.02$ & $\phantom{1}2.14  \pm  0.01$ & $\phantom{1}2.50  \pm  0.07$ & $\phantom{1}1.99  \pm  0.01$ & $\phantom{1}2.49  \pm  0.03$ \\
$M_{8}(2)$ & $\phantom{1}4.00  \pm  0.00$ & $\phantom{1}3.01  \pm  0.03$ & $\phantom{1}3.33  \pm  0.02$ & $\phantom{1}4.00  \pm  0.00$ & $\phantom{1}3.37  \pm  0.02$ & $\phantom{1}3.97  \pm  0.11$ & $\phantom{1}3.96  \pm  0.01$ & $\phantom{1}\bm{2.63  \pm  0.16}$ \\
$M_{9}(2)$ & $\phantom{1}3.52  \pm  0.02$ & $\phantom{1}2.25  \pm  0.01$ & $\phantom{1}2.41  \pm  0.01$ & $\phantom{1}3.98  \pm  0.07$ & $\phantom{1}2.56  \pm  0.01$ & $\phantom{1}3.54  \pm  0.10$ & $\phantom{1}\bm{2.17  \pm  0.02}$ & $\phantom{1}2.69  \pm  0.08$ \\
$M_{10}(2)$ & $\phantom{1}3.87  \pm  0.01$ & $\phantom{1}\bm{2.45  \pm  0.02}$ & $\phantom{1}2.72  \pm  0.02$ & $\phantom{1}4.00  \pm  0.00$ & $\phantom{1}2.80  \pm  0.02$ & $\phantom{1}3.76  \pm  0.10$ & $\phantom{1}3.07  \pm  0.04$ & $\phantom{1}2.76  \pm  0.04$ \\
\end{longtable}
}\end{landscape}

\bibliography{bibliography}

@Manual{Rdimtools,
  title  = {Rdimtools: Dimension Reduction and Estimation Methods},
  author = {Kisung You and Changhee Suh and Dennis Shung},
  year   = {2022},
  note   = {R package version 1.1.2},
  url    = {https://CRAN.R-project.org/package=Rdimtools},
}

@incollection{lee2003smooth,
  title={Smooth manifolds},
  author={Lee, John M},
  booktitle={Introduction to smooth manifolds},
  pages={1--29},
  year={2003},
  publisher={Springer},
  address={New York, NY, USA},
}

@book{lee2018introduction,
  title={Introduction to Riemannian manifolds},
  author={Lee, John M},
  volume={2},
  year={2018},
  publisher={Springer},
  address={New York, NY, USA},
}

@incollection{fedorchuk1990fundamentals,
  title={The fundamentals of dimension theory},
  author={Fedorchuk, Vitaly V},
  booktitle={General Topology I: Basic Concepts and Constructions Dimension Theory},
  pages={91--192},
  year={1990},
  publisher={Springer},
  address={Berlin, Germany},
}

@book{falconer2013fractal,
  title={Fractal geometry: mathematical foundations and applications},
  author={Falconer, Kenneth},
  year={2013},
  publisher={John Wiley \& Sons},
  address={Hoboken, NJ, USA}
}

@article{bruske2002intrinsic,
  title={Intrinsic dimensionality estimation with optimally topology preserving maps},
  author={Bruske, J{\"o}rg and Sommer, Gerald},
  journal={IEEE Transactions on pattern analysis and machine intelligence},
  volume={20},
  number={5},
  pages={572--575},
  year={2002},
  publisher={IEEE}
}

@inproceedings{bac2020local,
  title={Local intrinsic dimensionality estimators based on concentration of measure},
  author={Bac, Jonathan and Zinovyev, Andrei},
  booktitle={2020 International Joint Conference on Neural Networks (IJCNN)},
  pages={1--8},
  year={2020},
  organization={IEEE}
}

@article{nash1956imbedding,
  title={The imbedding problem for Riemannian manifolds},
  author={Nash, John},
  journal={Annals of mathematics},
  volume={63},
  number={1},
  pages={20--63},
  year={1956},
  publisher={JSTOR}
}

@article{campadelli2015intrinsic,
  title={Intrinsic dimension estimation: Relevant techniques and a benchmark framework},
  author={Campadelli, Paola and Casiraghi, Elena and Ceruti, Claudio and Rozza, Alessandro},
  journal={Mathematical Problems in Engineering},
  volume={2015},
  number={1},
  pages={759567},
  year={2015},
  publisher={Wiley Online Library}
}

@incollection{diaconis2013sampling,
  title={Sampling from a manifold},
  author={Diaconis, Persi and Holmes, Susan and Shahshahani, Mehrdad},
  booktitle={Advances in modern statistical theory and applications: a Festschrift in honor of Morris L. Eaton},
  volume={10},
  pages={102--126},
  year={2013},
  publisher={Institute of Mathematical Statistics},
  address={Beachwood, OH, USA},
}

@article{marsaglia1972choosing,
  title={Choosing a point from the surface of a sphere},
  author={Marsaglia, George},
  journal={The Annals of Mathematical Statistics},
  volume={43},
  number={2},
  pages={645--646},
  year={1972},
  publisher={Institute of Mathematical Statistics}
}

@article{kim2019minimax,
  title={Minimax rates for estimating the dimension of a manifold},
  author={Kim, Jisu and Rinaldo, Alessandro and Wasserman, Larry},
  journal={Journal of Computational Geometry},
  volume={10},
  number={1},
  pages={42--95},
  year={2019}
}

@article{bennett1969intrinsic,
  title={The intrinsic dimensionality of signal collections},
  author={Bennett, Robert},
  journal={IEEE Transactions on Information Theory},
  volume={15},
  number={5},
  pages={517--525},
  year={1969},
  publisher={IEEE}
}

@article{fukunaga1971algorithm,
  title={An algorithm for finding intrinsic dimensionality of data},
  author={Fukunaga, Keinosuke and Olsen, David R},
  journal={IEEE Transactions on computers},
  volume={100},
  number={2},
  pages={176--183},
  year={1971},
  publisher={IEEE}
}

@article{fan2010intrinsic,
  title={Intrinsic dimension estimation of data by principal component analysis},
  author={Fan, Mingyu and Gu, Nannan and Qiao, Hong and Zhang, Bo},
  journal={arXiv preprint arXiv:1002.2050},
  year={2010}
}

@article{lim2024tangent,
  title={Tangent space and dimension estimation with the wasserstein distance},
  author={Lim, Uzu and Oberhauser, Harald and Nanda, Vidit},
  journal={SIAM Journal on Applied Algebra and Geometry},
  volume={8},
  number={3},
  pages={650--685},
  year={2024},
  publisher={SIAM}
}

@article{cheng2013local,
  title={Local linear regression on manifolds and its geometric interpretation},
  author={Cheng, Ming-Yen and Wu, Hau-tieng},
  journal={Journal of the American Statistical Association},
  volume={108},
  number={504},
  pages={1421--1434},
  year={2013},
  publisher={Taylor \& Francis}
}

@article{aswani2011regression,
  title={REGRESSION ON MANIFOLDS: ESTIMATION OF THE EXTERIOR DERIVATIVE},
  author={Aswani, Anil and Bickel, Peter and Tomlin, Claire},
  journal={The Annals of Statistics},
  volume={39},
  number={1},
  pages={48--81},
  year={2011}
}

@article{chan2017factor,
  title={Factor modelling for high-dimensional time series: inference and model selection},
  author={Chan, Ngai Hang and Lu, Ye and Yau, Chun Yip},
  journal={Journal of Time Series Analysis},
  volume={38},
  number={2},
  pages={285--307},
  year={2017},
  publisher={Wiley Online Library}
}

@inproceedings{farahmand2007manifold,
  title={Manifold-adaptive dimension estimation},
  author={Farahmand, Amir Massoud and Szepesv{\'a}ri, Csaba and Audibert, Jean-Yves},
  booktitle={Proceedings of the 24th international conference on Machine learning},
  pages={265--272},
  year={2007}
}

@article{pettis1979intrinsic,
  title={An intrinsic dimensionality estimator from near-neighbor information},
  author={Pettis, Karl W and Bailey, Thomas A and Jain, Anil K and Dubes, Richard C},
  journal={IEEE Transactions on pattern analysis and machine intelligence},
  pages={25--37},
  year={1979},
  publisher={IEEE}
}

@article{bac2021scikit,
  title={Scikit-dimension: a python package for intrinsic dimension estimation},
  author={Bac, Jonathan and Mirkes, Evgeny M and Gorban, Alexander N and Tyukin, Ivan and Zinovyev, Andrei},
  journal={Entropy},
  volume={23},
  number={10},
  pages={1368},
  year={2021},
  publisher={MDPI}
}

@inproceedings{costa2004learning,
  title={Learning intrinsic dimension and intrinsic entropy of high-dimensional datasets},
  author={Costa, Jose A and Hero, Alfred O},
  booktitle={2004 12th European Signal Processing Conference},
  pages={369--372},
  year={2004},
  organization={IEEE}
}

@article{facco2017estimating,
  title={Estimating the intrinsic dimension of datasets by a minimal neighborhood information},
  author={Facco, Elena and d’Errico, Maria and Rodriguez, Alex and Laio, Alessandro},
  journal={Scientific reports},
  volume={7},
  number={1},
  pages={12140},
  year={2017},
  publisher={Nature Publishing Group UK London}
}

@article{gupta2012regularized,
  title={Regularized maximum likelihood for intrinsic dimension estimation},
  author={Gupta, Mithun Das and Huang, Thomas S},
  journal={arXiv preprint arXiv:1203.3483},
  year={2012}
}

@article{kalantana2013intrinsic,
  title={Intrinsic dimensionality estimation for high-dimensional data sets: New approaches for the computation of correlation dimension},
  author={Kalantana, Jochen Einbecka Zakiah and Einbecka, Z},
  journal={J. Emerg. Technol. Web Intell},
  volume={5},
  number={2},
  pages={91--97},
  year={2013}
}

@article{costa2004geodesic,
  title={Geodesic entropic graphs for dimension and entropy estimation in manifold learning},
  author={Costa, Jose A and Hero, Alfred O},
  journal={IEEE Transactions on Signal Processing},
  volume={52},
  number={8},
  pages={2210--2221},
  year={2004},
  publisher={IEEE}
}

@article{grassberger1983measuring,
  title={Measuring the strangeness of strange attractors},
  author={Grassberger, Peter and Procaccia, Itamar},
  journal={Physica D: nonlinear phenomena},
  volume={9},
  number={1-2},
  pages={189--208},
  year={1983},
  publisher={Elsevier}
}

@article{kegl2002intrinsic,
  title={Intrinsic dimension estimation using packing numbers},
  author={K{\'e}gl, Bal{\'a}zs},
  journal={Advances in neural information processing systems},
  volume={15},
  year={2002}
}

@inproceedings{little2009multiscale,
  title={Multiscale Estimation of Intrinsic Dimensionality of Data Sets.},
  author={Little, Anna V and Jung, Yoon-Mo and Maggioni, Mauro},
  booktitle={AAAI fall symposium: manifold learning and its applications},
  volume={9},
  pages={04},
  year={2009}
}

@article{trunk1968statistical,
  title={Statistical estimation of the intrinsic dimensionality of data collections},
  author={Trunk, Gerard V},
  journal={Information and Control},
  volume={12},
  number={5},
  pages={508--525},
  year={1968},
  publisher={Elsevier}
}

@article{verveer1995evaluation,
  title={An evaluation of intrinsic dimensionality estimators},
  author={Verveer, Peter J. and Duin, Robert P. W.},
  journal={IEEE Transactions on pattern analysis and machine intelligence},
  volume={17},
  number={1},
  pages={81--86},
  year={1995},
  publisher={IEEE}
}

@article{camastra2002estimating,
  title={Estimating the intrinsic dimension of data with a fractal-based method},
  author={Camastra, Francesco and Vinciarelli, Alessandro},
  journal={IEEE Transactions on pattern analysis and machine intelligence},
  volume={24},
  number={10},
  pages={1404--1407},
  year={2002},
  publisher={IEEE}
}

@inproceedings{hein2005intrinsic,
  title={Intrinsic dimensionality estimation of submanifolds in Rd},
  author={Hein, Matthias and Audibert, Jean-Yves},
  booktitle={Proceedings of the 22nd international conference on Machine learning},
  pages={289--296},
  year={2005}
}

@inproceedings{yang2007conical,
  title={Conical dimension as an intrisic dimension estimator and its applications},
  author={Yang, Xin and Michea, Sebastien and Zha, Hongyuan},
  booktitle={Proceedings of the 2007 SIAM International Conference on Data Mining},
  pages={169--179},
  year={2007},
  organization={SIAM}
}

@article{wang2008scale,
  title={A scale-based approach to finding effective dimensionality in manifold learning},
  author={Wang, Xiaohui and Marron, JS},
  journal={Electronic Journal of Statistics},
  volume={2},
  year={2008},
  publisher={Institute of Mathematical Statistics}
}

@article{fan2009intrinsic,
  title={Intrinsic dimension estimation of manifolds by incising balls},
  author={Fan, Mingyu and Qiao, Hong and Zhang, Bo},
  journal={Pattern Recognition},
  volume={42},
  number={5},
  pages={780--787},
  year={2009},
  publisher={Elsevier}
}

@inproceedings{kleindessner2015dimensionality,
  title={Dimensionality estimation without distances},
  author={Kleindessner, Matth{\"a}us and Luxburg, Ulrike},
  booktitle={Artificial Intelligence and Statistics},
  pages={471--479},
  year={2015},
  organization={PMLR}
}

@article{carter2009local,
  title={On local intrinsic dimension estimation and its applications},
  author={Carter, Kevin M and Raich, Raviv and Hero III, Alfred O},
  journal={IEEE Transactions on Signal Processing},
  volume={58},
  number={2},
  pages={650--663},
  year={2009},
  publisher={IEEE}
}

@inproceedings{sricharan2010optimized,
  title={Optimized intrinsic dimension estimator using nearest neighbor graphs},
  author={Sricharan, Kumar and Raich, Raviv and Hero, Alfred O},
  booktitle={2010 IEEE International Conference on Acoustics, Speech and Signal Processing},
  pages={5418--5421},
  year={2010},
  organization={IEEE}
}

@article{bouveyron2011intrinsic,
  title={Intrinsic dimension estimation by maximum likelihood in isotropic probabilistic PCA},
  author={Bouveyron, Charles and Celeux, Gilles and Girard, St{\'e}phane},
  journal={Pattern Recognition Letters},
  volume={32},
  number={14},
  pages={1706--1713},
  year={2011},
  publisher={Elsevier}
}

@inproceedings{gomtsyan2019geometry,
  title={Geometry-aware maximum likelihood estimation of intrinsic dimension},
  author={Gomtsyan, Marina and Mokrov, Nikita and Panov, Maxim and Yanovich, Yury},
  booktitle={Asian Conference on Machine Learning},
  pages={1126--1141},
  year={2019},
  organization={PMLR}
}

@inproceedings{rozza2011idea,
  title={IDEA: Intrinsic dimension estimation algorithm},
  author={Rozza, Alessandro and Lombardi, Gabriele and Rosa, Marco and Casiraghi, Elena and Campadelli, Paola},
  booktitle={International Conference on Image Analysis and Processing},
  pages={433--442},
  year={2011},
  organization={Springer}
}

@article{rozza2012novel,
  title={Novel high intrinsic dimensionality estimators},
  author={Rozza, Alessandro and Lombardi, Gabriele and Ceruti, Claudio and Casiraghi, Elena and Campadelli, Paola},
  journal={Machine learning},
  volume={89},
  number={1},
  pages={37--65},
  year={2012},
  publisher={Springer}
}

@article{liao2014intrinsic,
  title={Intrinsic dimension estimation via nearest constrained subspace classifier},
  author={Liao, Liang and Zhang, Yanning and Maybank, Stephen John and Liu, Zhoufeng},
  journal={Pattern recognition},
  volume={47},
  number={3},
  pages={1485--1493},
  year={2014},
  publisher={Elsevier}
}

@article{johnsson2014low,
  title={Low bias local intrinsic dimension estimation from expected simplex skewness},
  author={Johnsson, Kerstin and Soneson, Charlotte and Fontes, Magnus},
  journal={IEEE transactions on pattern analysis and machine intelligence},
  volume={37},
  number={1},
  pages={196--202},
  year={2014},
  publisher={IEEE}
}

@article{camastra2016intrinsic,
  title={Intrinsic dimension estimation: Advances and open problems},
  author={Camastra, Francesco and Staiano, Antonino},
  journal={Information Sciences},
  volume={328},
  pages={26--41},
  year={2016},
  publisher={Elsevier}
}

@article{you2022rdimtools,
  title={Rdimtools: An R package for dimension reduction and intrinsic dimension estimation},
  author={You, Kisung and Shung, Dennis},
  journal={Software Impacts},
  volume={14},
  pages={100414},
  year={2022},
  publisher={Elsevier}
}

@inproceedings{amsaleg2019intrinsic,
  title={Intrinsic dimensionality estimation within tight localities},
  author={Amsaleg, Laurent and Chelly, Oussama and Houle, Michael E and Kawarabayashi, Ken-Ichi and Radovanovi{\'c}, Milo{\v{s}} and Treeratanajaru, Weeris},
  booktitle={Proceedings of the 2019 SIAM international conference on data mining},
  pages={181--189},
  year={2019},
  organization={SIAM}
}

@article{amsaleg2018extreme,
  title={Extreme-value-theoretic estimation of local intrinsic dimensionality},
  author={Amsaleg, Laurent and Chelly, Oussama and Furon, Teddy and Girard, St{\'e}phane and Houle, Michael E and Kawarabayashi, Ken-ichi and Nett, Michael},
  journal={Data Mining and Knowledge Discovery},
  volume={32},
  number={6},
  pages={1768--1805},
  year={2018},
  publisher={Springer}
}

@article{aghajanyan2020intrinsic,
  title={Intrinsic dimensionality explains the effectiveness of language model fine-tuning},
  author={Aghajanyan, Armen and Zettlemoyer, Luke and Gupta, Sonal},
  journal={arXiv preprint arXiv:2012.13255},
  year={2020}
}

@article{thordsen2022abid,
  title={ABID: angle based intrinsic dimensionality—theory and analysis},
  author={Thordsen, Erik and Schubert, Erich},
  journal={Information Systems},
  volume={108},
  pages={101989},
  year={2022},
  publisher={Elsevier}
}

@article{meilua2024manifold,
  title={Manifold learning: What, how, and why},
  author={Meil{\u{a}}, Marina and Zhang, Hanyu},
  journal={Annual Review of Statistics and Its Application},
  volume={11},
  number={1},
  pages={393--417},
  year={2024},
  publisher={Annual Reviews}
}

@article{CERUTI20142569,
title = {DANCo: An intrinsic dimensionality estimator exploiting angle and norm concentration},
journal = {Pattern Recognition},
volume = {47},
number = {8},
pages = {2569-2581},
year = {2014},
issn = {0031-3203},
author = {Claudio Ceruti and Simone Bassis and Alessandro Rozza and Gabriele Lombardi and Elena Casiraghi and Paola Campadelli},
}

@article{levina2004maximum,
  title={Maximum likelihood estimation of intrinsic dimension},
  author={Levina, Elizaveta and Bickel, Peter},
  journal={Advances in neural information processing systems},
  volume={17},
  year={2004}
}

@book{mardia2009directional,
  title={Directional statistics},
  author={Mardia, Kanti V and Jupp, Peter E},
  year={2009},
  publisher={John Wiley \& Sons},
  address={Chichester, UK}
}

@article{gilbert2025pca,
  title={Ca-pca: Manifold dimension estimation, adapted for curvature},
  author={Gilbert, Anna C and O’Neill, Kevin},
  journal={SIAM Journal on Mathematics of Data Science},
  volume={7},
  number={1},
  pages={355--383},
  year={2025},
  publisher={SIAM}
}

@article{block2022intrinsic,
  title={Intrinsic dimension estimation using Wasserstein distance},
  author={Block, Adam and Jia, Zeyu and Polyanskiy, Yury and Rakhlin, Alexander},
  journal={Journal of Machine Learning Research},
  volume={23},
  number={313},
  pages={1--37},
  year={2022}
}

@book{villani2021topics,
  title={Topics in optimal transportation},
  author={Villani, C{\'e}dric},
  volume={58},
  year={2021},
  publisher={American Mathematical Soc.},
  address={Providence, RI, USA},
}

@article{dudley1969speed,
  title={The speed of mean Glivenko-Cantelli convergence},
  author={Dudley, Richard Mansfield},
  journal={The Annals of Mathematical Statistics},
  volume={40},
  number={1},
  pages={40--50},
  year={1969},
  publisher={JSTOR}
}

@article{balasubramanian2002isomap,
  title={The isomap algorithm and topological stability},
  author={Balasubramanian, Mukund and Schwartz, Eric L},
  journal={Science},
  volume={295},
  number={5552},
  pages={7--7},
  year={2002},
  publisher={American Association for the Advancement of Science}
}

@article{coifman2006diffusion,
  title={Diffusion maps},
  author={Coifman, Ronald R and Lafon, St{\'e}phane},
  journal={Applied and computational harmonic analysis},
  volume={21},
  number={1},
  pages={5--30},
  year={2006},
  publisher={Elsevier}
}

@article{lecun2002gradient,
  title={Gradient-based learning applied to document recognition},
  author={LeCun, Yann and Bottou, L{\'e}on and Bengio, Yoshua and Haffner, Patrick},
  journal={Proceedings of the IEEE},
  volume={86},
  number={11},
  pages={2278--2324},
  year={2002},
  publisher={Ieee}
}

@misc{asuncion2007uci,
  title={UCI machine learning repository},
  author={Asuncion, Arthur and Newman, David and others},
  year={2007},
  publisher={Irvine, CA, USA}
}

@article{denti2022generalized,
  title={The generalized ratios intrinsic dimension estimator},
  author={Denti, Francesco and Doimo, Diego and Laio, Alessandro and Mira, Antonietta},
  journal={Scientific Reports},
  volume={12},
  number={1},
  pages={20005},
  year={2022},
  publisher={Nature Publishing Group UK London}
}

@incollection{keogh2011curse,
  title={Curse of dimensionality},
  author={Keogh, Eamonn and Mueen, Abdullah},
  booktitle={Encyclopedia of machine learning},
  pages={257--258},
  year={2011},
  publisher={Springer},
  address={Boston, MA, USA},
}

@article{alberti2024manifold,
  title={Manifold learning by mixture models of VAEs for inverse problems},
  author={Alberti, Giovanni S and Hertrich, Johannes and Santacesaria, Matteo and Sciutto, Silvia},
  journal={Journal of Machine Learning Research},
  volume={25},
  number={202},
  pages={1--35},
  year={2024}
}

@book{zhou2021machine,
  title={Machine learning},
  author={Zhou, Zhi-Hua},
  year={2021},
  publisher={Springer nature},
  address={Singapore},
}

@article{floryan2022data,
  title={Data-driven discovery of intrinsic dynamics},
  author={Floryan, Daniel and Graham, Michael D},
  journal={Nature Machine Intelligence},
  volume={4},
  number={12},
  pages={1113--1120},
  year={2022},
  publisher={Nature Publishing Group UK London}
}

@article{binnie2025survey,
  title={A Survey of Dimension Estimation Methods},
  author={Binnie, James AD and Harvey, John and Malinowski, Jakub and Yim, Ka Man and others},
  journal={arXiv preprint arXiv:2507.13887},
  year={2025}
}

@article{lombardi2020intrinsic,
  title={Intrinsic dimensionality estimation techniques},
  author={Lombardi, Gabriele},
  journal={MATLAB Central File Exchange},
  year={2020}
}

\end{document}